\documentclass{ieeeaccess}
\pdfoutput=1
\usepackage{algpseudocode}
\usepackage[acronym,toc,shortcuts]{glossaries}
\usepackage{soul}
\usepackage{framed}
\usepackage{booktabs}
\usepackage{array}
\usepackage{multirow}

\usepackage[inline]{enumitem}

\usepackage[utf8]{inputenc} 
\usepackage[T1]{fontenc} 

\usepackage[norelsize, linesnumbered, ruled, lined, boxed, commentsnumbered]{algorithm2e}
\usepackage{qtree}

\newcommand{\nosemic}{\renewcommand{\@endalgocfline}{\relax}}
\newcommand{\dosemic}{\renewcommand{\@endalgocfline}{\algocf@endline}}

\let\oldnl\nl
\newcommand{\nonl}{\renewcommand{\nl}{\let\nl\oldnl}}
\PassOptionsToPackage{figuresleft}{rotating}
\usepackage{cite}
\usepackage{amsmath,amssymb,amsfonts}
\usepackage{graphicx}
\usepackage{textcomp}
\usepackage{array}
\newcolumntype{?}[1]{!{\vrule width #1}}
\usepackage{arydshln}
\usepackage{makecell}
\usepackage[utf8]{inputenc}
\usepackage[T1]{fontenc}
\usepackage{hyperref}
\usepackage{cleveref}
\usepackage{enumitem}
\usepackage{rotating}
\usepackage{multirow}
\usepackage{url}
\usepackage{comment}
\usepackage{mathabx}
\usepackage{relsize}
\ifCLASSOPTIONcompsoc
 \usepackage[caption=false,font={sf,scriptsize},labelfont={bf,color=accessblue}]{subfig}
\else
 \usepackage[caption=false,font={sf,scriptsize},labelfont={bf,color=accessblue}]{subfig}
\fi
\def\BibTeX{{\rm B\kern-.05em{\sc i\kern-.025em b}\kern-.08em
 T\kern-.1667em\lower.7ex\hbox{E}\kern-.125emX}}
            
\usepackage{mathtools}
\DeclarePairedDelimiter\round{\lfloor}{\rceil}

\usepackage{silence}
\usepackage{xcolor}
\usepackage{pgfplots}
\usepackage{pgfplotstable}
\pgfplotsset{compat=1.7}
\usepackage{tikz}
\NewSpotColorSpace{PANTONE}
\AddSpotColor{PANTONE} {PANTONE3015C} {PANTONE\SpotSpace 3015\SpotSpace C} {1 0.3 0 0.2}
\SetPageColorSpace{PANTONE}

\colorlet{yellow}{yellow!90}
\colorlet{pink}{red!60}
\colorlet{blue}{cyan!90}
\colorlet{white}{white}
\definecolor{DarkBlue}{RGB}{0, 0, 0}  % {0, 0, 255}

\usetikzlibrary{fit,calc}
\newcommand{\tikzmk}[1]{\tikz[remember picture,overlay] \node (#1) {};\ignorespaces}

\newcommand{\boxitLLS}[1]{\tikz[remember picture,overlay]{\node[yshift=3pt,fill=#1,fill opacity=.15,text opacity=1,fit={($(A)+(.005\linewidth,0.32\baselineskip)$)($(B)+(.885\linewidth,.8\baselineskip)$)}] {};}\ignorespaces}
\newcommand{\boxitLL}[1]{\tikz[remember picture,overlay]{\node[yshift=3pt,fill=#1,fill opacity=.15,text opacity=1,fit={($(A)+(.005\linewidth,0.13\baselineskip)$)($(B)+(.885\linewidth,.8\baselineskip)$)}] {};}\ignorespaces}
\newcommand{\boxitLLNBS}[1]{\tikz[remember picture,overlay]{\node[yshift=3pt,fill=#1,fill opacity=.15,text opacity=1,fit={($(A)+(.005\linewidth,0.13\baselineskip)$)($(B)+(.885\linewidth,1\baselineskip)$)}] {};}\ignorespaces}
\newcommand{\boxitLLLS}[1]{\tikz[remember picture,overlay]{\node[yshift=3pt,fill=#1,fill opacity=.15,text opacity=1,fit={($(A)+(.005\linewidth,0.32\baselineskip)$)($(B)+(.825\linewidth,.8\baselineskip)$)}] {};}\ignorespaces}
\newcommand{\boxitLLL}[1]{\tikz[remember picture,overlay]{\node[yshift=3pt,fill=#1,fill opacity=.15,text opacity=1,fit={($(A)+(.005\linewidth,0.13\baselineskip)$)($(B)+(.825\linewidth,.8\baselineskip)$)}] {};}\ignorespaces}

\SetKwFor{ForAll}{for all}{in parallel do}{end}
\SetKwFor{Jump}{\texttt{start}}{:}{}

\usepackage{stfloats}

\usepackage{soul}

\usepackage{bm}

\makeatletter
\def\blfootnote{\xdef\@thefnmark{}\@footnotetext}
\makeatother

\begin{document}
\history{Date of publication 2022 03, 08, date of current version 2022 03, 08.}
\doi{10.1109/ACCESS.2022.3157893}

\title{$\text{FxP-QNet: A Post-Training Quantizer for}$ $\text{the Design of Mixed Low-Precision DNNs}$ $\text{with Dynamic Fixed-Point Representation}$}

\author{
\uppercase{Ahmad Shawahna}\authorrefmark{1},
\uppercase{Sadiq M. Sait}\authorrefmark{1, 3}, \IEEEmembership{Senior Member, IEEE},
\uppercase{Aiman El-Maleh}\authorrefmark{1, 3}, \IEEEmembership{Member, IEEE},
\uppercase{and Irfan Ahmad}\authorrefmark{2, 3}}
\address[1]{Department of Computer Engineering, King Fahd University of Petroleum \& Minerals, Dhahran-31261, Saudi Arabia}
\address[2]{Information and Computer Science Department, King Fahd University of Petroleum \& Minerals, Dhahran-31261, Saudi Arabia}
\address[3]{Interdisciplinary Research Center for Intelligent Secure Systems, King Fahd University of Petroleum \& Minerals,
Dhahran-31261, Saudi Arabia}
\tfootnote{This work was supported by the King Fahd University of Petroleum \& Minerals, Dhahran, Saudi Arabia.
}
\markboth
{Shawahna \headeretal: A Post-Training Quantizer for the Design of Mixed Low-Precision DNNs with Dynamic Fixed-Point Representation}
{Shawahna \headeretal: A Post-Training Quantizer for the Design of Mixed Low-Precision DNNs with Dynamic Fixed-Point Representation}

\corresp{Corresponding author: Ahmad Shawahna (e-mail: g201206920@kfupm.edu.sa).}

\begin{abstract}
Deep neural networks (DNNs) have demonstrated their effectiveness in a wide range of computer vision tasks, with the state-of-the-art results obtained through complex and deep structures that require intensive computation and memory. In the past, graphic processing units enabled these breakthroughs because of their greater computational speed. Now-a-days, efficient model inference is crucial for consumer applications on resource-constrained platforms. As a result, there is much interest in the research and development of dedicated deep learning (DL) hardware to improve the throughput and energy efficiency of DNNs. Low-precision representation of DNN data-structures through quantization would bring great benefits to specialized DL hardware especially when expensive floating-point operations can be avoided and replaced by more efficient fixed-point operations. However, the rigorous quantization leads to a severe accuracy drop. As such, quantization opens a large hyper-parameter space at bit-precision levels, the exploration of which is a major challenge. In this paper, we propose a novel framework referred to as the \textbf{F}i\textbf{x}ed-\textbf{P}oint \textbf{Q}uantizer of deep neural \textbf{Net}works (${\textnormal{\textbf{FxP-QNet}}}$) that flexibly designs a mixed low-precision DNN for integer-arithmetic-only deployment. Specifically, the ${\textnormal{\textbf{FxP-QNet}}}$ gradually adapts the quantization level for each data-structure of each layer based on the trade-off between the network accuracy and the low-precision requirements. Additionally, it employs post-training self-distillation and network prediction error statistics to optimize the quantization of floating-point values into fixed-point numbers. Examining ${\textnormal{\textbf{FxP-QNet}}}$\footnotemark on state-of-the-art architectures and the benchmark ImageNet dataset, we empirically demonstrate the effectiveness of ${\textnormal{\textbf{FxP-QNet}}}$ in achieving the accuracy-compression trade-off without the need for training. The results show that ${\textnormal{\textbf{FxP-QNet}}}$-quantized AlexNet, VGG-16, and ResNet-18 reduce the overall memory requirements of their full-precision counterparts by $7.16\times$, $10.36\times$, and $6.44\times$ with less than $0.95$\%, $0.95$\%, and $1.99$\% accuracy drop, respectively.

\end{abstract}


\titlepgskip=-30pt
\maketitle

\section{Introduction}\label{sec:introduction}

%\blfootnote{©2022 IEEE. Personal use of this material is permitted. Permission from IEEE must be obtained for all other uses, in any current or future media, including reprinting/republishing this material for advertising or promotional purposes, creating new collective works, for resale or redistribution to servers or lists, or reuse of any copyrighted component of this work in other works.}
\blfootnote{The associate editor coordinating the review of this manuscript and approving it for publication was Paolo Crippa.}
\footnotetext{The code of the proposed framework will be available online in \textit{\textbf{https://github.com/Ahmad-Shawahna/FxP-QNet}}.}

\IEEEPARstart{D}{eep} neural networks (DNNs) have achieved state-of-the-art results on computer vision tasks, such as image classification~\cite{matiz2019inductive, huang2017densely, he2016deep, ren2015faster, szegedy2015going, krizhevsky2012imagenet, simonyan2014very}, object detection~\cite{ren2016faster, girshick2014rich}, face recognition~\cite{li2015convolutional}, and semantic segmentation~\cite{wei2019m3net, he2017mask, long2015fully}. Examples of recent fields of DNN applications include robot navigation, industrial automation, automated driving, aerospace and defense~\cite{wu2017squeezedet, goodfellow2016deep, deng2014deep}. These applications heavily depend on the complex DNN structures as well as on the deep and complicated models that over-fit the distribution of numerous training data. However, this also leads to over-parameterization resulting in a dramatic increase in memory and computational requirements for deployment.

A typical DNN often takes hundreds of MB of memory space. For instance, 170MB for ResNet-101~\cite{he2016deep}, 240MB for AlexNet~\cite{krizhevsky2012imagenet}, and 550MB for VGG-16~\cite{simonyan2014very}, and requires billions of floating-point operations (FLOPs) per image during inference that rely on powerful graphic processing units (GPUs), e.g., 0.72GFLOPs for AlexNet, 1.8GFLOPs for ResNet-18, and 15.3GFLOPs for VGG-16. This not only limits the real-time applications of pre-trained DNNs but also challenges the ubiquitous deployment of high-performance DNNs on embedded wearables and IoT devices such as smartphones and drones. Additionally, with the drive to put DNNs on low-power devices (unlike GPUs) now-a-days, developers and researchers often find themselves constrained by computational capabilities, such as the case when high-precision multiply-and-accumulate (MAC) instructions are not supported for hardware efficiency purposes~\cite{li2020survey}. Thus, network compression and acceleration are important issues in deep learning research and application. 

A promising approach to overcome the aforementioned challenges is to reduce the bit-precision level of DNN data-structures by quantization~\cite{sung2015resiliency}. With the network topology unchanged, quantization has the potential to reduce the model size to only a fraction of the original by utilizing low-precision representation of parameters. An example of quantizing the parameters is to represent them with a lightweight floating-point format that uses a configurable number of bits for the exponent and mantissa segments~\cite{liu2021improving}. Furthermore, the weights and activations could also be quantized to discrete values in the fixed-point representation. Then, the network inference is significantly accelerated by increasing the arithmetic intensity, that is, the ratio of operations performed to each memory byte accessed. Additionally, converting expensive floating-point arithmetic to more effective fixed-point and bit-wise operations gives an additional boost by lowering the inference latency~\cite{kouris2018cascade}.

Besides its benefits on memory and computation, quantization also offers large power savings~\cite{horowitz20141, han2016eie, chen2016eyeriss, sze2017efficient}. Reducing the wordlength of neurons also benefits DNN hardware accelerators such as application specific integrated circuits (ASICs) and field programmable gate arrays (FPGAs)~\cite{shawahna2018fpga}, because it means significantly lessening the resources required for calculations such as area and power requirements for multipliers. Table~\ref{tab:power_area_improvement} demonstrates the improvement in power consumption and area requirement when using $8$-bit fixed-point multiply-accumulate units compared to their fixed/floating-point counterparts in 45nm technology (0.9V)~\cite{horowitz20141}. The improvement in computation, power, and hardware efficiency through network quantization makes the deployment of DNNs on dedicated embedded systems feasible now-a-days. Thus, quantization has received considerable attention and has become an active research area~\cite{krishnamoorthi2018quantizing}.

However, an extremely low-precision representation of data-structures leads to a severe information loss, referred to as a quantization error. The substantial quantization error, in turn, leads to a significant degradation in accuracy. For instance, the binary neural networks (BNNs)~\cite{qin2020binary, bethge2018training, hubara2017quantized, hubara2016binarized, courbariaux2016binarized, rastegari2016xnor, hou2016loss} adopt a rigorous quantization method to reduce the wordlength of activations and weights to only $1$ bit. In a similar way, the ternary weight networks (TWNs)~\cite{li2016ternary, zhu2016trained, achterhold2018variational, hou2018loss} use aggressively quantized weights to benefit from the $16\times$ smaller model size. Accordingly, the BNNs and TWNs perform computations using bit-wise operations, but this comes at the cost of sacrificing model's accuracy. Even though several techniques have been proposed to alleviate BNNs/TWNs performance drop, including, but not limited to, optimizing the quantization scheme and the training algorithms, there is still a non-negligible accuracy gap between BNNs/TWNs and their full-precision counterparts. Therefore, researchers moved on to design $k$-bit DNNs, $k > 2$, with the intent of optimizing not only network accuracy but also reducing computational and memory costs~\cite{zhou2016dorefa}.

\begin{table}[tb!]
\renewcommand{\arraystretch}{1.4}
\begin{center}
\captionsetup{justification=centering}
\caption{Improvement in power consumption and area requirement when using 8-bit fixed-point multiply-accumulate operations compared to their fixed-point and floating-point counterparts in 45nm technology (0.9V)~\cite{horowitz20141}.}
\label{tab:power_area_improvement}
\setlength\tabcolsep{2.9pt}
\begin{tabular}{| l | c | c | c | c |}
\hline
\multicolumn{1}{| c |}{\multirow{2}{*}{\textbf{Bit-Precision}}} & \multicolumn{2}{ c |}{\textbf{Multiply}} & \multicolumn{2}{ c |}{\textbf{Accumulate}} \\ 
\cline{2-5}
& \textbf{\,Power\,} & \textbf{\,\,\,Area\,\,\,} & \textbf{\,Power\,} & \textbf{\,\,\,Area\,\,\,} \\
\hline\hline
\textbf{32-bit fixed-point} & 15.5$\times$ & 12.4$\times$ & 3.3$\times$ & 3.8$\times$ \\ 
\hline
\textbf{16-bit floating-point} & 5.5$\times$ & 5.8$\times$ & 13.3$\times$ & 37.8$\times$ \\ 
\hline
\textbf{32-bit floating-point} & 18.5$\times$ & 27.5$\times$ & 30$\times$ & 116$\times$ \\ 
\hline
\end{tabular}
\end{center}
\end{table}

In pursuit of maintaining the overall accuracy for the quantized DNNs, the majority of literature incorporates the quantization with the training process. In this way, the optimization algorithm used to train DNNs, e.g., stochastic gradient descent, will learn the network parameters that yield the best performance, and thus the accuracy degradation due to quantization is minimized. Hence, this approach is referred to as quantization-aware training. However, training requires a full-size dataset, which is often unavailable in real-world scenarios for reasons such as proprietary and privacy, especially when working on an off-the-shelf pre-trained model from community or industry. Training also requires tight integration of network design, training algorithms, initialization, and hyper-parameters configuration, all of which are not always feasible.

Therefore, many efforts have been made to quantize a pre-trained model using sophisticated methods with the intent of minimizing the perturbation in the distribution of quantized data-structures. Hence, this quantization approach is denoted by post-training quantization. Quantization methods utilizing this approach are easy to use and allow DNNs to be quantized even with limited data. Further, quantized DNNs based on a post-training quantization scheme are the fastest ones to be implemented and deployed. This is because they avoid retraining, a time-consuming process~\cite{han2015deep}, which is usually employed repeatedly in quantization-aware training.

Nevertheless, achieving near state-of-the-art accuracy on the large-scale image classification tasks such as ImageNet~\cite{russakovsky2015imagenet} with low-precision data-structures and without retraining remains a challenge. Generally, $8$-bit is considered to be a limit for post-training quantization of DNNs~\cite{gupta2015deep}. This gap has led to extensive research efforts by several tech giants to improve post-training quantization, such as, but not limited to, NVIDIA~\cite{migacz20178}, Samsung~\cite{lee2018quantization}, IBM~\cite{jain2018compensated}, Huawei~\cite{kravchik2019low}, Hailo Technologies~\cite{meller2019same}, and Intel~\cite{banner2019post}.

Most of the work to date homogeneously quantizes all layers, i.e., the same number of bits are allocated to all the layers. Yet, it has been asserted that DNN layers have a different structure and therefore different properties related to the quantization process~\cite{zhang2019all}. Furthermore, DNN layers have a different arithmetic intensity on hardware. Spurred by that, very recently, there have been a few approaches targeting the use of mixed-precision for representing DNN data-structures to design more efficient quantized DNNs~\cite{micikevicius2018mixed, wang2019haq, capotondi2020cmix, rajagopal2020multi}. This flexibility was originally not supported by chip vendors until recently when some advanced chips were released, including Apple A12 Bionic chip, Nvidia's Turing GPU architecture, and Imagination neural network accelerator IP, all of these support mixed-precision arithmetic~\cite{venieris2021reach}. Besides industry, academia also is working on bit-level flexible hardware design to accelerating DNNs. For instance, BitFusion~\cite{sharma2018bit} supports spatial multiplications of $2$, $4$, $8$, and $16$ bits, while BISMO~\cite{umuroglu2018bismo}, on the other hand, adopts a bit-serial multiplier to support scalable precision.

The problem now is, how to determine the wordlength of the data-structures for each layer so that network performance is preserved while minimizing computational and memory costs? Here, it must be emphasized that there is no systematic way to determine the quantization level of different data-structures. In~\cite{micikevicius2018mixed, mishra2018apprentice}, deciding the appropriate bit-precision level is manual and laborious so as to maintain accuracy. Thereafter, automated algorithms are designed that can discover the appropriate quantization level for each data-structure with accuracy in mind~\cite{wu2018mixed, khan2020learning}.

However, most of these methods rely on performing expensive MAC operations with low-precision integer arithmetic, and then multiplying the outputs with floating-point scalar(s) to improve accuracy by recovering model's dynamic range. The problem with all of these methods is extra memory and computational requirements for floating-point scaling. Thus, FPGA/ASIC-based hardware accelerators of DNNs still need to implement expensive floating-point multiplication and accumulation operations to handle activation update. For its significant benefits to device designers and application developers, the flexible design of efficient DNNs with mixed-precision for integer-only deployment has become an interesting research area.

Hence, in this paper, we review the current status of using quantization for the design of low-precision DNNs. We highlight the implementation details of existing techniques and state their drawbacks. Then, a new automated framework for fixed-point quantization of DNN data-structures referred to as the \textbf{F}i\textbf{x}ed-\textbf{P}oint \textbf{Q}uantizer of deep neural \textbf{Net}works (${\textnormal{\textbf{FxP-QNet}}}$) is proposed. The ${\textnormal{\textbf{FxP-QNet}}}$ flexibly designs a mixed low-precision DNN for integer-arithmetic-only deployment from a pre-trained model. The contributions of our paper can be summarized as below: \vspace{-0.4mm}\begin{enumerate}
\item We introduce ${\textnormal{\textbf{FxP-QNet}}}$ to design a mixed low-precision DNN based on the trade-off between network accuracy and low-precision requirements. The ${\textnormal{\textbf{FxP-QNet}}}$ employs a two-level clustering approach and the sensitivity of data-structures to quantization to gradually reduce the wordlength of each data-structure.
\item A detailed description of a post-training quantization method to represent both model parameters and activations as dynamic fixed-point numbers to implement highly resource/energy-efficient FPGA/ASIC-based hardware accelerators of DNNs.
\item We present a metric to quantify the effect of discretization on DNNs using the post-training self-distillation and network prediction error statistics and utilize it to optimize the quantization of floating-point values.
\item Three typical classification DNNs, including AlexNet, VGG, and ResNet-18 are designed with mixed low-precision data-structures for integer-only deployment based on the proposed ${\textnormal{\textbf{FxP-QNet}}}$ framework.
\item Examining the quantized architectures on the benchmark ImageNet dataset, we show that ${\textnormal{\textbf{FxP-QNet}}}$-quantized AlexNet, VGG-16, and ResNet-18 requires $85.99$\%, $91.62$\% and $84.96$\% less memory space for model parameters with less than $0.95$\%, $0.95$\%, and $1.99$\% accuracy drop compared to their full-precision counterparts, respectively.
\end{enumerate}\vspace{-0.4mm}

The rest of the paper is organized as follows. Section~{\color{DarkBlue}\ref{sec:Background_and_Terminology}} provides background information about the dynamic fixed-point representation and highlights its terminology. It also discusses the principally used quantization approaches and the benefits of each on hardware systems. Section~{\color{DarkBlue}\ref{sec:literature_review}} presents a literature review of the existing DNNs quantization frameworks. In Section~{\color{DarkBlue}\ref{sec:functional_components}}, we discuss the proposed ${\textnormal{\textbf{FxP-QNet}}}$ framework and describe its main functional components together with their implementation details. Section~{\color{DarkBlue}\ref{sec:experiments_results}} demonstrates the effectiveness of ${\textnormal{\textbf{FxP-QNet}}}$ via extensive experiments on several typical DNNs architectures. Finally, we conclude and describe the future directions of this paper in Section~{\color{DarkBlue}\ref{sec:conclusion}}.

\section{Background and Terminology} \label{sec:Background_and_Terminology}

This section provides an overview of the basics and terminologies of dynamic fixed-point representation. Additionally, it illustrates the quantization approaches used to map full-precision values to low-precision ones and highlights how hardware systems can benefit from these approaches.

\subsection{Dynamic Fixed-Point Representation} \label{sec:fixed_point_representation}

In fixed-point representation, a real number $x$ is represented as an integer number $\mathbf{x}$ with $k$ bits. An implied binary point is used to configure the integer length $\alpha$ and the fractional length $\beta$ of the binary word. Here, $k$ is the wordlength, $\alpha$ and $\beta \in \mathbb{Z}$, and $k = \alpha + \beta$. Generally, fixed-point representation is applied on a group basis. In other words, a group of $n$ numbers, each of which with $k$ bits, share a common fixed scaling factor equal to $2^{-\beta}$. That is why it is called \textit{fixed-point}. Thus, each integer number $\mathbf{X}_i$ in the group is interpreted to a rational number $\widetilde{X}_i$ expressible as

\begin{equation} \label{eq:fixed_point}
\widetilde{X}_i = \frac{\mathbf{X}_i}{2^\beta} = \mathbf{X}_i \gggtr \beta \;\;, \;\; \forall \; i \in [n]
\end{equation}

That is, the value represented $\widetilde{X}_i$ is $\mathbf{X}_i$ right-shifted $\beta$ bit positions. Consequently, the fixed-point representation allows using fractional numbers on low-cost integer hardware. Regarding the required memory space, it needs $(n \times k)$ bits for the internal representation of the numbers plus the memory space needed to store the scaling parameter, i.e., $e$ bits. Note that when the group size is large, the associated cost and storage overhead for $\beta$ is marginal because it is shared across many numbers.

If $\beta$ has the flexibility to change depending on the real numbers to be represented, the representation is named \textit{dynamic fixed-point}~\cite{kang2017dynamic}. In this paper, we adopt the dynamic fixed-point format, depicted in Figure~\ref{fig:Dynamic_Fixed_Point_Format}, to represent the data-structures of deep neural networks (DNNs). Therefore, we use a configurable $\beta$ to increase the flexibility as in the floating-point format but at a lower associated hardware cost because it is shared by a group of numbers. Specifically, given $n$ numbers to be represented, i.e., the weights or the activations of a layer, $\beta$ is tuned with the aim of minimizing a predefined cost function as will be illustrated in Section~{\color{DarkBlue}\ref{sec:quantization_parameters_optimization}}.

\Figure[t!][width=0.4\textwidth, trim={0cm 0cm 0cm 0cm}]{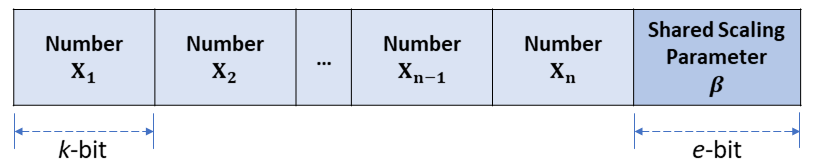}
{Generalized format for dynamic fixed-point numbers.
\label{fig:Dynamic_Fixed_Point_Format}}

More precisely, we employ the unsigned dynamic fixed-point format, referred to as $U\!\left(\alpha, \beta\right)$, to represent DNN data-structures that consist of only positive rationals. In contrast, the signed two's complement dynamic fixed-point format, denoted as $S\!\left(\alpha, \beta\right)$, is used for representing signed DNN data-structures. Here, we must emphasize that $\alpha$ includes the sign bit in the $S\!\left(\alpha, \beta\right)$ format. It is worth noting that the value of $\beta$ is not limited to the range [$0$, $k$] in dynamic fixed-point representation. This is very crucial for representing a group of large magnitude real numbers, or on the other hand, a group of small proper fractions $(\pm)$ with just a few bits. For a better representation in such cases, the representation may end up with a negative configuration for $\alpha$ or $\beta$.

For a given dynamic fixed-point format, the \textit{dynamic range} denotes the span of numbers that the format type can represent. Accordingly, a $U\!\left(\alpha, \beta\right)$ format has a dynamic range of ${[0, 2^\alpha - 2^{-\beta}]}$, whereas the dynamic range of $S\!\left(\alpha, \beta\right)$ format is ${[-2^{\alpha - 1}, 2^{\alpha - 1} - 2^{-\beta}]}$. Furthermore, the \textit{resolution} indicates the difference between two consecutive values in the dynamic range, which is equal to $2^{-\beta}$. For example, a $U\!\left(3, -1\right)$ format has a wordlength of $2$ bits for representing the integer numbers $\mathbf{X}_i$, a resolution of $2$, and $\widetilde{X}_i \in \{0, 2, 4, 6\}$. On the other hand, when $\mathbf{X}$ is in $S(-1, 3)$ format, then, $\mathbf{X}_i$ has a wordlength of $2$ bits, a resolution of $0.125$, and $\widetilde{X}_i \in \{-0.25, -0.125, 0, 0.125\}$.

\subsection{Quantization Approaches} \label{sec:q_methods}

In this section, we briefly discuss the approaches used to map $32$-bit floating-point (FP32) values to lower bit-precision and the benefits gained from each. Practitioners mainly use three types of quantization, namely, codebook-based, uniform, and non-uniform quantization approaches. Notwithstanding their differences, all methods aim at having compressed DNN data-structures to improve the memory efficiency and speed up the inference of neural networks.

In codebook-based quantization, a FP32 value $x$ is mapped to a $k$-bit codebook index $i$ using a pre-created codebook ${C = \left\{(i, v_i) : \forall \: i \in [0, 2^k), v_i \in \mathbb{R}\right\}}$ such that $v_i$ is the closest value to $x$~\cite{park2017weighted}. The idea behind this quantization method is to find a smallest possible optimal codebook for each data-structure. To perform a computation on the quantized data-structure, the codebook is used to retrieve the corresponding FP32 values. Accordingly, the computations in such a quantized DNN are floating-point arithmetic. However, computational and power efficiency increases through low bandwidth overhead and reduced memory requirements. To reduce the hardware cost, a simplified version uses integer codebook values with a single FP32 scalar~\cite{faraone2018syq}. 

On the other hand, the uniform quantization adopts an equal step size between the consecutive values. The linear~\cite{zhou2016dorefa} and fixed-point~\cite{lee2018quantization} quantization methods are examples of uniform quantization. These methods map a FP32 value $x$ to a $k$-bit integer value $\mathbf{x}$ using a scalar $\psi$ such that ${\mathbf{x} = \round*{x / \psi}}$. Here, $x / \psi$ is rounded to the nearest integer in a representation domain ${C \subseteq \left\{0, \pm 1, \dots, \pm 2^k \right\}}$, where $C$ is defined by the format type and the wordlength. Basically, the scalar of the linear quantization method depends on the considered range for the FP32 values. For instance, if the values are clipped/transformed into the range $[-\mu, \mu]$, then the scalar can be defined as $\mu/(2^{k - 1} - 1)$, where $\mu$ denotes the clipping/transformation parameter. On the other hand, the scalar can be equal to $\mu/(2^k - 1)$ when the considered range is $[0, \mu]$. Unlike linear quantization where the scalar is a FP32 value, the scalar in fixed-point quantization is a rational number, i.e., ${\psi = 1/2^{\beta}}$, $\beta \in \mathbb{Z}$. Thus, the idea behind the uniform quantization method is to find the optimal representation domain and scalar.

Accordingly, to perform the multiply-and-accumulate (MAC) operation between the FP32 vectors $X$ and $Y$ in a lower bit-precision, the input vectors are uniformly quantized based on their predefined scalars and representation domains. Here, we refer to the scalars of $X$ and $Y$ as $\psi_{\raisebox{-1.0pt}{$\scriptstyle x$}}$ and $\psi_{\raisebox{-1.0pt}{$\scriptstyle y$}}$, respectively. Thereafter, the MAC operation is performed on the lower bit-precision representations $\mathbf{X}$ and $\mathbf{Y}$ with integer arithmetic. Then, the result is rescaled to produce the output value $z$, such that

\begin{equation} \label{eq:linear_quantization}
 z = (\psi_{\raisebox{-1.0pt}{$\scriptstyle x$}} \cdot \: \psi_{\raisebox{-1.0pt}{$\scriptstyle y$}}) \cdot \text{MAC}\big(\mathbf{X}, \mathbf{Y}\big)
\end{equation} 

It should be noted that quantization and rescaling are done using floating-point arithmetic in linear quantization, whereas they are just simply bit-shift operations in fixed-point quantization. Finally, the non-uniform quantization adopts unequal step size between the consecutive values. Specifically, it quantizes the FP32 values in their logarithmic representation~\cite{miyashita2016convolutional}. Thus, non-uniform quantization method eliminates the need for digital multipliers. However, this comes at the cost of being more complicated during the quantization process and sacrificing model's accuracy.

\section{Literature Review} \label{sec:literature_review}

Inspired by the effectiveness and efficiency of low-precision operations, there have been a good number of studies on designing $k$-bit deep neural networks (DNNs). These efforts can roughly fall into two categories, quantization-aware training and post-training quantization. In the former approach, quantization is performed during network training so as to compensate for the effect of quantization. The latter approach, on the other hand, takes a pre-trained model and applies a static quantization formula with the intent of minimizing quantization error and maintaining network accuracy. Due to the emerging use of different bit-precision to represent DNN data-structures in different layers, a subsection is also devoted to mixed-precision quantization.

\subsection{Quantization-Aware Training} \label{sec:quantization_aware_training}

In DoReFa-Net~\cite{zhou2016dorefa}, activations are clipped with a bounded activation function and weights are transformed by the tanh function before being linearly quantized in the range $[0, 1]$. A much simpler weight quantization scheme is used in WRPN~\cite{mishra2017wrpn} through clipping the weights to the range $[-1, 1]$. On the other hand, PACT~\cite{choi2018pact} indicates that homogeneous clipping of all activations to $1$ is not effective for rigorous quantization. Instead, PACT learns the optimal clipping value for each layer during network training. Later in~\cite{choi2018bridging}, the idea of parametrized clipping extended to weights as well.

HWGQ-Net~\cite{cai2017deep} and TSQ~\cite{wang2018two} obtain the scalar for quantizing activations via Lloyd’s algorithm~\cite{lloyd1982least} on samples from the Gaussian distribution. Additionally, TSQ prunes activations below a handcrafted threshold to $0$. Then, it learns the quantized weights along with their kernel-wise scalars that minimize the mean square error (MSE) between full-precision (FP32) and quantized activations. QIL~\cite{jung2019learning}, on the other hand, learns both the clipping interval and the transformation parameter for weights and activations of each layer so as to quantize them linearly in the range $[0, 1]$.

LogQuant~\cite{miyashita2016convolutional} quantizes the FP32 values in their logarithmic representation based on manually fine-tuned layer-wise parameters, which is not a trivial process. Similarly, LPBN~\cite{graham2017low} quantizes activations in their log-space during the batch normalization (${\textnormal{\textit{BN}}}$) process using a manually defined formula for each wordlength. To simplify the log-based quantization, Hubara et al.~\cite{hubara2017quantized} quantizes the FP32 values to the index of the most significant bit in their binary representation. In WEQ~\cite{park2017weighted}, the authors automate the search for LogQuant parameters that maximize the weighted entropy of the quantized activations during training. On the other hand, they use an iterative search algorithm to find the codebook for the weights of each layer with the highest overall weighted entropy.

BalancedQ~\cite{zhou2017balanced} enforces frequency balance for the FP32 values mapped to each quantization bin. INQ~\cite{zhou2017incremental} quantizes the weights of a pre-trained model incrementally. On each iteration, a portion of the full-precision weights are quantized to fixed-point numbers, while the others are fine-tuned to compensate for the drop in accuracy. Apprentice~\cite{mishra2018apprentice} employs the knowledge distillation~\cite{hinton2015distilling} to improve the accuracy of the quantized network through the teacher network, which usually has a deeper architecture and a larger number of parameters and is trained in full-precision.

SYQ~\cite{faraone2018syq} quantizes each weight's kernel to binary or ternary values using ${s_k}^2$ codebooks, each of which with a learned FP32 scalar, where ${s_k}$ is the kernel size. With regards to activations, it uses $8$-bit fixed-point numbers where the fractional length is adjusted while training the network. In LQ-Nets~\cite{zhang2018lq}, the authors replace the standard basis of a $k$-bit quantized value with a basis vector consists of $k$ floating-point scalars. During the training phase, LQ-Nets learns the layer-wise basis vector for $k_a$-bit activations and the channel-wise basis vector for $k_w$-bit weights that minimizes the quantization error. Thus, unlike most of the techniques discussed where floating-point arithmetic is used once per multiply-and-accumulate (MAC) operation, SYQ and LQ-Nets use ${s_k}^2$ and ${(k_a \times k_w)}$ floating-point multiplication per MAC operation, respectively.

ALT~\cite{jain2019trained}, LSQ~\cite{esser2020learned}, and PROFIT~\cite{park2020profit} defined optimized gradient formulae. Accordingly, they learn low-precision parameters and quantization scalar of each layer when quantizing FP32 models to desired precision. While PROFIT and IBM's LSQ use FP32 quantization scalars, ALT restricts them to power-of-2. Furthermore, PROFIT proposed a training method in which sensitive weights, that cause activation instability, are progressively freezed.

\subsection{Post-Training Quantization}

In TWN~\cite{li2016ternary}, the authors use a layer-wise non-negative threshold ($\Delta$) and a scalar ($\alpha$) to design ternary weight networks. More precisely, TWN maps weights greater than $\Delta$ into $\alpha$ and weights less than $\Delta$ into $-\alpha$, while other wights are mapped into $0$. To increase TWN model capacity, FGQ~\cite{mellempudi2017ternary} partitions FP32 weights into groups along the input channels, and then ternarizes each group independently as in TWN but using separate thresholds for positive and negative values. The brute-force method is used in both approaches to find the quantization thresholds and scalars.

Nvidia’s TensorRT~\cite{migacz20178} adopts the linear quantization by using a clipping parameter that minimizes the Kullback-Leibler divergence between FP32 and quantized values. In~\cite{lee2018quantization}, Samsung proposed a channel-wise quantizer for $8$-bit fixed-point DNNs. The fractional length is estimated from the $n$-th moments of the channel-level distribution. In~\cite{jain2018compensated}, IBM introduced a new format, referred to as FPEC, for representing DNN data-structures. The FPEC includes two or more bits, denoted as compensation bits, which are used to estimate and compensate for the error due to fixed-point convolution. An accuracy-driven algorithm is proposed to reduce FPEC wordlength and adjust compensation bits.

In~\cite{kravchik2019low}, Huawei indicated that using linear quantization for data-structures with a high MSE may not be optimal. Therefore, a handcrafted threshold ($\tau$) is employed so that data-structures with an MSE larger than $\tau$ are quantized by codebook-based quantization, while others are quantized by linear quantization with the maximum absolute value as a clipping parameter. Then, quantization scalars of each layer are refined to minimize the MSE of the resulting activations of that layer. Meller et al.~\cite{meller2019same} at Hailo Technologies Ltd employ the weight factorization to make DNNs more robust to layer-wise post-training quantization. More precisely, for each weight layer, the output channels are scaled to match their dominant channel, that is, the channel with the largest absolute value, and thus equalizing channels.

KDE-KM~\cite{seo2019efficient} creates a quantization codebook for the weights of each layer from the center values of the clusters identified through the $k$-means clustering algorithm. Huang et al.~\cite{huang2021mxqn} proposed a mixed quantization framework, denoted as MXQN, for quantizing activations either to fixed-point numbers or quantizing them in their log-space based on the signal-to-quantization-noise ratio. On the other hand, the MXQN quantizes weights to fixed-point values.

To mitigate the significant accuracy drop due to ${\textnormal{base-}2}$ logarithmic quantization of weights, SegLog~\cite{xu2020memory} quantizes large magnitude weights in their base-$\sqrt{2}$ log-space. Specifically, SegLog creates a codebook with $N$ base-$2$ codes and ($2^{(k-1)} - N$) base-$\sqrt{2}$ codes for each layer, where $N$ is manually tuned to achieve the best accuracy. Additionally, SegLog approximates $\sqrt{2}$ to an expression of ${\textnormal{power-of-}2}$ integer values. Thus, it implements the multiplication between full-precision activations and quantized weights using shifters and adders.

\subsection{Mixed-Precision Quantization}

DNAS~\cite{wu2018mixed}, proposed by Facebook and Berkeley AI Research, represents the network as a computational directed acyclic graph (DAG). The nodes of the DAG represent activations while edges represent a parameterized convolution operation on the input node and its weights. Cascading nodes are connected using several edges each of which with a manually configured wordlengths. The SGD is used to select the edges, based on system requirements, as well as to optimize the weights of these edges. Note that DNAS follows PACT to quantize weights and activations. 

In ACIQ~\cite{banner2019post}, Intel proposed to allocate each data-structure channel with a different wordlength, whereby the average data-structure wordlength is maintained. Thereafter, with the use of a channel-wise clipping parameter, the range is divided into equal sized regions, and the FP32 values within each region are mapped to the midpoint. The optimal wordlength allocation and clipping thresholds are determined analytically to minimize the channel-wise MSE.

ADMM~\cite{ye2018unified} performs an iterative optimization to prune and quantize the weights in a layer-wise manner. The bit-precision of each layer is determined manually and the quantization scalars are defined so as to minimize the overall MSE. On each iteration, after pruning the weights, the ADMM quantizes a portion of FP32 weights that are closest to their discrete bins, and remaining weights are fine-tuned.

ReLeQ~\cite{elthakeb2019releq} and HAQ~\cite{wang2020hardware} adopt reinforcement learning to learn the wordlength of each data-structure in a layer-wise manner. Specifically, the ReLeQ uses the predicted bit-precision level to quantize the weights as in WRPN, whereas the HAQ quantizes both weights and activations on each explored wordlength in the same way as TensorRT. Subsequently, they retrain the quantized model to recover the performance, and then move on to reduce the wordlength of the next layer until the predefined constraints are met.

Rusci et al.~\cite{rusci2019memory} proposed to gradually reduce the bit-precision of weights and activations of convolution operation to one of the wordlengths in the set $\{8, 4, 2\}$ during the forward and backward passes. PACT quantization strategy is adopted for both weights and activations. To achieve the integer-only deployment, the authors combined the FP32 quantization scalars for activations and weights with ${\textnormal{\textit{BN}}}$ layer parameters, and then they converted the resulting parameters to $32$-bit fixed-point values.

HAWQ~\cite{dong2019hawq} defines the quantization sensitivity of each layer/block as the ratio between the top Hessian eigenvalue and the number of parameters. Then, it uses the measured sensitivity to indicate layer's prioritization to bit-precision reduction. Thus, HAWQ investigates the bit-precision space to determine the wordlength of each layer so that the assigned bit-precision ratio corresponds somewhat to the measured sensitivity of all layers. After quantization, HAWQ retrains the layers, one after another, in decreasing order of their quantization error and Hessian eigenvalue.

CCQ~\cite{khan2020learning} performs stages of competition and collaboration to gradually adapt weight's wordlength. The competition stage is carried out to measure the effect of quantizing randomly chosen layers to next bit-precision level on accuracy and memory. Thereafter, the CCQ randomly picks a layer based on the competition measurements and quantizes its weights to next level. On the other hand, the collaboration stage retrains all layers until the accuracy reaches a predefined threshold.

Chu et al.~\cite{chu2021mixed} heuristically assign the wordlength of activations and weights of each layer based on the separability of their hierarchical distribution. However, for large datasets such as ImageNet, it is unaffordable to obtain a complete separability matrix. Therefore, they used a random subset of the samples for this purpose. To reach the desired quantized network, the proposed framework utilized DoReFa-Net quantization technique with progressively decreasing wordlength during network training.

DoubleQExt~\cite{see2021doubleqext} quantizes weights and activations to $8$-bit integers using layer-wise FP32 scalar and offset parameters. Thereafter, it quantizes the integer weights again to represent them in power-of-$2$ form using $5$ bits, thus, reducing computational and memory cost. To recover the accuracy after the double quantization process, DoubleQExt iteratively reverts the most informative weights to their $8$-bit representation until the recovered accuracy reaches a predefined recovery threshold.

As a summary of the literature, improving the accuracy of quantized DNNs comes at the expense of floating-point computational cost in~\cite{zhou2016dorefa, li2016ternary, mishra2017wrpn, mellempudi2017ternary, migacz20178, cai2017deep, hubara2016binarized, rastegari2016xnor, zhu2016trained, choi2018pact, choi2018bridging, wang2018two, jung2019learning, kravchik2019low, esser2020learned, faraone2018syq, mishra2018apprentice, huang2021mxqn, wu2018mixed, ye2018unified, elthakeb2019releq, wang2020hardware, dong2019hawq, seo2019efficient, khan2020learning, chu2021mixed, see2021doubleqext, zhang2018lq}. Specifically, these approaches scale output activations of each layer with FP32 coefficient(s) to recover the dynamic range, and/or perform batch normalization as well as the operations of first and last layers with FP32 data-structures. Thus, the FPGA/ASIC-based hardware accelerators of such quantized DNNs still need to implement expensive floating-point units for multiplication and accumulation, and even the much more expensive non-linear quantization arithmetics in~\cite{miyashita2016convolutional, graham2017low, park2017weighted}. To overcome this issue, this paper adopts a simpler and a more realistic hardware-aware fixed-point quantization for integer-only deployment of quantized DNNs.

Another notable drawback of the discussed techniques in~\cite{zhou2017incremental, park2020profit, elthakeb2019releq, wang2020hardware, rusci2019memory, dong2019hawq, khan2020learning, chu2021mixed} is that they usually perform training repeatedly, which is highly inefficient and takes a lot of time to construct the quantized model~\cite{han2015deep}. Furthermore, training requires a full-size dataset, which is often unavailable in real-world scenarios for reasons such as proprietary and privacy, especially in the case when working on an off-the-shelf pre-trained model from a community or industry for which data is no longer accessible. Additionally, training involves hyper-parameters configuration, initialization, and training methods, which are not always feasible. In this paper, we tackle these drawbacks by avoiding training and quantizing pre-trained models.

At the same time, we achieve accuracy at par with the full-precision model, a major challenge in the discussed post-training quantization approaches, through an optimized quantization method with mixed-precision assignment. In contrast to methods that determine quantization parameters based on a local optimization of a metric, such as minimizing the KL divergence in~\cite{migacz20178} and MSE in~\cite{kravchik2019low, banner2019post} of each layer/channel, this paper involves the data-structures of all layers affected by quantization as well as the network outcome in this process. Thus, it insures a global optimization of quantization parameters.

To overcome the computationally expensive exploration and sorting phases during wordlength reduction in the discussed mixed-precision quantization approaches, this paper adopts a simple computation-and-memory based $2$-level clustering and traversal of data-structures. Additionally, as opposed to the mixed-precision design in~\cite{rusci2019memory}, this work considers the sensitivity of layers to quantization when decreasing their bit-precision level. This is because we experimentally found it to significantly affect the accuracy of the model. Last, but not least, this paper improves the computational and power efficiency of the channel-wise mixed-precision design in~\cite{banner2019post , rusci2019memory} through adopting layer-wise quantization and allocating less bits, $16$ bits or less, to ${\textnormal{\textit{BN}}}$ parameters and partial results.

\section{Designing Mixed Low-Precision Deep Neural Networks for Integer-Only Deployment} \label{sec:functional_components}

In this section, we provide an insight into our \textbf{F}i\textbf{x}ed-\textbf{P}oint \textbf{Q}uantizer of deep neural \textbf{Net}works (${\textnormal{\textbf{FxP-QNet}}}$). The ${\textnormal{\textbf{FxP-QNet}}}$ main components are the \textit{Preprocessor}, the \textit{Forward Optimizer}, and the \textit{Network Designer} as shown in Figure~\ref{fig:FxP_QNets_components}. The \textit{Preprocessor} imposes integer-only inference where floating-point data-structures are quantized to integers in dynamic fixed-point representation using the quantization method that will be illustrated in Section~{\color{DarkBlue}\ref{sec:dynamic_fixed_point_quantization}}. The \textit{Forward Optimizer} employs the optimization framework that will be presented in Section~{\color{DarkBlue}\ref{sec:quantization_parameters_optimization}} to improve the quality of the initially designed low-precision DNN. Finally, the \textit{Network Designer} finds the best quantization level for each data-structure to design a mixed low-precision DNN as will be discussed later in Section~{\color{DarkBlue}\ref{sec:bit_precision_level_reduction}}.

Here, we must emphasize that the dynamic fixed-point quantization is adopted in this paper because; \begin{enumerate*}[label=(\roman*)]
\item it allows using fractional numbers on low-cost integer hardware, \item it is more hardware-friendly than the non-uniform quantization, \item it facilitates the implementation of highly resource/energy-efficient FPGA/ASIC-based hardware accelerators, and \item its arithmetic is naturally supported in the instruction-set of all general-purpose programmable microcontroller units in edge devices.\end{enumerate*} Details of the ${\textnormal{\textbf{FxP-QNet}}}$ components and workflow are presented next.

\Figure[t!][width=0.375\textwidth, trim={0cm 0.5cm 0cm 2cm}]{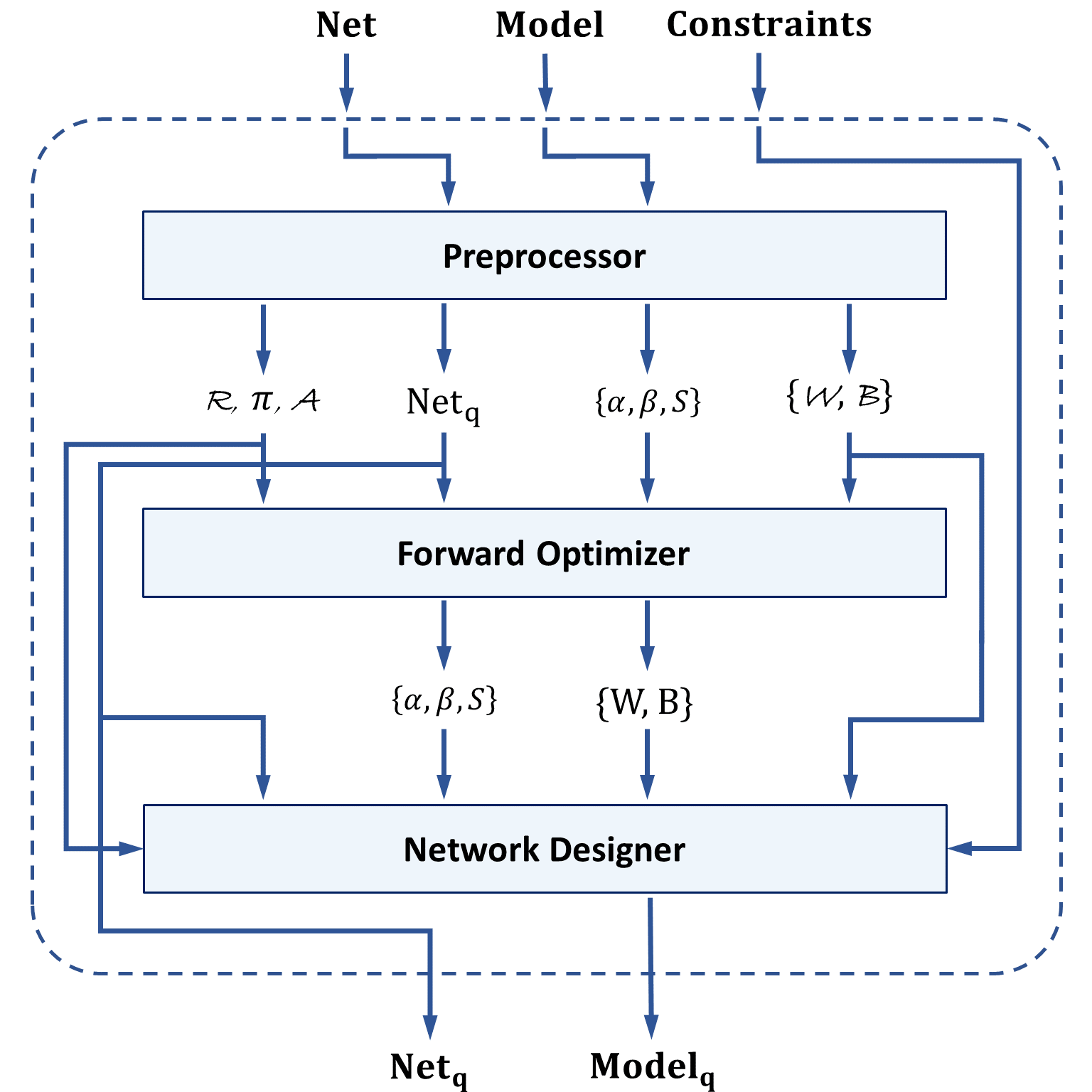}
{Components and workflow of ${\textnormal{\textbf{FxP-QNet}}}$. The ${\textnormal{\textbf{FxP-QNet}}}$ takes a pre-trained model with the description of network structure and the design constraints. Then, it designs a network structure for integer-only deployment and quantizes model parameters at different bit-precision levels.
\label{fig:FxP_QNets_components}}\vspace{-2mm}

\subsection{Integer-Only Inference} \label{sec:integer_only_inference}

In machine learning frameworks, the learnable parameters, e.g., weights (${w}$) and biases (${b}$), as well as the features, i.e., activations (${a}$), of a DNN consisting of $L$ layers are represented with a single-precision floating-point (FP32). Since not all DNN layers may have learnable parameters, we define the set $\mathfrak{L_p}$ as the layers with learnable parameters, $\mathfrak{L_p} \subseteq [L]$. Formally, we represent the set of all layers’ activations, weights, and biases as ${\mathcal{A} = \left\{\mathcal{A}_\ell : \ell \in [L]\right\}}$, ${\mathcal{W} = \left\{\mathcal{W}_\ell : \ell \in \mathfrak{L_p}\right\}}$, and ${\mathcal{B} = \left\{\mathcal{B}_\ell : \ell \in \mathfrak{L_p}\right\}}$, respectively. 

To avert using floating-point arithmetic, the computations of convolution (${\textnormal{\textit{CONV}}}$), fully-connected (${\textnormal{\textit{FC}}}$), batch normalization (${\textnormal{\textit{BN}}}$), pooling (${\textnormal{\textit{POOL}}}$), rectified linear unit (${\textnormal{\textit{ReLU}}}$), ${\textnormal{\textit{Scale}}}$, and element-wise (${\textnormal{\textit{Eltwise}}}$) layers must be performed on integers. For this purpose, ${\textnormal{\textbf{FxP-QNet}}}$ quantizes the FP32 data-structures required for these operations to integer data-structures either in unsigned dynamic fixed-point representation, $U\!\left(\alpha, \beta\right)$, or in signed 2's complement dynamic fixed-point representation, $S\!\left(\alpha, \beta\right)$. In the last subsection, we show how \textit{preprocessor} specifies the initial quantization parameters, i.e., integer length $\alpha$, fractional length $\beta$, and sign strategy $\mathcal{S}$, for each data-structure.

While model parameters can be quantized offline just before the deployment, quantizing activations requires on-the-fly shifting and clipping within the computational path of the network structure during the forward propagation. To this end, we implemented a layer, referred to as a ${\textnormal{\textit{FxP-QLayer}}}$, to carry out activation quantization. In Section~{\color{DarkBlue}\ref{sec:FxP_QLayer}}, we discuss where the ${\textnormal{\textit{FxP-QLayers}}}$ are inserted into the computation data path of the network architecture to ensure that the inference is performed using low-precision integer activations. It is noteworthy that ${\textnormal{\textbf{FxP-QNet}}}$ adopts the layer-wise quantization strategy for both model parameters and activations.
On the other hand, the ${\textnormal{\textit{BN}}}$ formulation hinders integer-only deployment because of its need for floating-point arithmetic. The next subsection explains how \textit{preprocessor} elevates this issue.

\subsubsection{Batch Normalization Refinement}

DNNs are designed with ${\textnormal{\textit{BN}}}$ layers to accelerate the training phase~\cite{ioffe2015batch}. ${\textnormal{\textit{BN}}}$ layer performs two operations, namely, normalizing the inputs to have zero mean and unit variance distributions, and then scaling the normalized inputs to any other distribution through a linear transform function. During network training time, each ${\textnormal{\textit{BN}}}$ layer learns its scale $\gamma$ and bias shift $\delta$ parameters. Additionally, it computes the global mean $\mu$ and variance ${\sigma^2}$ through a running average of the statistics of each mini-batch of training. During network forward pass, ${\textnormal{\textit{BN}}}$ transform is applied to the input activations $\mathcal{A}_{\ell - 1}$ to produce the output activations $\mathcal{A}_\ell$ as

\begin{equation}
 \mathcal{A}_\ell = \mathrm{BN}(\mathcal{A}_{\ell - 1}; \gamma_\ell, \delta_\ell, \mu_\ell, \sigma_\ell^2) = \gamma_\ell \cdot \frac{\mathcal{A}_{\ell - 1} - \mu_\ell}{\sqrt{\sigma_\ell^2 + \epsilon}} + \delta_\ell
\end{equation}
\\
where $\gamma_\ell$, $\delta_\ell$, $\mu_\ell$, and $\sigma_\ell^2$ are channel-wise FP32 parameters learned for ${\textnormal{\textit{BN}}}$ layer $\ell$, and $\epsilon$ is a small constant used for numerical stability. It is worth noting that this kind of formulation hinders integer-only deployment as it contains floating-point arithmetic. In~\cite{jacob2018quantization}, the authors fold ${\textnormal{\textit{BN}}}$ parameters into ${\textnormal{\textit{CONV}}}$ weights before quantization. However, the folding process itself deteriorated the accuracy. In this paper, we simplify and rewrite the ${\textnormal{\textit{BN}}}$ formula as

\begin{equation} \label{eq:BatchNorm}
 \mathrm{BN}(\mathcal{A}_{\ell - 1}; \gamma_\ell, \delta_\ell, \mu_\ell, \sigma_\ell^2) = \aleph_{1_\ell} \cdot \mathcal{A}_{\ell - 1} + \aleph_{2_\ell}
\end{equation}
\\
where $\aleph_{1_\ell}$ and $\aleph_{2_\ell}$ are new channel-wise FP32 parameters defined as

\begin{equation} \label{eq:BatchNorm_s1_s2}
\begin{split}
 \aleph_{1_\ell} = & \;\frac{\gamma_\ell}{\sqrt{\sigma_\ell^2 + \epsilon}}\\
 \aleph_{2_\ell} = & \;\delta_\ell - \frac{\gamma_\ell \cdot \mu_\ell}{\sqrt{\sigma_\ell^2 + \epsilon}}
\end{split}
\end{equation}

Accordingly, \textit{preprocessor} modifies the network structure by converting ${\textnormal{\textit{BN}}}$ layers into much simpler ${\textnormal{\textit{Scale}}}$ layers, updates the model parameters by assigning to each of these ${\textnormal{\textit{Scale}}}$ layers its scale $\aleph_{1_\ell}$ and bias shift $\aleph_{2_\ell}$ parameters, and finally removes the parameters related to ${\textnormal{\textit{BN}}}$ layers. At this point, performing integer-only inference becomes feasible by quantizing the parameters of ${\textnormal{\textit{Scale}}}$ layers in the same way as quantizing weights and biases of ${\textnormal{\textit{CONV}}}$/${\textnormal{\textit{FC}}}$ layers. Note that, for simplicity, we refer to $\aleph_{1_\ell}$ and $\aleph_{2_\ell}$ as $\mathcal{W}_\ell$ and $\mathcal{B}_\ell$, respectively, during the design of low-precision DNNs.

\subsubsection{FxP-QLayer: An On-the-Fly Quantization of Activations } \label{sec:FxP_QLayer}

The ${\textnormal{\textit{FxP-QLayer}}}$ is responsible for changing the wordlength of input activations to a lower bit-precision level. This is done through a simple on-the-fly shift and add operations without using expensive multipliers, as will be discussed in Section~{\color{DarkBlue}\ref{sec:dynamic_fixed_point_quantization}}. Here, an important issue is where to inject ${\textnormal{\textit{FxP-QLayer}}}$ into the computation data path of the network architecture. To perform the first computation on low-precision integers, ${\textnormal{\textit{FxP-QLayer}}}$ must be inserted at the beginning of the network to quantize input features. In addition, output activations of ${\textnormal{\textit{CONV}}}$, ${\textnormal{\textit{FC}}}$, ${\textnormal{\textit{POOL}}}$, ${\textnormal{\textit{ReLU}}}$, ${\textnormal{\textit{Scale}}}$, and ${\textnormal{\textit{Eltwise}}}$ layers must be quantized before being fed to the next layer. However, for efficient inference of DNNs in hardware, some DNN layers can be fused with each other to form what we call a computational block (${\text{C-Block}}$).

For instance, ${\textnormal{\textit{ReLU}}}$ and ${\textnormal{\textit{Scale}}}$ layers can be fused with their previous layer since they perform feature-wise operations. Thus, ${\textnormal{\textit{FxP-QLayer}}}$ is also required after each ${\text{C-Block}}$ that has a structure such as "${\mathcal{X}-{\textnormal{\textit{Scale}}}\text{ or/and }{\textnormal{\textit{ReLU}}}}$" or even as simple as "$\mathcal{X}$", where $\mathcal{X}$ here can be ${\textnormal{\textit{CONV}}}$, ${\textnormal{\textit{FC}}}$, ${\textnormal{\textit{POOL}}}$, or ${\textnormal{\textit{Eltwise}}}$ layer. Note that unlike ${\textnormal{\textit{AVG-POOL}}}$ operation where a patch of activations is replaced by their average, ${\textnormal{\textit{MAX/MIN-POOL}}}$ operations replace a patch of activations with one of them, which is the max/min value. Accordingly, there is no need for a ${\textnormal{\textit{FxP-QLayer}}}$ after ${\textnormal{\textit{MAX/MIN-POOL}}}$ layer. On the other hand, for ${\textnormal{\textit{AVG-POOL}}}$ layer, the reciprocal of the patch size is handled in the same way as the model parameters.

Last but not least, activations produced by last network layer are used to predict the model outcome and no other calculations are made based on them. So that, there is no need to insert a ${\textnormal{\textit{FxP-QLayer}}}$ after the last network layer. The \textit{preprocessor} modifies the network structure by injecting ${\textnormal{\textit{FxP-QLayers}}}$ as discussed above. Thus, DNNs can be viewed as a graph with ${\textnormal{\textit{FxP-QLayers}}}$ as vertices and ${\text{C-Blocks}}$ as edges. Each ${\text{C-Block}}$ performs its layers operations in a pipeline fashion, and thereafter ${\textnormal{\textit{FxP-QLayer}}}$ carries out quantization to its ${\text{C-Block}}$ output activations.

\subsubsection{Initial Design Generation} \label{sec:preprocessor_initial_solution}

The \textit{preprocessor} adopts a statistics-driven approach to determine the initial fixed-point representation for a given FP32 data-structure $X_\ell^{\left(d\right)}$, $X_\ell^{\left(d\right)} \in \left\{\mathcal{A}_\ell, \mathcal{W}_\ell, \mathcal{B}_\ell\right\}$, and a target wordlength $k_\ell^{(d)}$. Here, $d$ is the data-structure type, $d \in \{a, w, b\}$, and $\ell$ is the layer ID, $\ell \in [L]$. Specifically, the smallest FP32 value in the data-structure is used to determine its sign strategy as follows

\begin{figure*}[t]
\captionsetup[subfloat]{captionskip=0.5pt}
\centering
\subfloat[AlexNet architecture.]{\includegraphics[width=0.30\textwidth, trim= 0.0cm 0.0cm 0.0cm 0.35cm, clip]{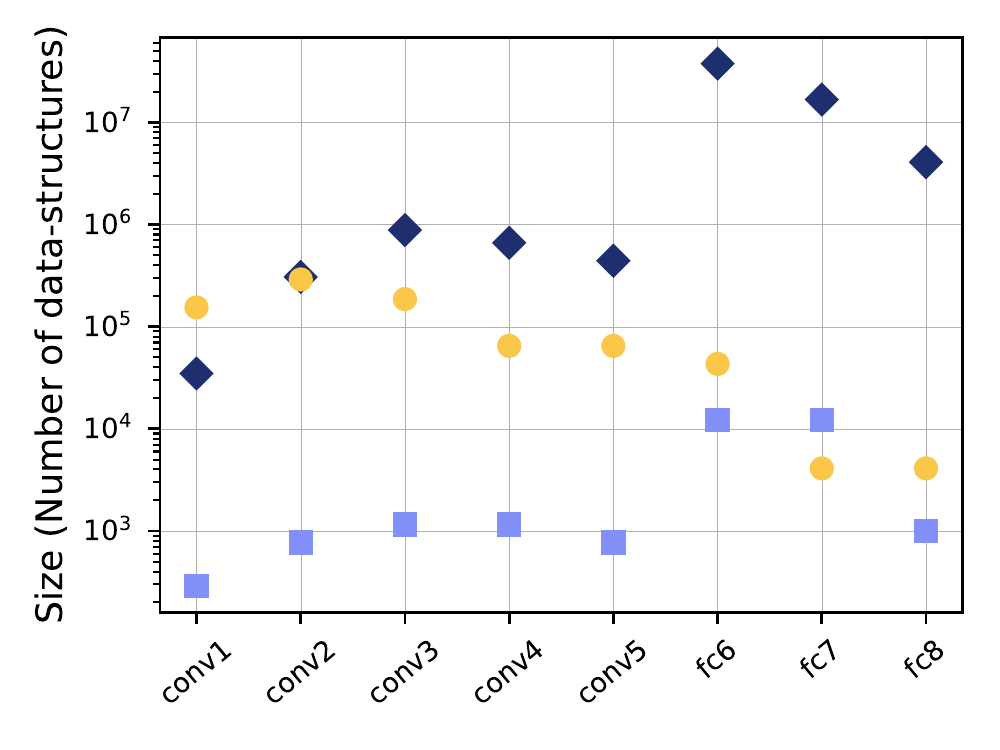}}
\hspace{2mm}
\subfloat[VGG-16 architecture.]{\includegraphics[width=0.60\textwidth, trim= 0.0cm 0.0cm 0.0cm 0.35cm, clip]{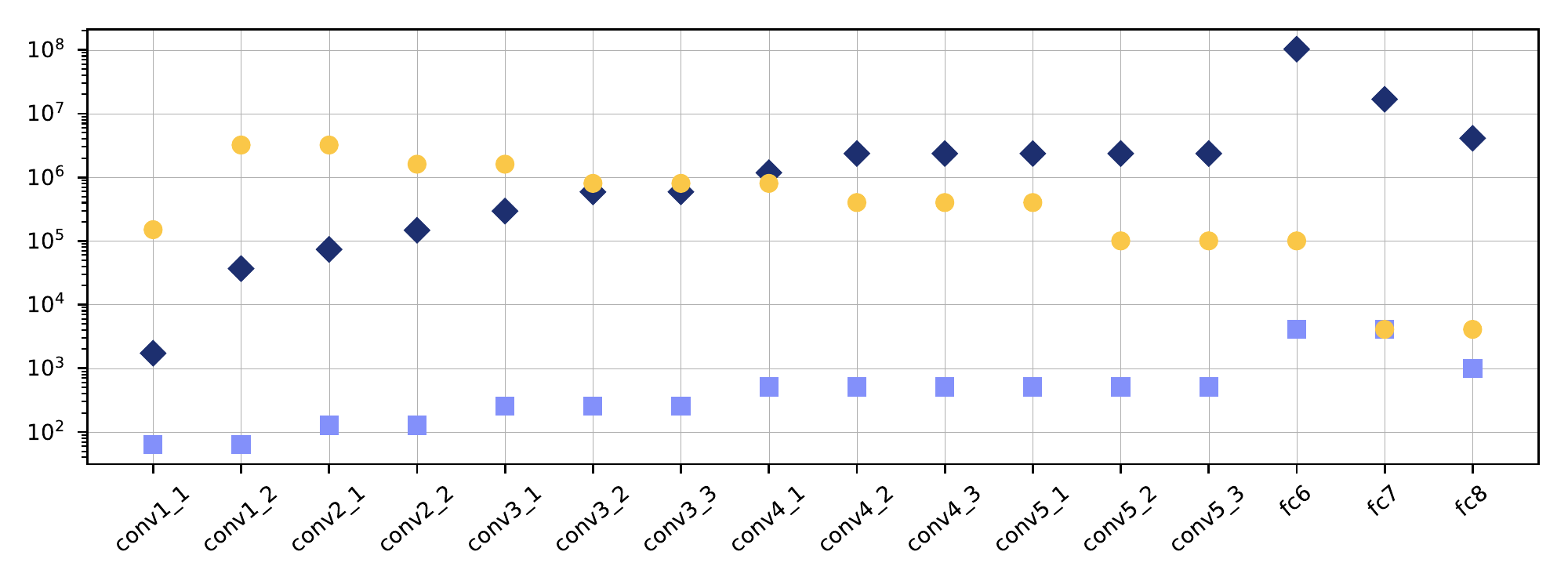}}
\hfil\\[-2ex]
\subfloat[ResNet-18 architecture.]{\includegraphics[width=0.79\linewidth,clip]{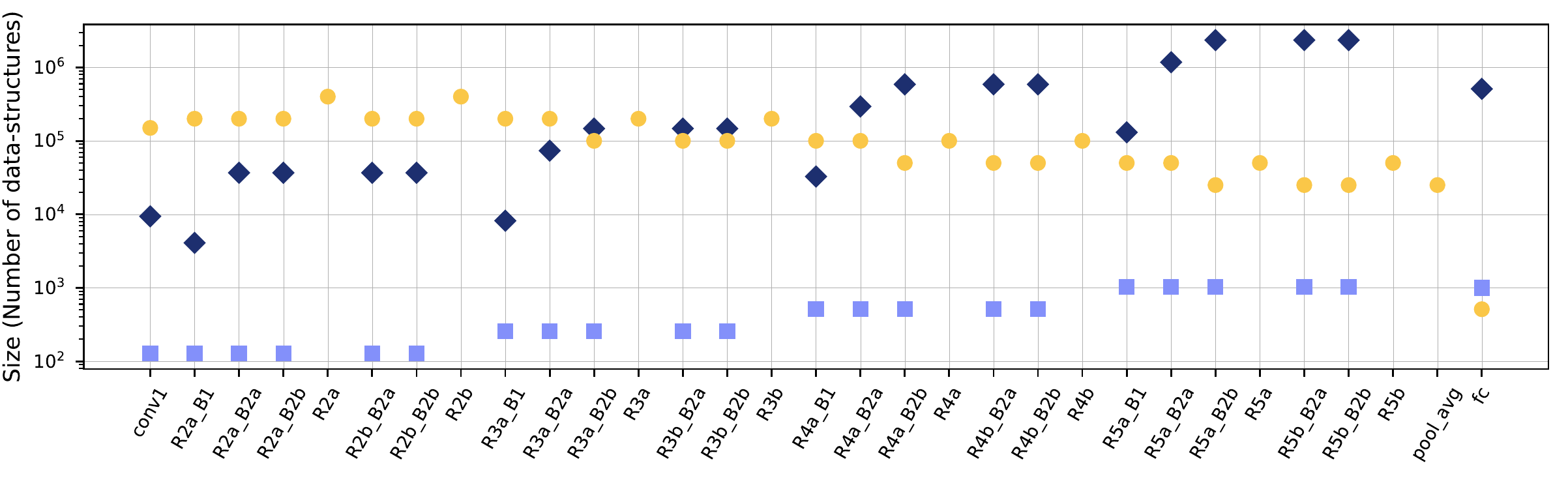}}
\subfloat{\includegraphics[width=0.11\linewidth, trim=16.95cm 0cm 0cm 0cm,clip]{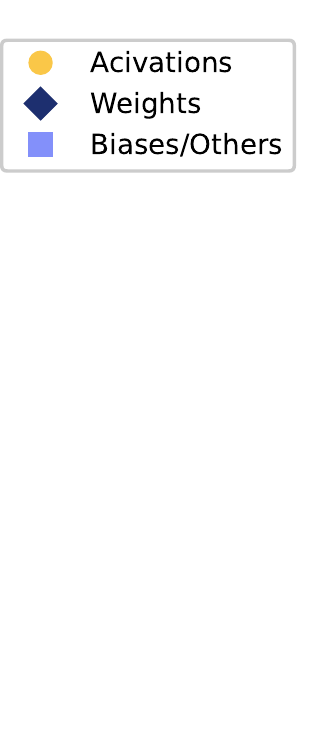}}
\caption[]{Statistics on size of input data-structures for each layer of \begin{enumerate*}[label=(\alph*)] \item AlexNet, \item VGG-16, and \item ResNet-18 \end{enumerate*} architectures in log-domain.}
\label{fig:data_structure_size}
\end{figure*}
\begin{equation}
 \mathcal{S}_\ell^{(d)} = \mathcal{I}\left({X_\ell^{\left(d\right)}}\strut_{\!\!\!\text{MAX}-} < 0\right)
\end{equation}
\\
where $\mathcal{I}\left(\cdot\right)$ is the indicator function; it returns $1$ if the condition is satisfied and $0$ otherwise. Thereafter, the \textit{preprocessor} analyzes the scalar $\psi$ based on the dynamic range, which is determined by wordlength and sign strategy as mentioned in Section~{\color{DarkBlue}\ref{sec:fixed_point_representation}}, such that

\begin{equation}
 \psi_\ell^{(d)} = 
 \begin{cases}
 \frac{{X_\ell^{\left(d\right)}}\strut_{\!\!\!\text{MAX}+}}{2^{k_\ell^{(d)}} - 1} \;\;\;\;\;\;\;\;\;\;\;\;\;\;\;\;\;\;\;\;\;\;\;\;\;\;\;\;\,, & \mathcal{S}_\ell^{(d)} = 0 \\
 \min\!\left(\frac{{X_\ell^{\left(d\right)}}\strut_{\!\!\!\text{MAX}-}}{-2^{k_\ell^{(d)} - 1}}, \frac{{X_\ell^{\left(d\right)}}\strut_{\!\!\!\text{MAX}+}}{2^{k_\ell^{(d)} - 1} - 1}\right), & \mathcal{S}_\ell^{(d)} \neq 0 
 \end{cases}
\end{equation}
\\
where ${X_\ell^{\left(d\right)}}\strut_{\!\!\!\text{MAX}+}$ and ${X_\ell^{\left(d\right)}}\strut_{\!\!\!\text{MAX}-}$ are the largest and smallest FP32 values in the data-structure $X_\ell^{\left(d\right)}$, respectively. Since the scalar of fixed-point representation equals ${2^{-\beta}}$, the FP32 scaling parameter can be calculated as 
\begin{equation}
 \widehat{\beta}_\ell^{(d)} = - \log_2 \psi_\ell^{(d)}
\end{equation}

Accordingly, the \textit{preprocessor} initializes the integer length $\alpha$ and fractional length $\beta$ based on $\widehat{\beta}_\ell^{(d)}$, $\mathcal{S}_\ell^{(d)}$, and a given rounding strategy $\mathfrak{Rs}$. Here, we adopt three rounding strategies, namely, \textit{conservative} ($\mathfrak{Rsc}$), \textit{neutral} ($\mathfrak{Rsn}$), and \textit{aggressive} ($\mathfrak{Rsa}$) strategies, such that

\begin{equation} \label{eq:initial_quantization_parameters}
\begin{split}
 & \beta_\ell^{(d)} = 
 \begin{cases}
 \big\lfloor\widehat{\beta}_\ell^{(d)}\big\rfloor, & \mathfrak{Rs} \textnormal{ \small\textbf{is} } \mathfrak{Rsc}\\
 \big\lfloor\widehat{\beta}_\ell^{(d)}\big\rceil, & \mathfrak{Rs} \textnormal{ \small\textbf{is} } \mathfrak{Rsn}\\
 \big\lceil\widehat{\beta}_\ell^{(d)}\big\rceil, & \mathfrak{Rs} \textnormal{ \small\textbf{is} } \mathfrak{Rsa}
 \end{cases}\\[4pt]
 & \alpha_\ell^{(d)} = k_\ell^{(d)} - \beta_\ell^{(d)} - \mathcal{S}_\ell^{(d)}
 \end{split}
\end{equation}

In Section~{\color{DarkBlue}\ref{sec:analysis_rounding_strategies}}, we analyze these strategies by demonstrating the performance of a DNN that was quantized using initial quantization parameters derived from Equation~(\ref{eq:initial_quantization_parameters}) at different homogeneous wordlengths. Here, an important issue is deciding which initial bit-precision level should be assigned to each data-structure. The experimental results for data-structure sensitivity to quantization, illustrated in Section~{\color{DarkBlue}\ref{sec:sensitivity_results}}, show that data-structures in shallower layers are less robust to quantization than the ones in deeper layers. On the other hand, output activations of the last network layer are used to predict the model outcome. Hence, the computation of first and last network layers should initially use accurate neurons, i.e., neurons with a high bit-precision level, in order to maintain the performance.

This fact that the first and the last layers should be treated differently has also been observed in various quantization techniques~\cite{zhou2016dorefa, mishra2017wrpn, cai2017deep, hubara2016binarized, rastegari2016xnor, zhu2016trained, choi2018pact, wang2018two, esser2020learned, chu2021mixed, faraone2018syq}, where quantization of these layers is avoided as it may have devastating effects on the overall accuracy. However, we found that quantizing these two layers can greatly increase the compression ratio which in turn significantly reduces the computational power required. Thus, the proposed framework starts the bit-precision level of activations and weights of first and last layers with $32$ bits, i.e., $k_0^{(a)} = k_1^{(w)} = k_{L-1}^{(a)} = k_L^{(w)} = 32$.

On the other hand, to achieve the accuracy-compression trade-off, the framework sets the wordlength of other activations and weights of ${\textnormal{\textit{CONV}}}$ and ${\textnormal{\textit{FC}}}$ layers to $8$ bits. This is because it has been shown that representing these data-structures with $8$-bit integers achieves comparable accuracy to a full-precision network without the need for fine-tuning~\cite{lee2018quantization}. Other data-structures, such as biases of ${\textnormal{\textit{CONV}}}$ and ${\textnormal{\textit{FC}}}$ layers as well as the parameters of ${\textnormal{\textit{Scale}}}$ layer, have a relatively small memory footprint in typical DNNs as shown in Figure~\ref{fig:data_structure_size}. Thus, the framework uses $32$ bits to initially represent them.

\subsection{Quantization Methodology} \label{sec:dynamic_fixed_point_quantization}

The quantization method, denoted as $Q(.; ., ., .)$, converts a FP32 data-structure $X_\ell^{\left(d\right)}$, $X_\ell^{\left(d\right)} \in \left\{\mathcal{A}_\ell, \mathcal{W}_\ell, \mathcal{B}_\ell\right\}$, to a low-precision integer data-structure $\mathbf{X}_\ell^{\left(d\right)}$, ${\mathbf{X}_\ell^{\left(d\right)} \in \left\{\mathbf{A}_\ell, \mathbf{W}_\ell, \mathbf{B}_\ell\right\}}$, either in unsigned dynamic fixed-point representation, $U\big(\alpha_\ell^{\left(d\right)}, \beta_\ell^{\left(d\right)}\big)$, or in signed 2's complement dynamic fixed-point representation, $S\big(\alpha_\ell^{\left(d\right)}, \beta_\ell^{\left(d\right)}\big)$, depending on the layer-wise quantization parameters $\alpha_\ell^{\left(d\right)}$, $\beta_\ell^{\left(d\right)}$, and $\mathcal{S}_\ell^{\left(d\right)}$, such that

\begin{equation} \label{eq:fp_k_bit_quantization}
\begin{split}
\!\mathbf{X}_\ell^{\left(d\right)} &
= {} Q\left(X_\ell^{\left(d\right)}; \alpha_\ell^{\left(d\right)}, \beta_\ell^{\left(d\right)}, \mathcal{S}_\ell^{\left(d\right)}\right) \\
& = {} \text{clip}\Big(\round*{X_\ell^{\left(d\right)} \cdot 2^{\beta_\ell^{\left(d\right)}}}; \mathbf{Q}_\text{min}, \mathbf{Q}_\text{max}\Big) \\
\end{split}
\end{equation}
\\
where $d$ denotes the data-structure type, $d \in \left\{a, w, b\right\}$, such that $\mathbf{X}^{\left(a\right)} \equiv \mathbf{A}$, $\alpha_\ell^{\left(d\right)}$ and $\beta_\ell^{\left(d\right)}$ refer to the integer length and fractional length, respectively, $\alpha_\ell^{\left(d\right)}$ and $\beta_\ell^{\left(d\right)} \in \mathbb{Z}$. Thus, each integer number in $\mathbf{X}_\ell^{\left(d\right)}$ is represented with $k_\ell^{\left(d\right)}$ bits, ${k_\ell^{\left(d\right)} = \alpha_\ell^{\left(d\right)} + \beta_\ell^{\left(d\right)}}$. Note that the values represented by $\mathbf{X}_\ell^{\left(d\right)}$ are calculated as mentioned in Equation (\ref{eq:fixed_point}) using the parameter $\beta_\ell^{\left(d\right)}$. On the other hand, the sign strategy $\mathcal{S}_\ell^{\left(d\right)}$ indicates whether the data-structure type $d$ at layer $\ell$ consists of unsigned ($0$) or signed ($1$) numbers, $\mathcal{S}_\ell^{\left(d\right)} \in \left\{0, 1\right\}$. The sign strategy is employed in order to avoid losing half of the dynamic range when, for instance, using signed number representation for non-negative data-structures. Additionally, the $\round*{\bullet}$ represents the round operation where the half is rounded to the nearest even to prevent bias~\cite{jacob2018quantization}, and

\begin{equation} \label{eq:clip}
 \text{clip}\!\left(\widehat{\mathbf{X}}; \text{min}, \text{max}\right) = \begin{cases}
 \text{min}, & \forall \, \widehat{\mathbf{X}}_i < \text{min} \\
 \widehat{\mathbf{X}}_i, & \text{min} \leq \forall \, \widehat{\mathbf{X}}_i \leq \text{max}\\ 
 \text{max}, & \forall \, \widehat{\mathbf{X}}_i > \text{max}
 \end{cases}
\end{equation}

Furthermore, the $\mathbf{Q}_\text{min}$ and $\mathbf{Q}_\text{max}$ represent the number of positive and negative quantization levels, respectively, and are computed as

\begin{equation} \label{eq:quantization_levels}
\begin{split}
 \mathbf{Q}_\text{min} & = {} -\mathcal{S}_\ell^{\left(d\right)} / \, 2^{\mathcal{S}_\ell^{\left(d\right)} - \alpha_\ell^{\left(d\right)} - \beta_\ell^{\left(d\right)}}\\
 \mathbf{Q}_\text{max} & = {} 2^{\alpha_\ell^{\left(d\right)} + \beta_\ell^{\left(d\right)} - \mathcal{S}_\ell^{\left(d\right)}} - 1
\end{split}
\end{equation}

\Figure[t!](topskip=0pt, botskip=0pt, midskip=0pt)[width=0.975\textwidth, trim={0cm 1cm 0cm 2cm}]{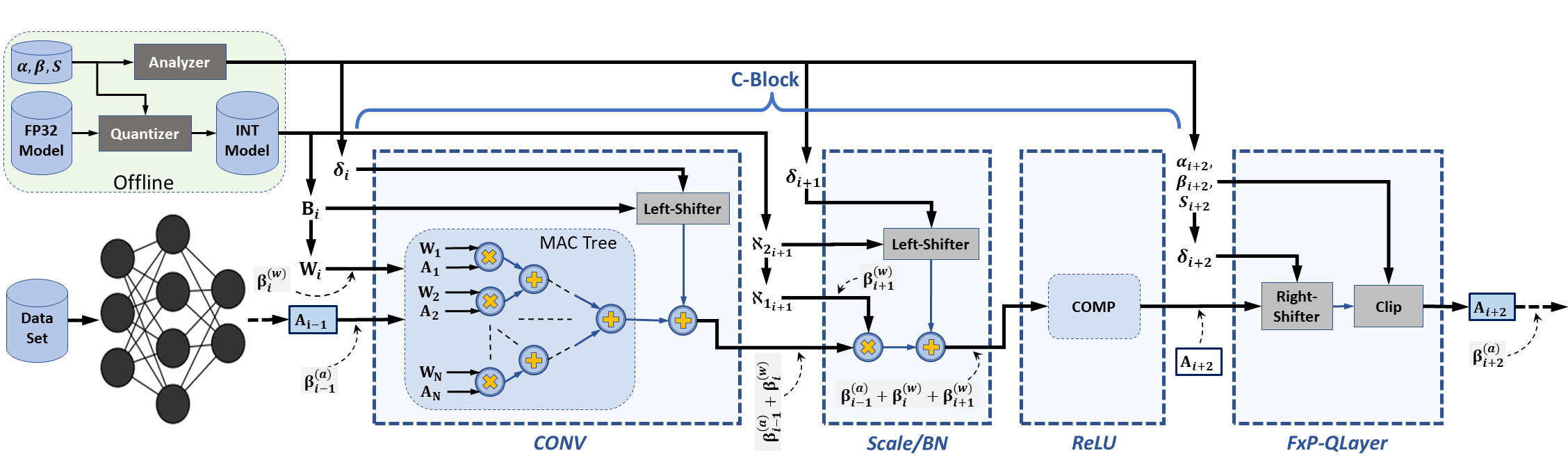}
{An overview of ${\textnormal{\textbf{FxP-QNet}}}$ low-precision calculations in a computational block consisting of ${\textnormal{\textit{CONV}}}$, ${\textnormal{\textit{Scale}}}$, and ${\textnormal{\textit{ReLU}}}$ layers, as envisioned here. Note that shift, wordlength, and sign strategy are layer-wise parameters consisting of a very small number of bits, and therefore their memory footprint is negligible.
\label{fig:FxP_QNets_Hardware}}

For the special case of quantizing a signed data-structure to a very low-precision level, i.e., $1$-bit, the quantization method utilizes the binary quantization~\cite{rastegari2016xnor}. More precisely, it quantizes the elements of the input $X_\ell^{\left(d\right)}$ based on their \textit{sign} as follows

\begin{equation} \label{eq:fp_1_bit_quantization}
\begin{split}
 \mathbf{X}_\ell^{\left(d\right)} & = {} Q\left(X_\ell^{\left(d\right)}; \alpha_\ell^{\left(d\right)}, \beta_\ell^{\left(d\right)}, \mathcal{S}_\ell^{\left(d\right)}\right)\\
 & = {} 
 \begin{cases}
 -1, & \forall \, \text{${X_i}_\ell^{\left(d\right)} \in (-\infty, 0)$} \\
 0, & \forall \, \text{${X_i}_\ell^{\left(d\right)} \in [0, \infty)$} 
 \end{cases}
\end{split}
\end{equation}

Then, before performing operations on $\mathbf{X}_\ell^{\left(d\right)}$, the values of $\mathbf{X}_\ell^{\left(d\right)}$ are decoded such that $0$s are mapped into $+1$. Thus, the values represented by $\mathbf{X}_\ell^{\left(d\right)}$ become $\pm 2^{-\beta_{\ell}^{\left(d\right)}}$ allowing the multiplication to be conducted efficiently through bit-shifter. Here, we must emphasize that ${\textnormal{\textbf{FxP-QNet}}}$ performs a layer-wise quantization. One can also note that, compared to the previous quantization techniques discussed in Section~{\color{DarkBlue}\ref{sec:q_methods}}, the proposed quantization method is more computationally efficient as \textit{all} DNN data-structures are quantized into fixed-point numbers, and thus, \textit{all} the performed operations are integer arithmetic, or even bit-wise operations, whereas floating-point arithmetic is not involved at all.

In Figure~\ref{fig:FxP_QNets_Hardware}, we provide an overview of ${\textnormal{\textbf{FxP-QNet}}}$ low-precision computations. During forward propagation, integer activations and model parameters are used as inputs to a low-precision integer ${\text{C-Block}}$ consisting of ${\textnormal{\textit{CONV}}}$, ${\textnormal{\textit{Scale}}}$, and ${\textnormal{\textit{ReLU}}}$ layers. A MAC tree of integer arithmetic can be used to convolve an activation window $\mathbf{A}_\text{i-1,j}$ with a weight kernel $\mathbf{W}_\text{i,m}$ each of $N$ integers. To avoid the overflow, the partial results of the MAC tree in first and last layers are computed with $32$ bits, while on the other hand, MAC tree partial results in other layers are computed with $16$ bits. This is because these data-structures have a low bit-precision level during deployment and a limited data range with weights in bell-curve distribution.

Thereafter, the resultant feature from MAC tree needs to be accumulated with the bias parameter $\mathbf{B}_\text{i,m}$ to produce the output feature of ${\textnormal{\textit{CONV}}}$ layer. However, these two values have different scaling parameters. Hence, $\mathbf{B}_\text{i,m}$ must be left-shifted and sign-extended to have the same scalar and wordlength as MAC tree result before accumulation. When considering arithmetic operations within MAC tree, one can notice that multiplying two integers having scalars $2^{-\beta_{i-1}^{(a)}}$ and $2^{-\beta_{i}^{(w)}}$ yields an integer having $2^{-\big(\beta_{i-1}^{(a)} + \beta_{i}^{(w)}\big)}$ as its scalar. On the other hand, accumulating integers results in a value that has a scalar equal to that of the integers. Thus, the amount of shifting required to align fractional parts is specified by ${\textnormal{\textbf{FxP-QNet}}}$ as ${\delta_i = \beta_{i-1}^{(a)} + \beta_{i}^{(w)} - \beta_{i}^{(b)}}$, where $\beta_{i}^{(b)}$ is the scaling parameter of $\mathbf{B}_\text{i}$. 

Similarly, a $16$-bit integer multiplier and a $32$-bit integer accumulator can be used to carry out the computation of ${\textnormal{\textit{BN}}}$/${\textnormal{\textit{Scale}}}$ layers. Thereafter, the produced feature is compared to $0$ and the output value is the larger of them. The output feature from the ${\text{C-Block}}$ is then rounded and clipped after a right-shift operation by $\delta_{i+2}$, which is equal to ${\beta_{i-1}^{(a)} + \beta_{i}^{(w)} + \beta_{i+1}^{(w)} - \beta_{i+2}^{(a)}}$. Here, we must emphasize that each layer within a $\textnormal{C-Block}$ hypothetically considers activations produced by $\textnormal{C-Block}$ as the activations it produced. Hence, it is necessary to note that when mentioning layer $\ell$ activations, we mean the hypothetical activations of layer $\ell$, i.e., activations produced by $\textnormal{C-Block}$ in which layer $\ell$ is. Additionally, if a $\textnormal{C-Block}$ is followed by a ${\textnormal{\textit{FxP-QLayer}}}$, then when referring to the activations produced by $\textnormal{C-Block}$, we actually mean those produced by ${\textnormal{\textit{FxP-QLayer}}}$.

\subsection{Quantization Parameters Optimization} \label{sec:quantization_parameters_optimization}

A $k$-bit quantized data-structure can have several quantization domains, $\mathrm{domQ}$s in-short, based on the selected quantization parameters ($\mathrm{QPs}$), i.e., the integer length $\alpha$ and fractional length $\beta$. The $\mathrm{domQ}$ refers to the set of possible values that can be represented by the given $\mathrm{QPs}$. Figure~\ref{fig:quantization_parameters_optimization} shows the effect of these parameters on quantizing the weights of the first fully-connected layer in the ${\textnormal{VGG-}16}$ model~\cite{simonyan2014very}, i.e., the ${\textnormal{FC-}6}$ layer. As we can see, using a large $\beta$ represents many small weights more finely but at the expense of losing the accuracy in the representation of large weights. Here, we mean the magnitude of the weights, not the sign. One can also note that such data-structures tend to have a bell-curve distribution, which enables them to be quantized effectively with error-minimization-based techniques without the need for retraining~\cite{banner2019post}.

\Figure[!t](topskip=0pt, botskip=0pt, midskip=0pt)[trim=1.95cm 0.5cm 1.2cm 0.9cm, clip=true, width=0.99\textwidth]{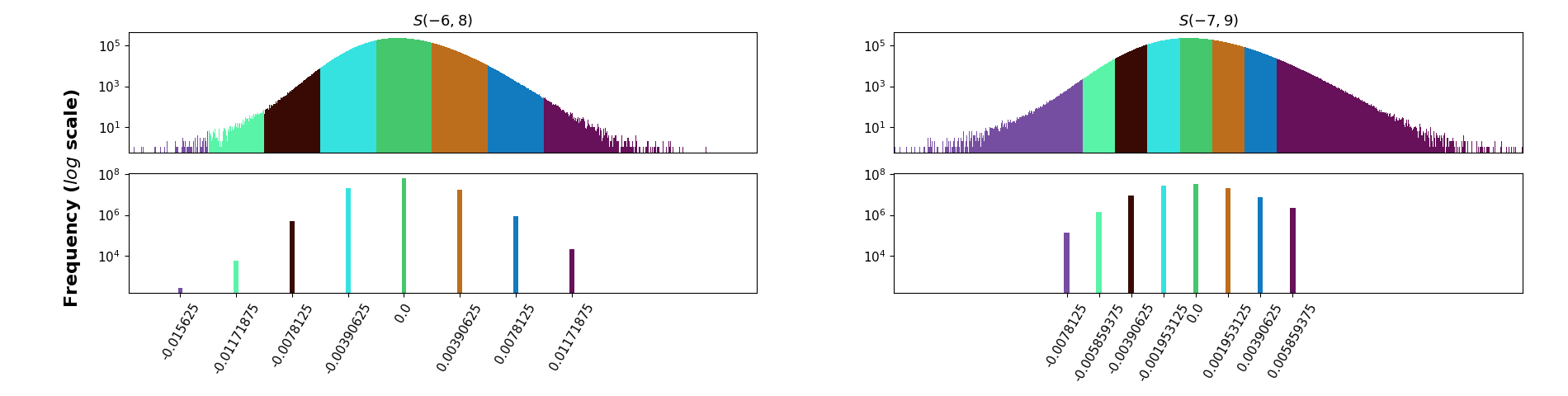}
{Distribution of ${\textnormal{\textit{FC}}}$-$6$ weights in VGG-16 model
before (\textit{top}) and after (\textit{bottom}) $3$-bit fixed-point quantization. The two figures on the left show the quantization of the weights with $\alpha$ and $\beta$ as $-5$ and $8$, respectively. While, the two figures on the right show the quantized weights with $\alpha$ equals $-6$ and $\beta$ equals $9$.
\label{fig:quantization_parameters_optimization}}

Therefore, an important step during data-structure quantization is to select appropriate $\mathrm{QPs}$ that result in the optimal $\mathrm{domQ}$. Prior quantization policies~\cite{jain2018compensated, guo2018angel, khan2020learning} choose the $\mathrm{domQ}$ with the highest overall network accuracy. However, it turns out, empirically, that the resulting network is prone to overfitting. Thus, retraining is mandatory to maintain the performance of the quantized network with an accuracy-based selection of $\mathrm{domQ}$. Alternatively, other quantization policies~\cite{migacz20178, qiu2016going, banner2018aciq, kravchik2019low, banner2019post} assume that if the perturbation in the distributions of data-structures due to discretization is minimized, then the network accuracy can be maintained. Consequently, we initially considered the optimal $\mathrm{QPs}$ as the ones that minimize the data-structure-based quantization error, denoted as $\mathcal{L}$, given by

\begin{equation} \label{eq:DSB_QE}
 \mathcal{L}\left(X; \alpha, \beta, \mathcal{S}\right) = \left\| X -\widetilde{X} \right\|_2^2
\end{equation}
\\
Here, $\widetilde{X}$ is the low-precision representation of the FP32 $X$ that is computed as mentioned before in Equations~\eqref{eq:fixed_point}, \eqref{eq:fp_k_bit_quantization}, and \eqref{eq:fp_1_bit_quantization}, and $\|\bullet\|_2^2$ is the squared $L2$ norm. However, we empirically observed that minimizing the quantization error for each layer and data-structure independently produces a local optimum solution. Such kind of local optimization results in a massive degradation in the overall accuracy. This is because the quantization error propagates to the succeeding layers and magnifies. The way of quantizing the activations, weights, and biases of a layer affects the activation distribution of all layers succeeding it. Additionally, quantizing data-structures independently may not allow for rigorous model compression. 

Hence, while quantizing a data-structure of a specific layer, the proposed optimization algorithm considers the quantization error of all data-structures in the same layer as well as the quantization error of data-structures in all the succeeding layers. We refer to these combined errors as the post-training self-distillation error ($\mathcal{E}$). However, we expect the data to be quantized less finely as dispersion increases. This is because the distance between consecutive values in $\mathrm{domQ}$ becomes large to better encode outliers. Additionally, as the number of elements in the data-structure increases, the quantization becomes more significant because many occurrences of quantization error propagate to layer output. Thus, the higher the dispersion and size of data-structure get, the more the skew in the statistics of the output activations can be incurred, which affects the following layers, and therefore degrades model's accuracy.

To take that into account, we assign a weight of importance, $\pi$, to each data-structure. Such that, $\pi$ is a quantitative metric for evaluating the impact of quantizing a data-structure on the output quality of its layer. The quantization error of a data-structure with a large $\pi$ is more expensive than the error due to quantizing a data-structure with a small $\pi$. The value of $\pi$ is set as follows\vspace{-2.08mm}

\begin{equation} \label{eq:importance}
 \pi_\ell^{(d)} = \frac{{\sigma_x}\strut_\ell^{(d)} \cdot \nu_\ell^{(d)}}{
 \mathlarger{\mathlarger{\sum}}_{i \, \in \, \{a, w, b\}} \! \! \! \! {\sigma_x}\strut_\ell^{(i)} \cdot \nu_\ell^{(i)}}
\end{equation}
where $\pi$ is normalized so as $\sum_{i \in \{a, w, b\}} \pi_\ell^{(i)} = 1, $ ${\forall \, \ell \in [L]}$, ${\sigma_x\strut_\ell^{(d)}}$ and $\nu_\ell^{(d)}$ are dispersion and size of data-structure type $d$ in layer $\ell$, respectively, and ${\sigma_x}\strut_\ell^{(d)}$ is computed as

\begin{equation}
 {\sigma_x}\strut_\ell^{(d)} = \sqrt{\frac{1}{\nu_\ell^{(d)}} \cdot \sum_{i=1}^{\nu_\ell^{(d)}} \left({X_i}_\ell^{\left(d\right)} - \widebar{X}_\ell^{\left(d\right)}\right)^2}
\end{equation}

Here, $\widebar{X}_\ell^{\left(d\right)}$ is the arithmetic mean of $X_\ell^{\left(d\right)}$, and ${{\sigma_x}\strut_\ell^{(a)}}$ is calculated for all of the calibration dataset samples and the mean is taken. Note that the calibration dataset is a subset of the validation dataset used to do what is called $\mathrm{QPs}$ calibration, hence the name. Remember that data-structures in different layers have different distributions. Therefore, the errors due to quantizing them must be normalized. Thus, post-training self-distillation error on quantizing the data-structure type $d$ in layer $\ell$ is calculated as

\begin{equation} \label{eq:CPB_QE}
 \mathcal{E}(d, \ell) = \mathlarger{\sum}_{i \, = \,\ell}^{L} \!\!\!\!\!\!\!\!\!\! \mathlarger{\sum}_{\;\;\;\;\;\;\;\;\;\;j \, \in \, \{a, w, b\}} \!\!\!\!\!\!\!\!\! \pi_i^{(j)} \cdot \frac{\mathcal{L}\left(X_i^{\left(j\right)}; \alpha_i^{(j)}, \beta_i^{(j)}, \mathcal{S}_i^{(j)}\right)}{\mathcal{R}_i^{(j)}}
\end{equation}
\\
where $\mathcal{L}(.; ., ., .) \, / \, \mathcal{R}$ is the normalized quantization error, and $\mathcal{R}_i^{(j)}$ is the range of data-structure type $j$ in layer $i$, 

\begin{equation} \label{eq:data_range}
 \mathcal{R}_i^{(j)} = {X_i^{\left(j\right)}}\strut_{\!\!\!\text{MAX}+} - {X_i^{\left(j\right)}}\strut_{\!\!\!\text{MAX}-}
\end{equation}
\\
and ${X_i^{\left(j\right)}}\strut_{\!\!\!\text{MAX}+}$ and ${X_i^{\left(j\right)}}\strut_{\!\!\!\text{MAX}-}$ are the largest and smallest FP32 values in the data-structure $X_i^{\left(j\right)}$, respectively. It is worth noting that, in the classification task, the last layer of the network produces the probabilities of classifying the input sample for each class. A classifier performs a prediction mapping from the input sample to the model outcome based on these probabilities, the model outcome is the class with the highest probability. Consequently, the classifier fails in predicting the class of the input sample when two, or more, classes have the same highest probability, which might occur when classification is performed on a very-low-precision DNN. Hence, we take into consideration the percentage of unclassified samples ($\mathrm{UnClass}$) as well as the percentage of mispredicted classes ($\mathrm{MisPred}$) in those unclassified samples, and minimize them to maintain the performance, given by

\setlength{\textfloatsep}{20pt}
\begin{algorithm}[!t]
\small
\SetAlgoLined
\DontPrintSemicolon
\KwIn{$\textnormal{An initial } \mathrm{QPs} \textnormal{ (}\alpha_\mathrm{init}\textnormal{, }\beta_\mathrm{init}\textnormal{) for the data-structure type}$ ${d\textnormal{ in layer }\ell\textnormal{ of a quantized network structure }\mathbf{Net_q}}$ ${\textnormal{and model }\{\mathbf{W}, \mathbf{B}\}\textnormal{ using }\mathrm{QPs}\textnormal{ sets }\{\pmb\alpha, \pmb\beta, \mathcal{S}\}\textnormal{, a}}$ ${\textnormal{reference FP32 model }\{\mathcal{W}, \mathcal{B}\}\textnormal{, and a validation set }\mathcal{T}}$.} 
\KwOut{${\textnormal{The optimum }\mathrm{QPs}\textnormal{ and network accuracy (Top-1/5)}}$.}
\SetKwFunction{QFO}{QParamsOpt}
\SetKwProg{Fn}{Procedure}{:}{}
\nonl\SetAlgoNoLine\Fn{\QFO{$\alpha_\mathrm{init}, \beta_\mathrm{init}, d, \ell, \mathbf{Net_q},$ 
${\,\,\,\,\,\,\,\,\,\,\,\,\,\,\,\,\,\,\,\,\,\,\,\,\,\,\,\,\,\,\,\,\,\,\,\,\,\,\,\,\,\,\,\,\,\,\,\,\,\,\,\,\,\,\,\,\,\,\,\,\,\,\{\mathbf{W}, \mathbf{B}\}, \{\pmb\alpha, \pmb\beta, \mathcal{S}\}, \{\mathcal{W}, \mathcal{B}\}, \mathcal{T}}$}}{
\SetAlgoLined
\tikzmk{A}{Initialize $\pmb\alpha_\ell^{\left(d\right)} \textnormal{ and } \alpha_\mathrm{best} \textnormal{ with } \alpha_\mathrm{init} \textnormal{, } \pmb\beta_\ell^{\left(d\right)} \textnormal{ and } \beta_\mathrm{best}$ ${\,\textnormal{ with } \beta_\mathrm{init}}$

Forward the network on calibration set
${\mathrm{forward}\big(\mathbf{Net_q}, \{\mathbf{W}, \mathbf{B}\}, \{\pmb\alpha, \pmb\beta, \mathcal{S}\}; \mathrm{sample}(\mathcal{T})\big)}$

Calculate the error $\mathcal{E}(d, \ell)$ using (\ref{eq:CPB_QE})

Calculate the percent of $\mathrm{UnClass}$ samples and $\mathrm{MisPred}$ classes using (\ref{eq:unclassified_samples}, \ref{eq:mispredicted_classes})

Initialize the best cost $\mathrm{Cost}_{\textnormal{best}} \gets \mathcal{E} + \mathrm{UnClass} + \mathrm{MisPred}$

}\tikzmk{B}\boxitLLS{pink}
\tikzmk{A}{\For(\textcolor{DarkBlue}{\Comment{\text{\footnotesize Range Saturation vs. Resolution}}}){$i \in \{-1, +1\}$}{
$k \gets 1$, $j \gets 0$

\While(\textcolor{DarkBlue}{\Comment{\text{\footnotesize Search Size}}}){$k \leq (K + j)$}
{
Update the quantization parameters $\pmb\alpha_\ell^{\left(d\right)} \gets \alpha_\mathrm{init} - i \times k\textnormal{, }\;\; \pmb\beta_\ell^{\left(d\right)} \gets \beta_\mathrm{init} + i \times k$

${\textnormal{Quantize the data-structure }(d, \ell)\textnormal{ using (\ref{eq:fp_k_bit_quantization}, \ref{eq:fp_1_bit_quantization})}}$

Forward the network on calibration set
${\mathrm{forward}\big(\mathbf{Net_q}, \{\mathbf{W}, \mathbf{B}\}, \{\pmb\alpha, \pmb\beta, \mathcal{S}\}; \mathrm{sample}(\mathcal{T})\big)}$

Calculate the error $\mathcal{E}(d, \ell)$ using (\ref{eq:CPB_QE})

Calculate the percent of $\mathrm{UnClass}$ samples and $\mathrm{MisPred}$ classes using (\ref{eq:unclassified_samples}, \ref{eq:mispredicted_classes})

Calculate the current cost $\mathrm{Cost}_{\textnormal{curr}} \gets \mathcal{E} + \mathrm{UnClass} + \mathrm{MisPred}$

\If{$\mathrm{Cost}_{\textnormal{curr}} < \mathrm{Cost}_{\textnormal{best}}$}{
 $\mathrm{Cost}_{\textnormal{best}} \gets \mathrm{Cost}_{\textnormal{curr}}$
 
 $\alpha_\mathrm{best} \gets \pmb\alpha_\ell^{\left(d\right)}\textnormal{, } \;\;\beta_\mathrm{best} \gets \pmb\beta_\ell^{\left(d\right)}\textnormal{, }\;\;j \gets j + 1$
}
$k \gets k + 1$
}}

}\tikzmk{B}\boxitLLNBS{blue}
\tikzmk{A}{${\textnormal{Set }\mathrm{QPs}\textnormal{ as    }\:\:\pmb\alpha_\ell^{\left(d\right)} \gets \alpha_\mathrm{best}\textnormal{, } \pmb\beta_\ell^{\left(d\right)} \gets \beta_\mathrm{best}}$

${\textnormal{Quantize the data-structure }(d, \ell)\textnormal{ using (\ref{eq:fp_k_bit_quantization}, \ref{eq:fp_1_bit_quantization})}}$

Forward the network on validation set
${\mathrm{Acc}_{\textnormal{best}} \gets \mathrm{forward}(\mathbf{Net_q}, \{\mathbf{W}, \mathbf{B}\}, \{\pmb\alpha, \pmb\beta, \mathcal{S}\}; \mathcal{T})}$
}\tikzmk{B}\boxitLLS{yellow}
\KwRet $\;\alpha_\mathrm{best}$, $\beta_\mathrm{best}$, $\mathrm{Acc}_{\textnormal{best}}$

}
\caption{${\textnormal{Quantization Parameters Optimization.}}$}
\label{Algorithm:QFO}
\end{algorithm}

\begin{equation} \label{eq:unclassified_samples}
 \mathrm{UnClass} = \frac{1}{N} \cdot \sum_{i=1}^{N} \mathcal{I}\left(\left|\mathcal{C}_i^{(1)}\right| > 1\right)
\end{equation}

\begin{equation} \label{eq:mispredicted_classes}
 \mathrm{MisPred} = \frac{1}{N \cdot \mathcal{T_C}} \cdot \sum_{i=1}^{N} \left(\left|\mathcal{C}_i^{(1)}\right| - 1\right)
\end{equation}
\\
where $\mathcal{I}\left(\cdot\right)$ is the indicator function; it returns $1$ if the condition is satisfied and $0$ otherwise, $\big|\mathcal{C}_i^{(1)}\big|$ is the number of predicted classes with the same highest probability for the $i$-th input sample, $N$ is the number of calibration dataset samples, and $\mathcal{T_C}$ is the number of dataset classes. Based on the discussion above, for optimizing $\mathrm{QPs}$ of the data-structure type $d$ in layer $\ell$ with an initial $\mathrm{QPs}$, denoted as ($\alpha_\mathrm{init}$, $\beta_\mathrm{init}$), the optimization framework systematically explores the trade-off between the resolution and range saturation spaces by modulating $(\alpha_\ell^{\left(d\right)}, \: \beta_\ell^{\left(d\right)})$ to minimize the overall post-training self-distillation and network prediction errors, denoted by $\mathcal{F}$, as\vspace{-1mm}

\begin{equation} \label{eq:quantization_parameters_optimization}
\begin{aligned}
& \underset{\alpha_\ell^{\left(d\right)}, \: \beta_\ell^{\left(d\right)}}{\text{minimize}}
& & \!\!\!\!\!\!\!\!\!\!\! \mathcal{F}(d, \ell) = \mathcal{E}(d, \ell) + \mathrm{UnClass} + \mathrm{MisPred}\\
& \;\;\;\;\;\; \text{subject to}
&& \alpha_\ell^{\left(d\right)} + \beta_\ell^{\left(d\right)} = \alpha_\mathrm{init} + \beta_\mathrm{init}
\end{aligned}
\end{equation}
\vspace{0.4mm}
Algorithm~\ref{Algorithm:QFO} summarizes the steps involved in finding the optimal $\mathrm{QPs}$. In the initialization step, the optimization framework initializes the quantization \textit{cost}, i.e., the $\mathcal{F}$ error, based on the initial $\mathrm{QPs}$ as shown in lines $1-5$. The search step, highlighted in blue, is used to explore the $2K$ neighboring $\mathrm{QPs}$ for the one with the minimum \textit{cost} as defined in Equation (\ref{eq:quantization_parameters_optimization}), where $K$ is the search space limit. Given that post-training self-distillation error, $\mathcal{E}$, is a convex function and network prediction error, i.e., $\mathrm{UnClass} + \mathrm{MisPred}$, has a relatively small magnitude, the search space can be safely reduced by setting $K$ to a small value and the global optimum $\mathrm{QPs}$ is insured by expanding the search space in the direction of minimum \textit{cost}. In other words, if a neighbor reduces the \textit{cost}, the framework takes a step further and expands the search space to verify the \textit {cost} of additional adjacent $\mathrm{QPs}$. It is worth noting that the forward process is adopted after each quantization operation because the impact of data-structure quantization with various $\mathrm{QPs}$ is unpredictable on both \textit{cost} and accuracy. Fortunately, this process is cheap in terms of computation when feed-forwarded on calibration and validation sets.

On the other hand, it is important to ensure, before starting to design a mixed low-precision DNN, that the accuracy drop for the initial solution proposed by \textit{preprocessor} is negligible. Therefore, the \textit{forward optimizer} fine-tunes the initial solution by employing Algorithm~\ref{Algorithm:QFO}. In doing so, it optimizes the initial $\mathrm{QPs}$ for the first layer, after which it moves to the next layer and so on until it reaches and optimizes $\mathrm{QPs}$ of the last layer, and hence the name \textit{forward optimizer}. Specifically, for each layer, \textit{forward optimizer} identifies the optimal $\mathrm{QPs}$ for weights, then for bias, and finally for activations.

The reason behind adopting this order for layers and data-structures is that the quantization of activations, weights, and biases of a layer does not affect the activation distribution of all the layers that precede it, while on the other hand the activations produced by a layer are affected by how the learnable parameters of the same layer are quantized. Note that while optimizing $\mathrm{QPs}$ for a layer, the data-structures in all the succeeding layers are floating-point values, conversely, all the data-structures in preceding layers are fixed-point values.

\subsection{Bit-Precision Level Reduction} \label{sec:bit_precision_level_reduction}

The primary goal of quantization is to reduce the computational cost as well as having a minimal memory footprint to accelerate DNN inference. Recalling that DNNs are designed to achieve superior accuracy on computer vision tasks, such as image classification, it is therefore important to investigate the role each layer plays in forming a good predictor before reducing its bit-precision. In light of the above, we conducted an experiment to investigate the sensitivity of each data-structure to quantization. The results, which are illustrated in Section~{\color{DarkBlue}\ref{sec:sensitivity_results}}, show that quantizing different DNN data-structures causes various levels of network accuracy degradation, and the bit-precision of some data-structures can be significantly reduced without causing noticeable performance degradation.

Even more importantly, though, the results show that data-structures in deeper layers are more robust to quantization than the ones in shallower layers. During the forward propagation in DNNs, the stacked layers extract internal features and progressively enhance the dissimilarity between the hierarchical features. Compared to the more separable features in deeper layers, the features in shallower layers are distributed on complex manifolds and overlap mutually. Therefore, extracting the meaningful intermediate representations in shallower layers requires the use of more accurate neurons, neurons with high bit-precision level, than the ones in deeper layers~\cite{chu2021mixed}.

Motivated by the aforementioned observations, we propose an iterative accuracy-driven bit-precision reduction framework that designs a quantized DNN flexibly with a mixed-precision rather than a $k$-bit homogeneous network. The proposed framework aims to find the best quantization level for each data-structure based on the trade-off between accuracy and low-precision requirements. Nonetheless, the search space for choosing the appropriate quantization level of each data-structure in different layers is exponential in the number of layers and data-structures, which makes the exhaustive search impractical. For instance, VGG-16 exposes a search space of size $N_{{lv}_1}^{2L-2} \times N_{{lv}_2}^{L+4} = 8^{30} \times {32}^{20} > {10}^{57}$, where $L$ is the number of layers, $N_{{lv}_1}$ is the number of possible quantization levels for activations and weights in all layers except the first and last layers, and $N_{{lv}_2}$ is the number of possible quantization levels for other data-structures. 

To overcome this issue, we present a simple deterministic technique for dividing this huge search space into relatively small feasible sets of solutions. The bit-precision level of the data-structures within each set is determined based on their quantization robustness. With this technique, first and last layers can also be quantized to lower levels without having a drastic impact on performance. As illustrated in Algorithm~\ref{Algorithm:BPR_1}, the proposed framework starts with an initial quantized DNN where the data-structures are in fixed-point representation, each of which with possibly different wordlength from $32$ to $8$. Thereafter, it progressively quantizes the data-structures until it reaches the maximum compression rate under a given accuracy degradation constraint as elaborated in the three steps given below.

\subsubsection{Data-Structure Clustering}

In typical DNNs, forward propagation analysis demonstrated that the convolution (${\textnormal{\textit{CONV}}}$) operation takes more than $90$\% of the total computational time~\cite{cong2014minimizing}. Thus, to improve the computational efficiency, the proposed framework focuses on reducing the bit-precision level of the weights of ${\textnormal{\textit{CONV}}}$ and ${\textnormal{\textit{FC}}}$ layers as well as the activations. This is done through dividing the data-structures into two sets of pairs $(d, \ell)$ such that $d$ is the data-structure type, $d \in \{a, w, b\}$, and $\ell$ is the layer ID, $\ell \in [L]$, as

\begin{equation} \label{eq:level_1_clustering}
 \begin{aligned}
 \!\!\mathbf{G}_1 \! = & \, \big\{(a, \ell) : \ell \in [L]\big\} \,\, {\textstyle \bigcup} \\
 & \, \big\{(w, \ell) : \ell \in \mathfrak{L_p}, \, \mathrm{type}(w, \ell) \in \{\text{\small CONV}, \text{\small FC}\}\big\} & \\
 \!\!\mathbf{G}_2 \!= & \, \big\{(d, \ell) : d \neq a, \,\ell \in \mathfrak{L_p}, \, (d, \ell) \notin \mathbf{G}_1\big\}
\end{aligned}
\end{equation}

Here, $\mathfrak{L_p}$ is a set of all layers containing learnable parameters. Thus, the first set, $\mathbf{G}_1$, contains precisely the elements for ${\textnormal{\textit{CONV}}}$ and ${\textnormal{\textit{FC}}}$ weights as well as the activations. On the other hand, the second set, $\mathbf{G}_2$, contains the elements for the learnable parameters not in $\mathbf{G}_1$. Note that there is no need to quantize the activations of the output layer, whereas the activations of other layers, including the input layer, are potential for bit-precision level reduction.

Another important issue that the proposed framework takes into account is the effect of quantizing each data-structure on the memory footprint. As shown in Figure~\ref{fig:data_structure_size}, the data-structures vary greatly in their need for memory space. One can also note that majority of learnable parameters are concentrated in deeper layers. Hence, for instance, reducing the bit-precision of AlexNet ${\textnormal{\textit{CONV}}}$-1 and ${\textnormal{\textit{FC}}}$-6 weights to the next level reduces the memory requirements by $4.25$KB and $4.72$MB, respectively.

\setlength{\textfloatsep}{20pt}
\begin{algorithm}[!t]
\small
\SetAlgoLined
\DontPrintSemicolon
\KwIn{${\textnormal{A quantized network structure }\mathbf{Net_q}\textnormal{ and model}}$ ${\{\mathbf{W}, \mathbf{B}\}\textnormal{ using }\mathrm{QPs}\textnormal{ sets }\{\pmb\alpha, \pmb\beta, \mathcal{S}\}\textnormal{, a reference FP32 }}$ ${\textnormal{model }\{\mathcal{W}, \mathcal{B}\} \textnormal{ and accuracy }\mathrm{Acc}_{\textnormal{FP32}}\textnormal{, an accuracy}}$ ${\textnormal{degradation threshold }\Delta_{\textnormal{thld}} \textnormal{, and a validation set } \mathcal{T}}$.} 
\KwOut{A mixed low-precision deep neural network.}
\SetKwFunction{BPR}{QReduction}
\SetKwFunction{Reduce}{Reduce}
\SetKwProg{Fn}{Procedure}{:}{}
\nonl\SetAlgoNoLine\Fn{\BPR{$\mathbf{Net_q}, \{\mathbf{W}, \mathbf{B}\}, \{\pmb\alpha, \pmb\beta, \mathcal{S}\}, $
${\,\,\,\,\,\,\,\,\,\,\,\,\,\,\,\,\,\,\,\,\,\,\,\,\,\,\,\,\,\,\,\,\,\,\,\,\,\,\,\,\,\,\,\,\,\,\,\,\,\,\,\,\,\,\,\,\,\,\,\,\,\,\{\mathcal{W}, \mathcal{B}\}, \mathrm{Acc}_{\textnormal{FP32}}, \Delta_{\textnormal{thld}}, \mathcal{T}}$}}{
\SetAlgoLined
\tikzmk{A}{Cluster the data-structures using (\ref{eq:level_1_clustering}, \ref{eq:level_2_clustering}) ${\,\,\,\mathcal{G}^{(1)} \gets \left\{\mathcal{G}^{(1)}_{\nu_1}, \dots, \mathcal{G}^{(1)}_{\nu_{n_1}}\right\}, \,\, \mathcal{G}^{(2)} \gets \left\{\mathcal{G}^{(2)}_{\nu_1}, \dots, \mathcal{G}^{(2)}_{\nu_{n_2}}\right\}}$

Sort ${\text{Level-}2}$ clusters in decreasing order such that ${\nu_{1} > \nu_{2} > \dots > \nu_{n_t}, \, \forall \, t \in \{1, 2\}}$

}\tikzmk{B}\boxitLL{pink}
\tikzmk{A}{Initialize the traversal algorithm ${\,\,\, visit[i] \gets true, \forall \, i \in \{1, 2\}}$ \textcolor{DarkBlue}{\Comment{\text{\scriptsize Visit all clusters}}}

\nonl $\,\,\,t \gets 1$ \textcolor{DarkBlue}{\Comment{\text{\scriptsize Start with $\mathcal{G}^{(1)}$}}}

\nonl $\,\,\,\lambda \gets 0$ \textcolor{DarkBlue}{\Comment{\text{\scriptsize Restrict the degradation threshold}}}

\Repeat(\textcolor{DarkBlue}{\Comment{\text{\footnotesize ${\text{Level-}1}$ traversal}}}){\textnormal{\textbf{not}} $visit[t]$}{

$visit[t] \gets false$

$j \gets 1$

\While(\textcolor{DarkBlue}{\Comment{\text{\footnotesize ${\text{Level-}2}$ traversal}}}){$j \leq n_t$}{
Select a candidate data-structure in $\mathcal{G}^{(t)}_{\nu_j}$ cluster
${\left(d, \ell, \alpha_\mathrm{new}, \beta_\mathrm{new}\right), change \gets \texttt{Reduce}(\mathcal{G}^{(t)}_{\nu_j}, }$ ${\,\,\,\,\,\,\,\,\,\,\,\,\,\,\,\,\,\,\,\,\,\,\,\,\,\,\,\,\,\,\,\,\,\,\,\,\,\,\,\,\,\,\,\,\,\,\,\,\,\,\,\,\,\,\,\Delta_{\textnormal{thld}} \cdot \mathcal{I}\left(t \neq 2\right), \lambda, \cdots)}$

\uIf(\textcolor{DarkBlue}{\Comment{\text{\scriptsize Perform the reduction}}}){$change$}{
Update the quantization parameters $\pmb\alpha_\ell^{\left(d\right)} \gets \alpha_\mathrm{new}\textnormal{, }\;\; \pmb\beta_\ell^{\left(d\right)} \gets \beta_\mathrm{new}$

${\textnormal{Quantize the data-structure }(d, \ell)\textnormal{ using (\ref{eq:fp_k_bit_quantization}, \ref{eq:fp_1_bit_quantization})}}$

\If(\textcolor{DarkBlue}{\Comment{\text{\scriptsize Saturated data}}}){$\pmb\alpha_\ell^{\left(d\right)} + \pmb\beta_\ell^{\left(d\right)} = 1$}{
$\mathcal{G}^{(t)}_{\nu_j} \gets \mathcal{G}^{(t)}_{\nu_j} - \left\{(d, \ell)\right\}$
}

$visit[i] \gets true, \forall \, i \in \{1, 2\}$
}
\Else(\textcolor{DarkBlue}{\Comment{\text{\scriptsize Advance to next ${\text{Level-}2}$ cluster}}}){$j \gets j + 1$}

}

\If(\textcolor{DarkBlue}{\Comment{\text{\scriptsize Advance to other ${\text{Level-}1}$ cluster}}}){\textnormal{\textbf{not}} $visit[t]$}{$t \gets t + {(-1)}^{t + 1}$}
$\lambda \gets 1$
}

}\tikzmk{B}\boxitLL{yellow}
\KwRet $\;\{\mathbf{W}, \mathbf{B}\}$, $\{\pmb\alpha, \pmb\beta, \mathcal{S}\}$

}
\caption{$\textnormal{Bit-Precision Level Reduction - Part 1.}$}
\label{Algorithm:BPR_1}
\end{algorithm}

As it is evident from the given example, the contribution of learnable parameters in deeper layers to the compression rate is much more than that of the shallower layers when both are quantized to the next bit-precision level. Accordingly, the proposed framework adopts the two-level clustering approach~\cite{larose2014discovering} to represent the data-structures for wordlength reduction based on their type and size, such that

\begin{equation} \label{eq:level_2_clustering}
 \begin{aligned}
 \mathcal{G}^{(t)} \! = & \, \bigg\{\mathcal{G}^{(t)}_{\nu_1}, \mathcal{G}^{(t)}_{\nu_2}, \dots, \mathcal{G}^{(t)}_{\nu_{n_t}} : \forall (d_i, \ell_i) \in \mathbf{G}_t, \\
 & \,\,\,\, (d_i, \ell_i) \in \mathcal{G}^{(t)}_{\nu_j} \iff \nu_{\ell_i}^{(d_i)} = \nu_j, t \in \{1, 2\}, \\
 & \,\,\,\, 1 \leq i \leq m, m = |\mathbf{G}_t|, 1 \leq j \leq {n_t}, \\
 & \,\,\,\, {n_t} = \left|\left\{\nu_{\ell_1}^{(d_1)}, \nu_{\ell_2}^{(d_2)}, \dots, \nu_{\ell_m}^{(d_m)}\right\}\right| \bigg\}
\end{aligned}
\end{equation}

In other words, the first clustering level maps a data-structure represented as a $2$-tuple $(d, \ell)$ into either $\mathcal{G}^{(1)}$ or $\mathcal{G}^{(2)}$ clusters on the basis of data-structure type as discussed in Equation~(\ref{eq:level_1_clustering}). Then, in the second clustering level, the tuples within each ${\text{Level-}1}$ cluster are grouped into $n_t$ sub-clusters based on their sizes, i.e., $n_t$-sizes clustering. Consequently, a ${\text{Level-}2}$ cluster, say $\mathcal{G}^{(t)}_{\nu_j}$, contains all elements in $\mathbf{G}_t$ whose data-structure size is ${\nu_j}$. Here, $n_t$ is the number of unique sizes in $\mathbf{G}_t$.

\subsubsection{Clusters Traversal}

The order of picking data-structures for quantization plays an important role in achieving rigorous compression for DNNs designed with accuracy degradation constraint. For this purpose, the proposed framework imposes to visit cluster $\mathcal{G}^{(1)}$ before cluster $\mathcal{G}^{(2)}$. Within each ${\text{Level-}1}$ cluster, the sub-cluster with the larger $\nu$ is given a higher priority for reducing the bit-precision level of its data-structures. When visiting a ${\text{Level-}2}$ cluster, the proposed framework carries out a progressive wordlength reduction to its data-structures, one data-structure at a time, until no further reduction is possible. Thereafter, it advances to the next sibling cluster. 

\setlength{\textfloatsep}{20pt}
\begin{algorithm}[!t]
\ContinuedFloat
\small
\SetAlgoLined
\DontPrintSemicolon
\KwIn{${\textnormal{A quantized network structure }\mathbf{Net_q}\textnormal{ and model}}$ ${\{\mathbf{W}, \mathbf{B}\}\textnormal{ using }\mathrm{QPs}\textnormal{ sets }\{\pmb\alpha, \pmb\beta, \mathcal{S}\}\textnormal{, a reference FP32 }}$ ${\textnormal{model }\{\mathcal{W}, \mathcal{B}\} \textnormal{ and accuracy }\mathrm{Acc}_{\textnormal{FP32}}\textnormal{, a quantized}}$ ${\textnormal{data-structure set }\mathcal{G}^{(t)}_{\nu_j} \textnormal{, a degradation threshold }\Delta_\textnormal{thld}}$ $\textnormal{and mask }\lambda\textnormal{, and a validation set } \mathcal{T}$.} 
\KwOut{A selected candidate data-structure for the wordlength reduction, if exists.}
\SetKwFunction{BPR}{QReduction}
\SetKwFunction{Reduce}{Reduce}
\SetKwProg{Fn}{Procedure}{:}{}
\nonl\SetAlgoNoLine\Fn{\Reduce{$\mathcal{G}^{(t)}_{\nu_j}, \Delta_{\textnormal{thld}}, \lambda, \mathbf{Net_q}, \{\mathbf{W}, \mathbf{B}\} $
${\,\,\,\,\,\,\,\,\,\,\,\,\,\,\,\,\,\,\,\,\,\,\,\,\,\,\,\,\,\,\,\,\,\,\,\,\,\,\,\,\,\,\,\,\,\,\,\,\,\{\pmb\alpha, \pmb\beta, \mathcal{S}\}, \{\mathcal{W}, \mathcal{B}\}, \mathrm{Acc}_{\textnormal{FP32}}, \mathcal{T}}$}}{
\SetAlgoLined
Initialize the reduction parameters ${\,\,\,cand \gets \{\}, \,\,climb \gets false}$

\SetAlgoNoLine\Jump{}{
\SetAlgoLined
\tikzmk{A}{\ForAll{$u \in \{1, \cdots, U\}, U = |\mathcal{G}^{(t)}_{\nu_j}|$}{
$sol_u \gets \{\}$

\uIf(\textcolor{DarkBlue}{\Comment{\text{\footnotesize Hill climb reduction}}}){$climb$}{
$\theta_{high} \gets \theta_{low} \gets  2$
}
\Else{
\uIf(\textcolor{DarkBlue}{\Comment{\text{\footnotesize Gradual}}}){$\pmb\alpha_{\ell_u}^{\left(d_u\right)} + \pmb\beta_{\ell_u}^{\left(d_u\right)} \leq 8$}{
$\theta_{high} \gets \theta_{low} \gets  1$
}
\Else(\textcolor{DarkBlue}{\Comment{\text{\footnotesize Aggressive}}}){
$\theta_{high} \gets \pmb\alpha_{\ell_u}^{\left(d_u\right)} + \pmb\beta_{\ell_u}^{\left(d_u\right)} - 1$, $\theta_{low} \gets  1$
}
}

\While(\textcolor{DarkBlue}{\Comment{\text{\footnotesize Binary search}}}){$\theta_{low} \leq \theta_{high}$}{
$\theta_u \gets \left\lceil \left(\theta_{low} + \theta_{high}\right)/\,2\right\rceil$

Optimize the quantization parameters
${\alpha_{u}^{new}, \beta_{u}^{new}, \mathrm{Acc}_{u}^{new} \gets \texttt{QParamsOpt}(}$ ${\,\,\,\,\,\,\,\,\,\,\,\,\,\,\,\,\,\,\,\,\,\,\,\,\,\,\,\pmb\alpha_{\ell_u}^{\left(d_u\right)}, \pmb\beta_{\ell_u}^{\left(d_u\right)} - \theta_u, d_u, \ell_u, \cdots)}$

$\Delta_u \gets \left(\mathrm{Acc}_{\textnormal{FP32}} - \mathrm{Acc}_{u}^{new}\right)$

$\varphi \gets \lambda + (1 - \lambda) \cdot \theta_u \cdot \nu_{\ell_u}^{(d_u)} / \sum_{\:\ell \, \in \, [L]}\nu_{\ell}^{(d_u)}$

\uIf(\textcolor{DarkBlue}{\Comment{\text{\scriptsize Valid reduction}}}){$\Delta_u \leq \varphi \cdot \Delta_{\textnormal{thld}}$}{
${sol_u \gets \left\{\left(d_u, \ell_u, \theta_u, \Delta_u, \alpha_{u}^{new}, \beta_{u}^{new}\right)\right\}}$

$\theta_{low} \gets \theta_u + 1$
}
\lElse{
$\theta_{high} \gets \theta_u - 1$
}
}
$cand \gets cand + sol_u$
}
}\tikzmk{B}\boxitLLLS{blue}
\tikzmk{A}{\uIf(\textcolor{DarkBlue}{\Comment{\text{\footnotesize Select a candidate for reduction}}}){$cand \neq \phi$}{
\KwRet $\;\texttt{select}(cand)$, $true$
}
\Else(\textcolor{DarkBlue}{\Comment{\text{\footnotesize No candidate reductions}}}){
\uIf(\textcolor{DarkBlue}{\Comment{\text{\scriptsize Hill climbing}}}){\textit{\textbf{not}} $climb$}{
$climb \gets true$

\textbf{go to \texttt{start}}
}
\lElse{
\KwRet $\;\left(\textnormal{None}, \cdots, \textnormal{None}\right)$, $false$
}}
}\nonl\tikzmk{B}\vskip-10pt\vskip-10pt\rlap{\boxitLLL{red}}
}
}\caption{$\textnormal{Bit-Precision Level Reduction - Part 2.}$}
\label{Algorithm:BPR_2}
\end{algorithm}

When any of the data-structure belonging to the currently visited ${\text{Level-}1}$ cluster succeed in reducing the bit-precision level, the proposed framework retraverses all of its ${\text{Level-}2}$ clusters again. Otherwise, it advances to visit the other ${\text{Level-}1}$ cluster. In this context, we define an \textit{episode} as a single pass through $\mathcal{G}^{(1)}$ and $\mathcal{G}^{(2)}$ clusters. The cluster traversal algorithm continues to iterate between $\mathcal{G}^{(1)}$ and $\mathcal{G}^{(2)}$ clusters until it converges and no changes in the bit-precision level are observed in an \textit{episode}.

\subsubsection{Intra-Cluster Bit-Precision Level Reduction}

At this stage, the framework receives a set data-structures, each in fixed-point representation, for the currently visited Level-$2$ cluster as ${\mathcal{G}^{(t)}_{\nu_j} \! = \! \left\{(d_1, \ell_1), \cdots, (d_U, \ell_U)\right\}}$, and then, a data-structure, say $(d_u, \ell_u)$, is chosen and quantized to one of the next bit-precision levels from $(\alpha_{\ell_u}^{\left(d_u\right)} + \beta_{\ell_u}^{\left(d_u\right)} - 1)$ to $1$, where ${1 \leq u \leq U}$, and $U = \big|\mathcal{G}^{(t)}_{\nu_j}\big|$. It is noteworthy that during wordlength reduction process, data-structures belonging to $\mathcal{G}^{(t)}_{\nu_j}$ may have different bit-precision levels ranging from $32$ to $2$.

Here, all the data-structures in $\mathcal{G}^{(t)}_{\nu_j}$ compete with each other to be picked and quantized to a next bit-precision level. More precisely, each data-structure is assigned its next level of bit-precision, and then, quantized as discussed in Section~{\color{DarkBlue}\ref{sec:dynamic_fixed_point_quantization}}, and finally the quantization parameters are optimized based on the discussion in Section~{\color{DarkBlue}\ref{sec:quantization_parameters_optimization}}. Note that the reduction in the bit-precision level is done through lessening the fractional length. As thus, to reduce the bit-precision level of data-structure $(d, \ell)$ by $\theta$ bits, the bit-precision reduction framework updates $\beta_{\ell}^{\left(d\right)}$ as $\beta_{\ell}^{\left(d\right)} = \beta_{\ell}^{\left(d\right)} - \theta$. When a data-structure is quantized to the minimum bit-precision level, it will be removed from its ${\text{Level-}2}$ cluster and therefore will not be considered for further reduction.

To determine the next bit-precision level for data-structure $(d, \ell)$, we advocate the use of the gradual reduction, i.e., $\theta = 1$, when ${\alpha_{\ell}^{\left(d\right)} + \beta_{\ell}^{\left(d\right)} \leq 8}$. On the other hand, we adopt the use of the binary search algorithm to look for the lowest acceptable bit-precision level, in terms of accuracy, when ${\alpha_{\ell}^{\left(d\right)} + \beta_{\ell}^{\left(d\right)} > 8}$. While the aggressive reduction speeds up the reduction process in the initial stages, the gradual reduction prohibits the potential early presence of bottleneck quantization layer~\cite{zhuang2018towards}. Such a layer prevents other data-structures in the pipeline from being quantized to the next level without violating the accuracy degradation constraint.

Upon completion of the competition, the information about all the data-structures that successfully reduced their bit-precision level without violating the accuracy degradation threshold ($\Delta_{\textnormal{thld}}$) will be in the candidate set, referred to as $cand$. These information include the optimum quantization parameters $(\alpha_{u}^{new}, \beta_{u}^{new})$ and the amount of accuracy drop $(\Delta_u)$ after reducing the bit-precision level of data-structure $(d_u, \ell_u)$ by an amount of $\theta_u$ bits. If more than one candidate was observed during any reduction stage, only one candidate is accepted. Therefore, we defined the operation \texttt{select}$(cand)$ to be the selection of the candidate $sol_u$ as $\left(d_u, \ell_u, \alpha_{u}^{new}, \beta_{u}^{new}\right)$ based on the following ranking order: \begin{enumerate*}[label=(\roman*)]
\item reduction amount $(\gg \theta)$, \item amount of accuracy drop $(\ll \Delta)$, \item layer ID $(\gg \ell)$, and \item data-structure type $(w \textnormal{, then } a \textnormal{, then } b)$\end{enumerate*}. 

On the other hand, we apply the hill climbing optimization technique~\cite{sait1999iterative} when no candidates are found for wordlength reduction. Hill climbing heuristic is adopted to avoid getting stuck in a local optima quantized DNN solution through accepting a bad wordlength reduction solution. Here, we need to emphasize that while the framework tries to reduce bit-precision level, it also tries smartly and innovatively to add a positive error that reduces the overall network error by selecting $\alpha$ and $\beta$ which results in the lowest $\mathcal{F}$ as mentioned in Equation \eqref{eq:quantization_parameters_optimization}. We found hill climbing to be very useful for this.

It is worth noting that reducing the wordlength of a data-structure can impact the level of quantization that other data-structures can reach. Thus, to prevent an early over-quantized data-structure, the bit-precision reduction framework rejects candidate data-structures that cause degradation more than their size ratio of the degradation threshold times the reduction amount during the first traversal visit of ${\text{Level-}1}$ data-structures. In other words, a data-structure $(d_u, \ell_u)$ that is a member of $\mathcal{G}^{(1)}$ can be a candidate during the first traversal visit of $\mathcal{G}^{(1)}$ elements if and only if,

\begin{equation}
 \Delta_u \leq \Delta_{\textnormal{thld}} \cdot \theta_u \cdot \frac{\nu_{\ell_u}^{(d_u)}}{\sum\limits_{\ell \, \in \, [L]} \! \nu_{\ell}^{(d_u)}}
\end{equation}

Note that the bit-precision reduction framework allows $\mathcal{G}^{(2)}$ data-structures to reduce their bit-precision level only when they do not cause further accuracy degradation. As such, the proposed framework is accuracy-driven as it decides when and to what extent to quantize each data-structure in order to have the minimum accuracy degradation and the maximum compression rate. Furthermore, ${\textnormal{\textbf{FxP-QNet}}}$ does not require domain experts, and more important, it frees the human labor from exploring the vast search space to choose the appropriate quantization level for each data-structure in various layers.

\section{Experiments and Results} \label{sec:experiments_results}

In this section, we first describe our experimental settings and then proceed to analyze the rounding strategies used to determine the initial quantization parameters ($\mathrm{QPs}$). Next, we discuss and compare the performance of the proposed post-training self-distillation and network prediction errors function, denoted as $\mathcal{F}$, with the commonly used loss functions to further demonstrate the effectiveness of the proposed $\mathrm{QPs}$ optimization framework. We also investigate the sensitivity of data-structures in different layers to quantization. Then, we employ the proposed ${\textnormal{\textbf{FxP-QNet}}}$ framework to design mixed low-precision networks for three widely used typical DNNs and discuss the results and findings. Finally, we compare the mixed low-precision DNNs designed using ${\textnormal{\textbf{FxP-QNet}}}$ framework with the state-of-the-art low bit-precision quantization frameworks.

\subsection{Experimental Settings}

To demonstrate the versatility of ${\textnormal{\textbf{FxP-QNet}}}$ in designing mixed low-precision DNNs, we conduct extensive experiments on the ImageNet large scale visual recognition challenge (ILSVRC) dataset~\cite{russakovsky2015imagenet}, which is known as one of the most challenging image classification benchmark so far. The ILSVRC-2012 ImageNet dataset has about $1.28$ million images for training and $50$ thousand images for validation, all are natural high-resolution images. Each image is annotated as one of $1,000$ classes. To avoid using the training set, which is the motivation part for post-training quantization, samples from the ILSVRC-2012 validation set are used to do what is called calibration of $\mathrm{QPs}$. Hence, it is referred to as the calibration set. The goal is to keep the calibration set as small as possible. Using the validation images, we report the evaluation results with two standard measures; top-1 accuracy/error-rate and top-5 accuracy/error-rate.

\Figure[t!][width=0.48\textwidth, trim={0cm 0.5cm 0cm 2cm}]{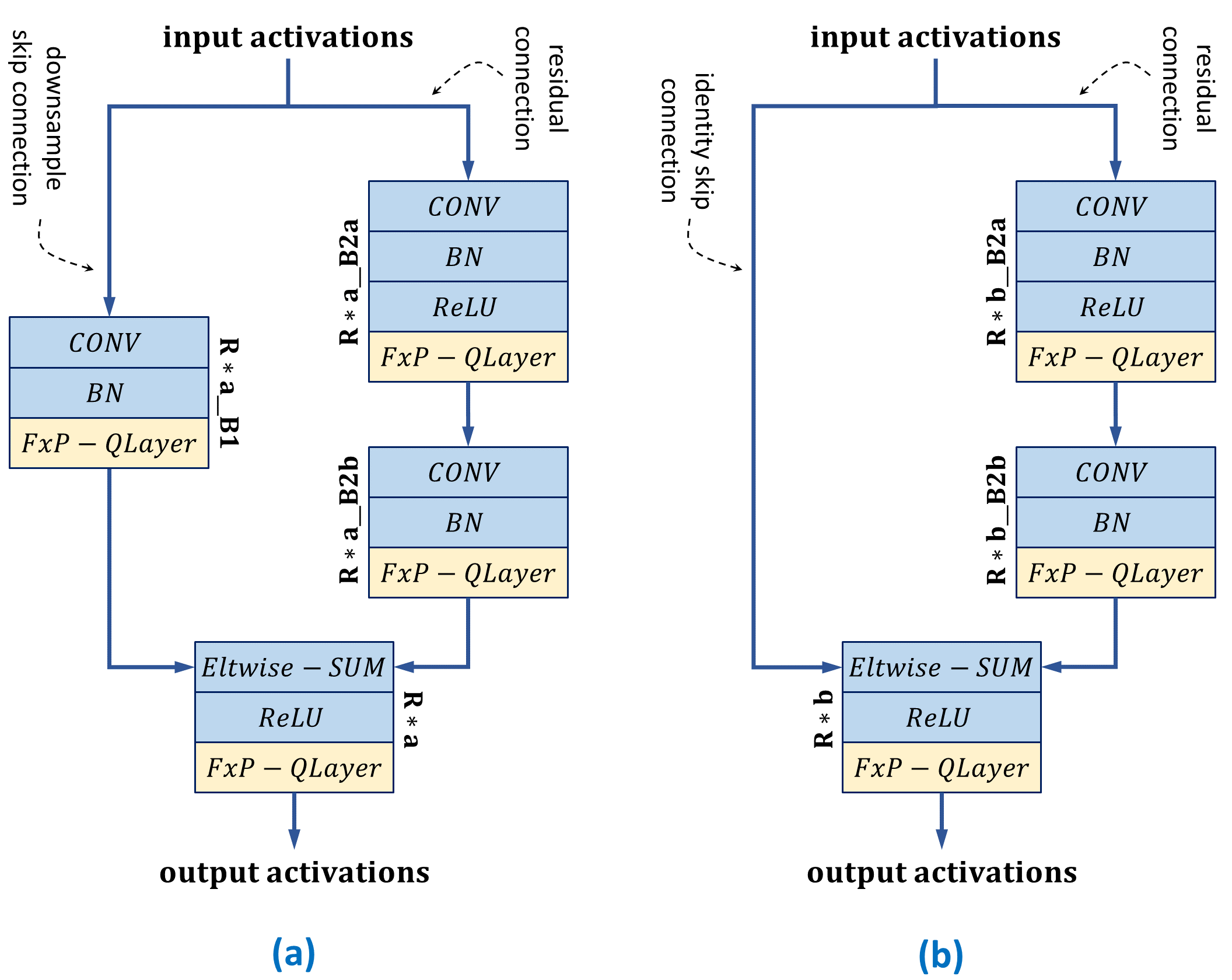}
{The different types of basic residual blocks used in ResNet-18; (a) with downsample skip connection, and (b) with identity skip connection. The name of each computational block is also given where, for instance, the star symbol in the name \textbf{R$\star$a\_B2b} denotes the stage ID, \textbf{R$\star$a} means the first residual block in this stage, \textbf{R$\star$a\_B2} refers to the second branch in this block, i.e., the residual skip connection, and finally \textbf{R$\star$a\_B2b} indicates the second computational block in this branch.
\label{fig:ResNet_Blocks}}

For those experiments, we used our ${\textnormal{\textbf{FxP-QNet}}}$ to design mixed low-precision networks for three benchmark architectures, namely, AlexNet~\cite{krizhevsky2012imagenet}, VGG-16~\cite{simonyan2014very}, and ResNet-18~\cite{he2016deep}, as they are well-known in the field of image classification and have been extensively studied in the literature. The ${\textnormal{\textbf{FxP-QNet}}}$ is implemented on top of Caffe~\cite{jia2014caffe}, a popular deep learning framework. The pre-trained full-precision models are taken from the Caffe Model Zoo without any fine-tuning or extra retraining. For the ResNet-18 model, we use the publicly available re-implementation by Facebook. Details of ResNet-18, VGG-16, and AlexNet architectures are presented next.
\\
\textbf{AlexNet} consists of five convolutional (${\textnormal{\textit{CONV}}}$) layers and three fully-connected (${\textnormal{\textit{FC}}}$) layers. Each of these layers, except for the last ${\textnormal{\textit{FC}}}$ layer, is followed by batch normalization (${\textnormal{\textit{BN}}}$) and rectified linear unit (${\textnormal{\textit{ReLU}}}$) layers. In addition, three ${\textnormal{\textit{MAX-POOL}}}$ layers are employed with the first, second, and last ${\textnormal{\textit{CONV}}}$ layers.
\\
\textbf{VGG-16} is similar to AlexNet architecture in terms of the number of ${\textnormal{\textit{FC}}}$ layers. However, it contains five groups of ${\textnormal{\textit{CONV}}}$ layers, and each group is followed by a ${\textnormal{\textit{MAX-POOL}}}$ layer. The first and second groups consist of two ${\textnormal{\textit{CONV}}}$ layers each while each of the last three groups consist of three ${\textnormal{\textit{CONV}}}$ layers according to model-D configurations.
\\
\textbf{ResNets-18} is a five-stage network followed by an $\textnormal{\textit{AVG-POOL}}$ and finally a ${\textnormal{\textit{FC}}}$ layer. The first stage consists of ${\textnormal{\textit{CONV}}}$, ${\textnormal{\textit{BN}}}$, ${\textnormal{\textit{ReLU}}}$, and ${\textnormal{\textit{MAX-POOL}}}$ layers. Each of the remaining four stages consists of two blocks, where layers in the residual connection of each block learn the residual, hence the name residual network. The first residual block contains a downsample skip connection, while the second residual block uses an identity skip connection. Figure~\ref{fig:ResNet_Blocks} illustrates the structure of these residual blocks after inserting ${\textnormal{\textit{FxP-QLayer}}}$ for activations quantization.

\begin{figure*}[t!]
\captionsetup[subfloat]{captionskip=0.01pt}
\centering
\subfloat[AlexNet architecture.]{\includegraphics[width=0.97\textwidth,clip]{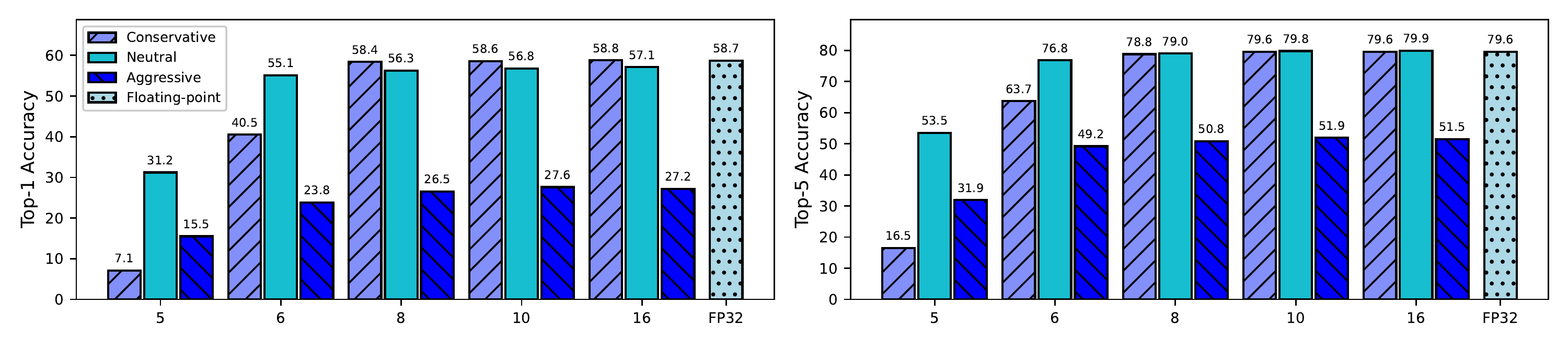}}
\hfil\\[-0.25ex]
\subfloat[VGG-16 architecture.]{\includegraphics[width=0.97\textwidth,clip]{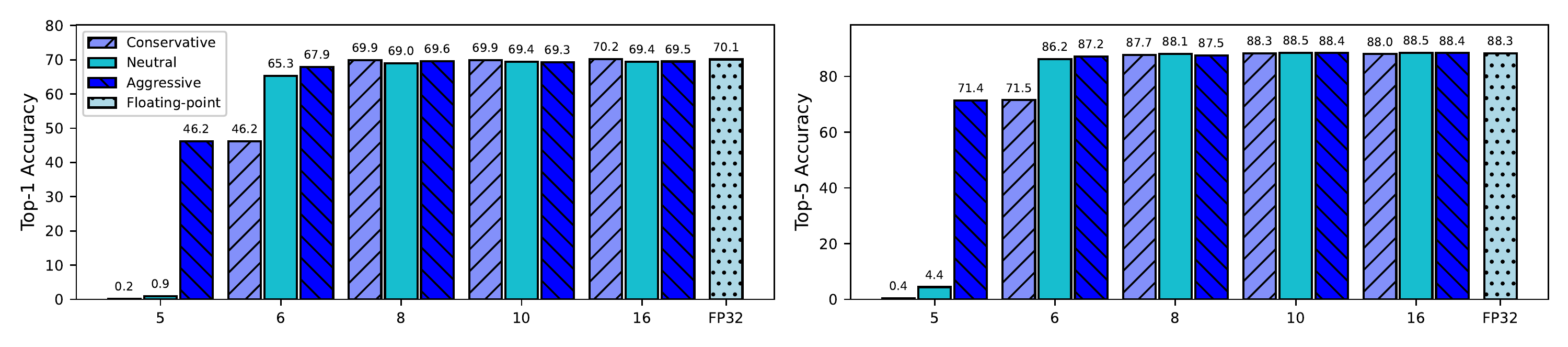}}
\hfil\\[-0.25ex]
\subfloat[ResNet-18 architecture.]{\includegraphics[width=0.97\linewidth,clip]{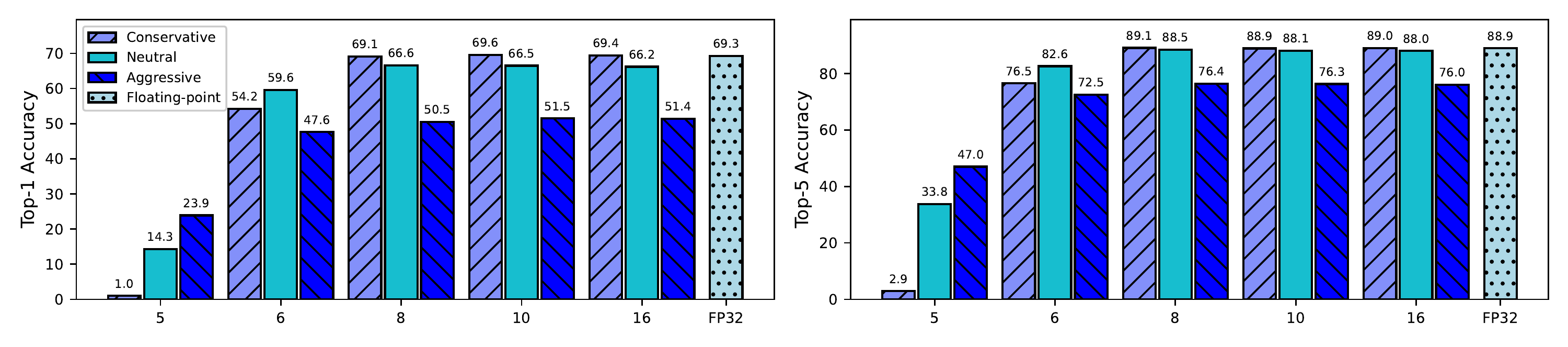}}
\caption[]{The performance of (a) AlexNet, (b) VGG-16, and (c) ResNet-18 architectures after quantization using the initial quantization parameters determined by the rounding strategies for each of the studied bit-precision levels. The FP32 stands for 32-bit floating-point bit-precision level.
}
\label{fig:round_strategies}
\end{figure*}

\subsection{Analysis of Rounding Strategies used to determine the Initial Quantization Parameters} \label{sec:analysis_rounding_strategies}

The design of a mixed low-precision DNN starts with the \textit{preprocessor} that generates initial $\mathrm{QPs}$ for each data-structure based on a given wordlength and round strategy as discussed in Equation~(\ref{eq:initial_quantization_parameters}). Specifically, \textit{preprocessor} adopts three rounding strategies for this purpose; the conservative, neutral, and aggressive strategies. In this section, we evaluate the use of each rounding strategy in the performance of quantized DNNs considering five bit-precision levels, namely, $16$, $10$, $8$, $6$, and $5$ bits. Recall from Section~{\color{DarkBlue}\ref{sec:preprocessor_initial_solution}} that these bit-precision levels are used to quantize $\mathbf{G}_1$ data-structures set defined in Equation~(\ref{eq:level_1_clustering}), except for activations and weights in first and last layers, while all other data-structures are represented in the $32$-bit fixed-point format.

The data ranges, which are used for the quantization of activations, are collected after forwarding the network on the calibration set comprising of $800$ images. Then, the top-1 and top-5 accuracies are measured on $200$ test images. After $5$ runs of each experiment, the average test accuracies for AlexNet, VGG-16, and ResNet-18 architectures are shown in Figure~\ref{fig:round_strategies}. Note that we use very small sets in this experiment to demonstrate the efficacy of the quantization solution proposed by \textit{preprocessor} even with limited data.

First of all, we can observe that the rounding strategy plays a crucial role in determining network’s performance. One can also notice that the conservative strategy achieves better performance in all the considered architectures for $8$-, $10$-, and $16$-bit fixed-point quantization. Focusing on $8$-bit results, it is clear that the bit-precision reduction framework can start the design of mixed low-precision networks with an initial solution that causes less than $0.4$\% accuracy drop. 

Going down to $6$-bit quantized data, the obtained $\mathrm{QPs}$ by the neutral strategy increases the top-1 accuracy gap with those of the second-best strategy to $14.6$\% and $5.4$\% for AlexNet and ResNet-18, respectively. The top-5 accuracy results also show similar behavior where conservative strategy degrades AlexNet and ResNet-18 top-5 accuracies by $13.1$\% and $6.1$\%, respectively. On the other hand, the best results for the $6$-bit initial solution of VGG-16 architecture come from the aggressive strategy as it has a $2.6$\% and $1$\% improvement for the top-1 and top-5 accuracies over neutral strategy, respectively.

For the $5$-bit quantization, we can see that neutral strategy achieves the best top-1/5 results for AlexNet, whereas for VGG-16 and ResNet-18, the aggressive strategy performs better than other strategies. One can also note that the homogeneous quantization of activations and weights to $6$ bits, or even fewer number of bits, without retraining causes a severe information loss. Thus, using $6$ bits as a starting wordlength for the bit-precision reduction framework is not appropriate due to the significant accuracy drop. In subsequent experiments, we use the best strategy found in this section for each architecture and wordlength as \textit{preprocessor} strategy to generate initial $\mathrm{QPs}$ and report the numbers based on that. Note that we also analyzed several methods for rounding quantized data in Equation~\eqref{eq:fp_k_bit_quantization}, such as the ceil, floor, round, and truncate, and found the round method to be the best. For the sake of brevity, we do not show these experiments.

\subsection{Analysis of Optimization Strategies for Quantization Parameters}

In this section, we demonstrate the effectiveness of the proposed post-training self-distillation and network prediction errors function, $\mathcal{F}$, defined in Equation~(\ref{eq:quantization_parameters_optimization}), in optimizing $\mathrm{QPs}$. In order to do that, we use Algorithm~\ref{Algorithm:QFO} as a standalone framework to optimize the $\mathrm{QPs}$ regardless of the optimization function mentioned in lines $5$ and $14$. We set the search space limit, $K$ in line $8$, to $5$ so that the $10$-nearest $\mathrm{QPs}$ neighbors are checked to make sure we get the one with the minimum \textit{cost}. We compare $\mathcal{F}$ with three popular optimization functions; top-1/5 error-rate, $L1$ error, and $L2$ error. The top-1/5 error-rate is equal to one minus the top-1/5 accuracy. The $L2$ error, denoted by $\mathcal{L}$ in this paper, is calculated as stated in Equation~(\ref{eq:DSB_QE}). On the other hand, the $L1$ error, also named mean absolute error, is computed as

\begin{equation} \label{eq:MAE}
 L1\left(X, \widetilde{X}\right) = \frac{1}{N} \cdot \sum_{i = 1}^{N} \left|X_i - \widetilde{X}_i\right|
\end{equation}
\\
where $\widetilde{X}$ is the low-precision representation of $X$ that is computed as mentioned in Equation~(\ref{eq:fixed_point}), and $N$ is the size of $X$. We use VGG-16 as a test case to compare the performance of these four optimization functions. Additionally, we use $50$ images to calibrate the $\mathrm{QPs}$ with a given optimization function and then test the classification accuracy of the quantized fixed-point network on $5,000$ images. The size of the datasets is defined with the aim of measuring the effectiveness of the proposed optimization function even when given a very small sample set for calibration.

In the experiments, the \textit{preprocessor} generates the initial $\mathrm{QPs}$, referred to as the range-based $\mathrm{QPs}$, to homogeneously quantize $\mathbf{G}_1$ data-structures into $16$-, $8$-, and $6$-bit fixed-point numbers, whereas $\mathbf{G}_2$ data-structures are quantized to $32$-bit fixed-point numbers. It is noteworthy that $\mathbf{G}_1$ denotes the weights of VGG-16 model as well as the activations of all layers, and on the other hand, $\mathbf{G}_2$ denotes the biases of VGG-16 model. Thereafter, the \textit{forward optimizer} is used to optimize range-based $\mathrm{QPs}$ with a given optimization function. After running each experiment on $10$ randomly selected datasets, the average calibration and test accuracies are reported in Table~\ref{tab:qf_method_anaysis}.

\begin{table}[tb!]
\begin{center}
\captionsetup{justification=centering}
\caption{Homogeneous data-structure quantization using different optimization methods for designing 16-, 8-, and 6-bit fixed-point network of VGG-16 architecture.}
\label{tab:qf_method_anaysis}
\scriptsize
\renewcommand{\arraystretch}{1.4}
\setlength\tabcolsep{1.3pt}
\begin{tabular}{| l | l | l | c | c | c | c | c |}
\hline
\multicolumn{3}{| c |}{\multirow{2}{*}{\textbf{Configuration}}} & \multirow{2}{*}{\makecell{\textbf{Top-1/5} \\ \textbf{Error}}} & \multirow{2}{*}{\makecell{\textbf{Range} \\ \textbf{Eqn.~(\ref{eq:initial_quantization_parameters})}}} & \multirow{2}{*}{\makecell{\textbf{$L1$} \\ \textbf{Eqn.~(\ref{eq:MAE})}}} & \multirow{2}{*}{\makecell{$\mathcal{L}$ \\ \textbf{Eqn.~(\ref{eq:DSB_QE})}}} & \multirow{2}{*}{\makecell{$\mathcal{F}$ \\ \textbf{Eqn.~(\ref{eq:quantization_parameters_optimization})}}} \\ 
\multicolumn{3}{| c |}{} & & & & & \\
\hline\hline
\multirow{4}{*}{\makecell{\textbf{16-bit} \\ \textbf{Fixed-Point}}} & \multirow{2}{*}{\textbf{\,Calib\:}} & \textbf{\,Top-1\:}	& 74.29\% & 70.86\%	& 71.50\% & 70.86\%	& 70.86\% \\ 
\cdashline{3-8}[2.8pt/1pt]
 & & \textbf{\,Top-5}	& 90.86\% & 91.14\%	& 91.75\% & 91.14\%	& 91.14\% \\ 
 \cdashline{2-8}[5.0pt/1.2pt]
 & \multirow{2}{*}{\textbf{\,Test}} & \textbf{\,Top-1}	& 66.51\% & \textbf{67.83\%}	& 67.68\% & \textbf{67.83\%}	& \textbf{67.83\%} \\ 
\cdashline{3-8}[2.8pt/1pt]
 & & \textbf{\,Top-5}	& 88.09\% & \textbf{88.91\%}	& 88.80\% & \textbf{88.91\%}	& \textbf{88.91\%} \\ 
\hline
\multirow{4}{*}{\makecell{\textbf{8-bit} \\ \textbf{Fixed-Point}}} & \multirow{2}{*}{\textbf{\,Calib}} & \textbf{\,Top-1}	& 75.43\% & 70.57\%	& 55.43\% & 71.14\%	& 69.71\% \\ 
\cdashline{3-8}[2.8pt/1pt]
 & & \textbf{\,Top-5}	& 90.00\% & 91.14\%	& 80.00\% & 91.14\%	& 91.14\% \\ 
 \cdashline{2-8}[5.0pt/1.2pt]
 & \multirow{2}{*}{\textbf{\,Test}} & \textbf{\,Top-1}	& 65.14\% & 67.66\%	& 55.74\% & 67.60\%	& \textbf{67.83\%} \\ 
\cdashline{3-8}[2.8pt/1pt]
 & & \textbf{\,Top-5}	& 86.77\% & 88.86\%	& 79.23\% & 89.14\%	& \textbf{88.71\%} \\ 
\hline
\multirow{4}{*}{\makecell{\textbf{6-bit} \\ \textbf{Fixed-Point}}} & \multirow{2}{*}{\textbf{\,Calib}} & \textbf{\,Top-1}	& 74.86\% & 66.29\%	& 35.14\% & 67.43\%	& 68.00\% \\ 
\cdashline{3-8}[2.8pt/1pt]
 & & \textbf{\,Top-5}	& 87.14\% & 90.00\%	& 57.71\% & 87.14\%	& 88.86\% \\ 
 \cdashline{2-8}[5.0pt/1.2pt]
 & \multirow{2}{*}{\textbf{\,Test}} & \textbf{\,Top-1}	& 62.17\% & 66.40\%	& 33.20\% & 64.69\%	& \textbf{66.83\%} \\ 
\cdashline{3-8}[2.8pt/1pt]
 & & \textbf{\,Top-5}	& 84.11\% & 87.63\%	& 58.23\% & 86.26\%	& \textbf{88.06\%} \\ 
\hline
\end{tabular}
\end{center}\vspace{-1.9mm}
\end{table}

Focusing on $\mathcal{L}$ and the proposed $\mathcal{F}$ optimization functions, we see that they both achieve similar performance results for $16$-bit quantization level. With $8$-bit data quantization, $\mathcal{F}$ performs slightly better than $\mathcal{L}$ top-1 accuracy on test datasets. Going down to $6$-bit, it can be seen that the quantized networks achieves top-1 accuracy of $66.83$\% and top-5 accuracy of $88.06$\% on test datasets by using our $\mathcal{F}$ optimization function. Comparatively, $\mathcal{L}$ optimization function achieves $64.69$\% and $86.26$\% in top-1 and top-5 accuracies on test datasets, respectively. In fact, $\mathcal{L}$ optimization function brings $1.71$\% and $1.37$\% decrease in initial solution top-1 and top-5 accuracies, respectively. Apparently, range-based quantized networks are better than low-precision networks optimized with $\mathcal{L}$ optimization function.

\begin{figure*}[b!]
\centering
\subfloat[$\textnormal{conv1\_1, }S(1, 5)$.]{\includegraphics[width=0.315\textwidth,trim=0.4cm 0.4cm 0.36cm 0.7cm, clip]{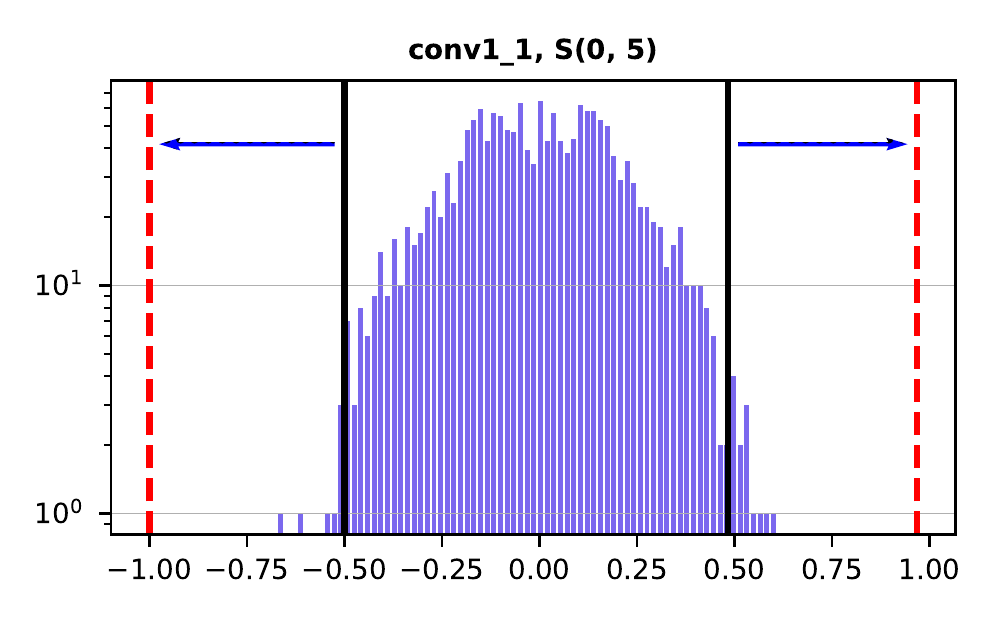}}
\hspace{1.30mm}
\subfloat[$\textnormal{conv3\_1, }S(-1, 7)$.]{\includegraphics[width=0.315\textwidth,trim=0.4cm 0.4cm 0.36cm 0.7cm, clip]{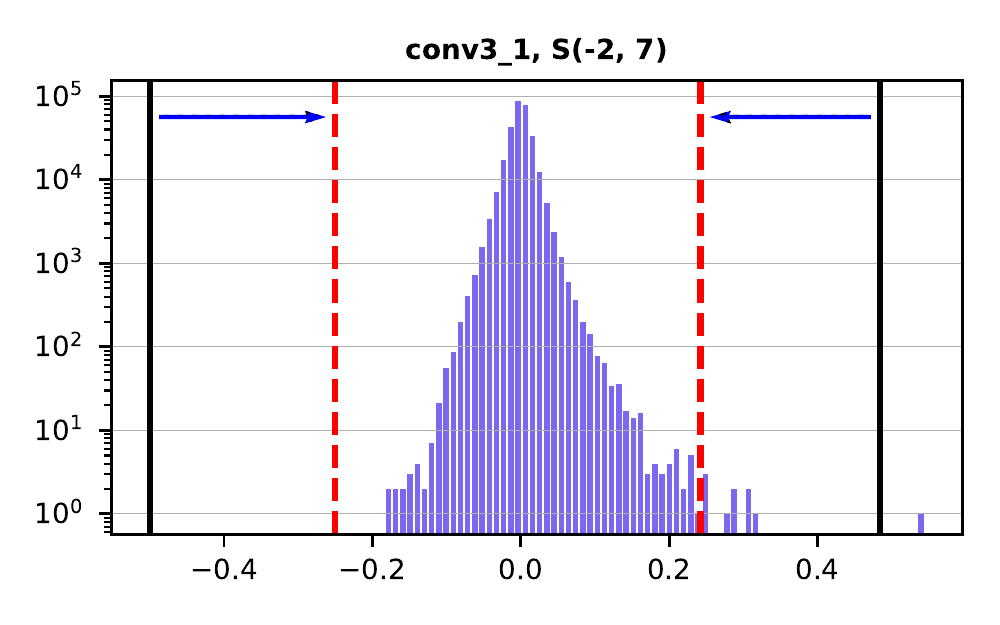}}
\hspace{1.30mm}
\subfloat[$\textnormal{conv4\_2, }S(-2, 8)$.]{\includegraphics[width=0.315\textwidth,trim=0.4cm 0.4cm 0.36cm 0.7cm, clip]{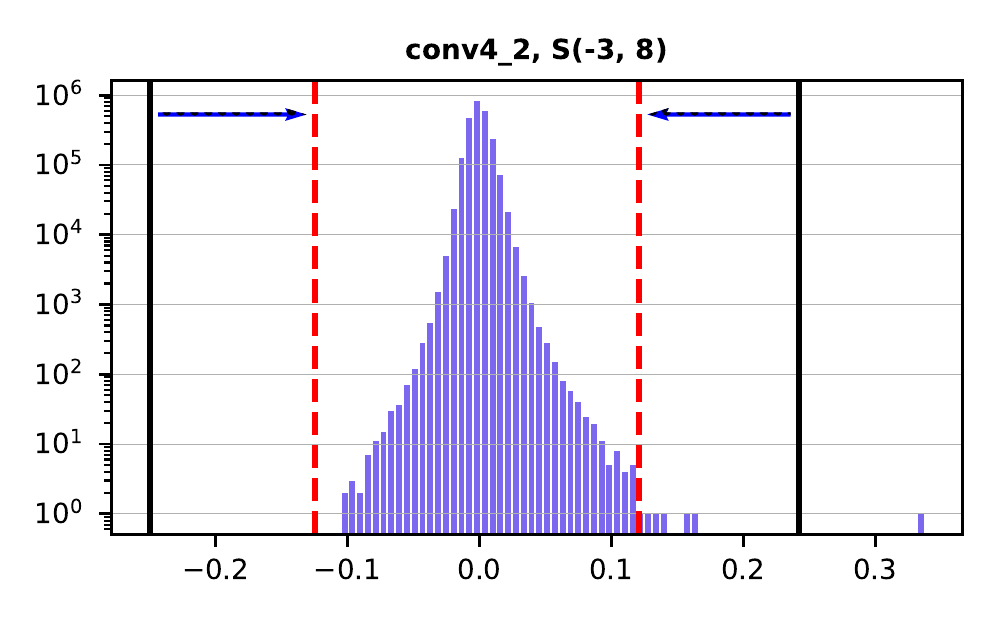}}
\hfil\\[-0.5ex]
\subfloat[$\textnormal{conv4\_3, }S(-2, 8)$.]{\includegraphics[width=0.315\textwidth,trim=0.4cm 0.4cm 0.36cm 0.7cm, clip]{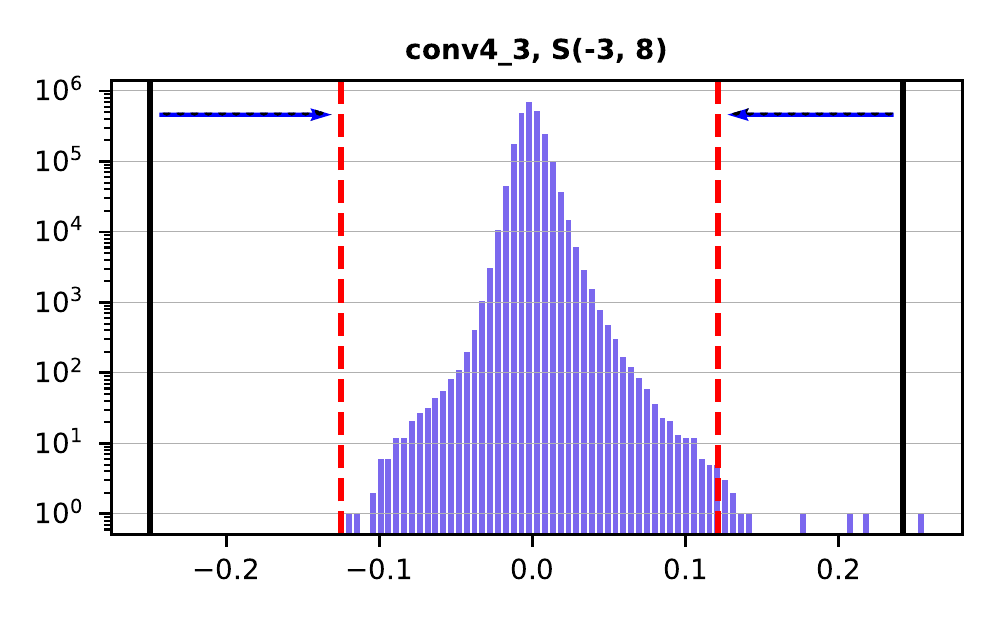}}
\hspace{1.30mm}
\subfloat[$\textnormal{conv5\_2, }S(-3, 9)$.]{\includegraphics[width=0.315\textwidth,trim=0.4cm 0.4cm 0.36cm 0.7cm, clip]{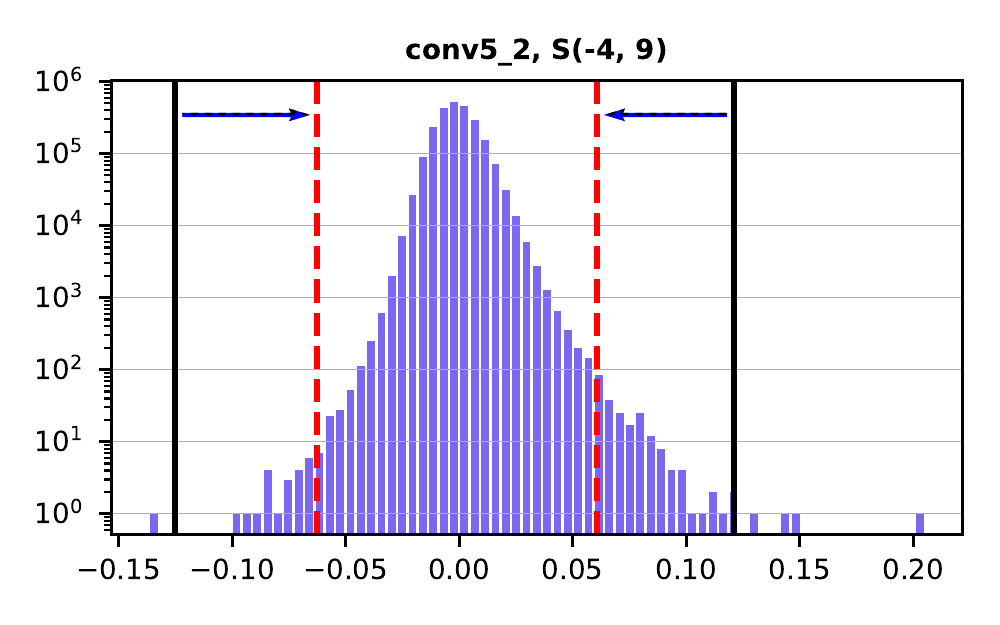}}
\hspace{1.30mm}
\subfloat[$\textnormal{conv5\_3, }S(-3, 9)$.]{\includegraphics[width=0.315\textwidth,trim=0.4cm 0.4cm 0.36cm 0.7cm, clip]{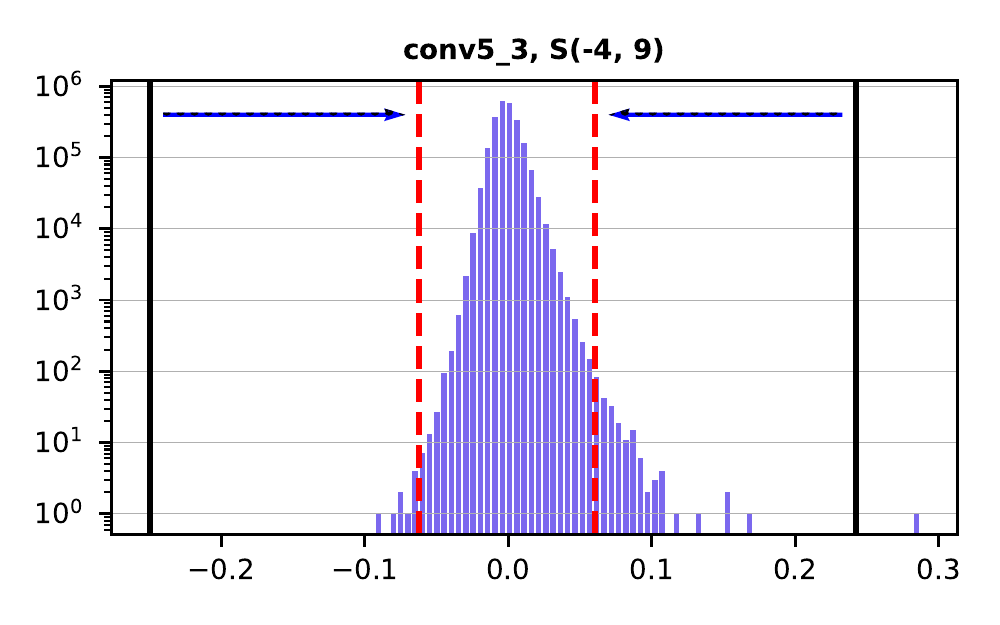}}
\caption{Weight distributions of VGG-16 layers that changed their clipping thresholds by non-zero integer amount in log-domain. Initial and calibrated clipping thresholds are indicated as black-solid and red-dashed lines, respectively. The optimized fixed-point representation for each of these layers are also given.}
\label{fig:precision_range_tradeoff}
\end{figure*}

As $\mathcal{F}$ achieves better accuracy than optimization functions that directly and independently seek to minimize the quantization error for each layer and data-structure, this suggests that simply fitting a quantizer to its corresponding data distribution may not be optimal for task performance. As seen, $\mathcal{F}$ optimization function can quantize state-of-the-art network to $8$-bit fixed-point format with negligible accuracy loss. Furthermore, the $6$-bit data quantization using the proposed optimization function brings within $1$\% accuracy drop on test datasets. We believe this is due to the fact that our optimization function formulation is able to balance range-resolution trade-off effectively.

In Figure~\ref{fig:precision_range_tradeoff}, we analyze the optimized clipping thresholds for a few $6$-bit quantized weight data-structures in VGG-16 architecture, highlighting the importance of range-resolution trade-off. Remember that the dynamic range of a $6$-bit number in signed 2's complement fixed-point representation is $[-32 \times 2^{-\beta}, 31 \times 2^{-\beta}]$, where $\beta$ denotes the fractional length. As seen with conv1\_1 layer, the calibrated clipping thresholds move out from their initialized values by $1$ integer bin in the log-domain, favoring dynamic range over resolution. For other layers, the clipping thresholds move in from their initialized values by up to $2$ integer bins in the log-domain, which in turn leads to a higher resolution and, therefore, smaller round-off error upon quantizing the data within the clipping thresholds. From these results, we argue that for better accuracy of quantized models, weights in shallow layers should focus on saturating their range. On the other hand, as we go deeper into the network, layers should focus more on increasing their resolution.

For top-1/5 error-rate when used as an optimization function, it achieves the best accuracy results on calibration sets for all considered quantization levels, but its low-precision networks suffer significant accuracy drop when tested with unseen samples. Hence, the generalization ability of $\mathcal{F}$-based quantized networks is better than that of the error-rate-based. For $L1$ optimization function, $16$-bit data quantization incurs almost no accuracy drop on test sets. Going down to $8$- and $6$-bit fixed-point weights and activations, the quantized networks with $L1$ optimization function crash. The reason for this is due to the low sensitivity of $L1$ optimization to outliers which have a potentially important impact on the underlying dot-product operation. Hence, $L1$ optimization is not a robust error measure for finding the best design for low-precision networks. To summarize, the experimental results show that the proposed optimization function achieves better performance than other tested optimization functions in all of the three precision cases.

\begin{figure*}[t]
\centering
\subfloat{\includegraphics[width=0.335\textwidth,trim=0.7cm 0.8cm 1.4cm 1.9cm, clip]{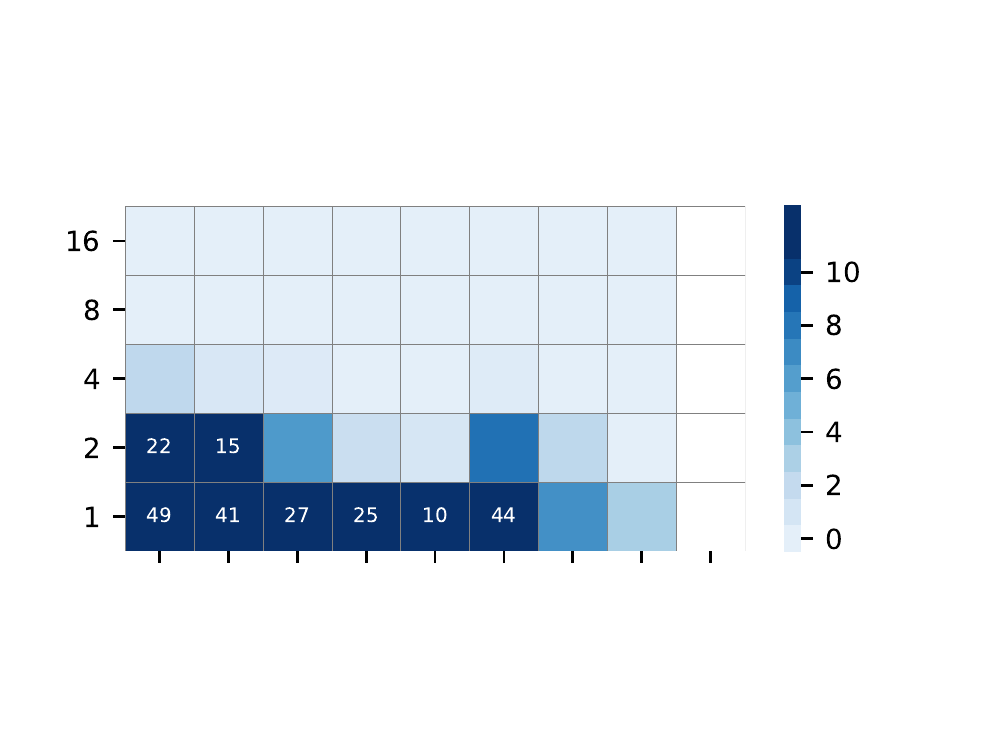}}
\hspace{0.55mm}
\subfloat{\includegraphics[width=0.615\textwidth, trim=1.9cm 0.6cm 3.6cm 1.88cm, clip]{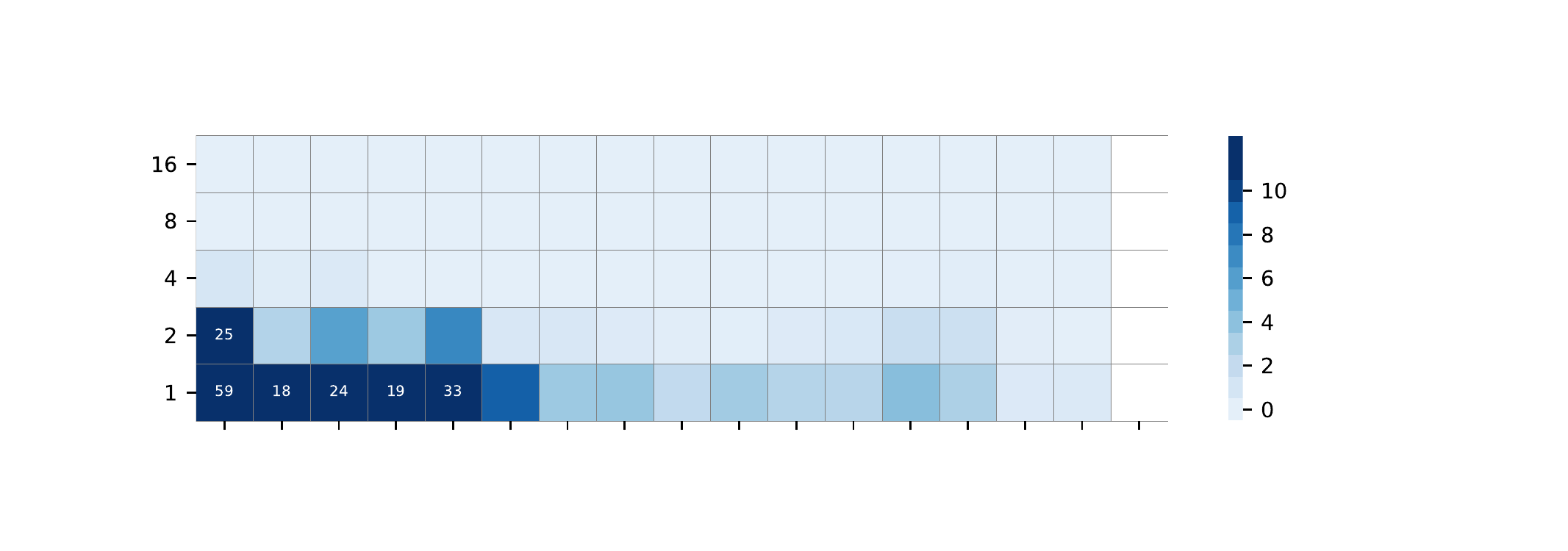}}
\hfil\\[-5.5ex]
\setcounter{subfigure}{0}%
\subfloat[AlexNet architecture.]{\includegraphics[width=0.335\textwidth,trim=0.7cm 0.8cm 1.4cm 1.9cm, clip]{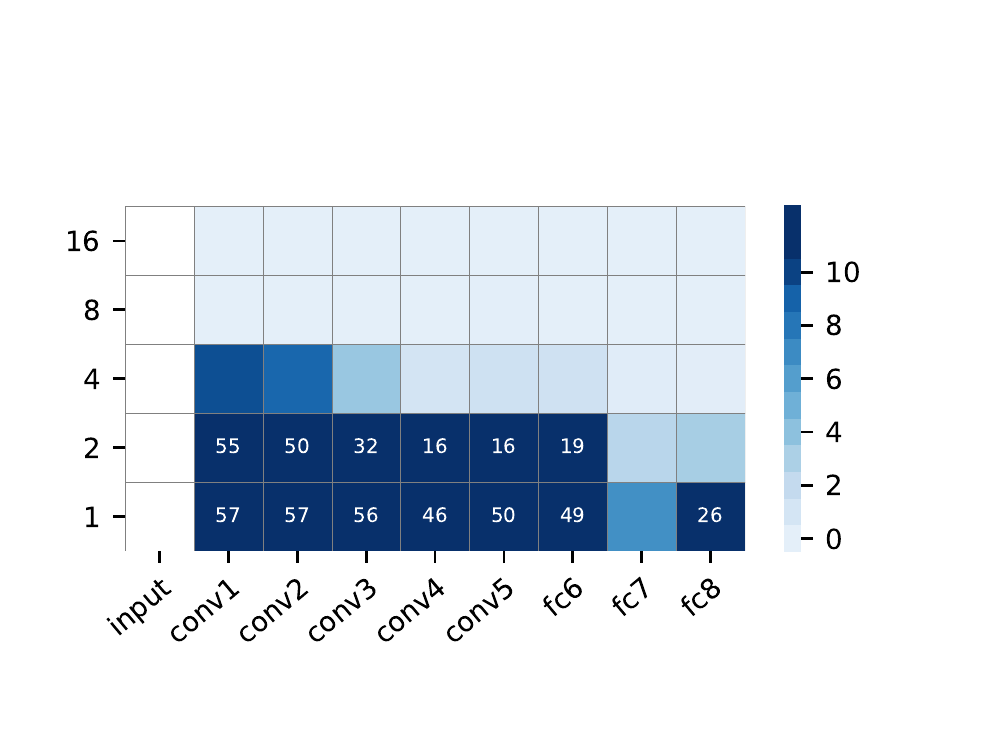}}
\hspace{0.55mm}
\subfloat[VGG-16 architecture.]{\includegraphics[width=0.615\textwidth, trim=1.9cm 0.6cm 3.6cm 1.88cm, clip]{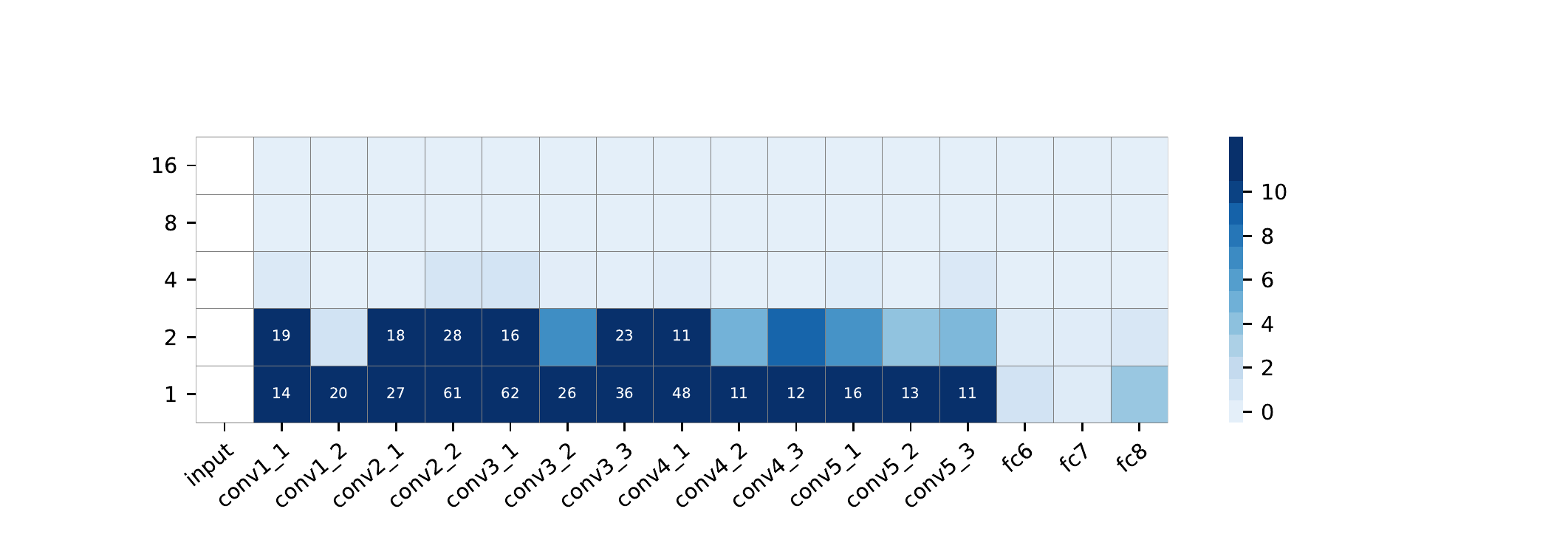}}
\hfil
\subfloat{\includegraphics[width=0.97\textwidth, trim=2.485cm 2.5cm 4.6cm 2.65cm, clip]{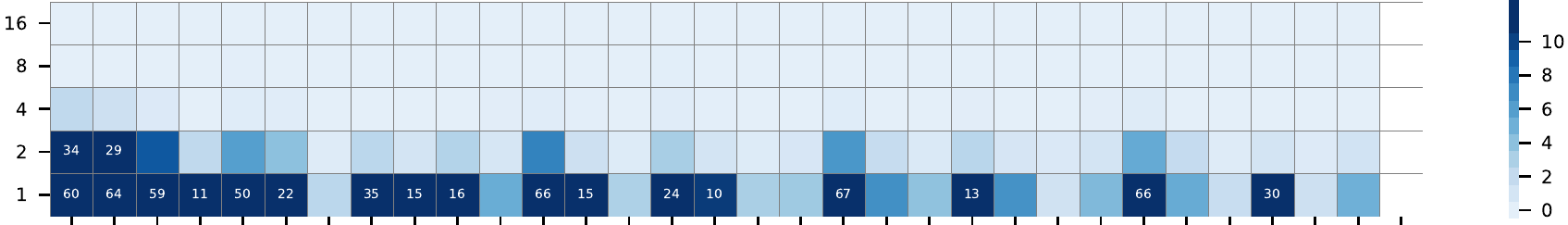}}
\hfil\\[1ex]
\setcounter{subfigure}{2}%
\subfloat[ResNet-18 architecture.]{\includegraphics[width=0.97\textwidth, trim=2.485cm 1.55cm 4.6cm 2.65cm, clip]{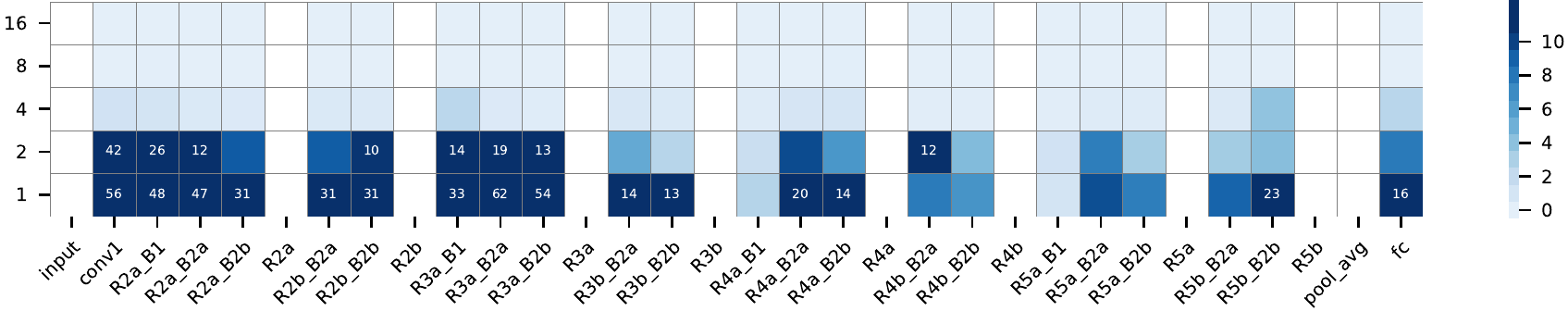}}
\captionsetup{singlelinecheck=off}
\caption[]{Data-structures robustness to quantization for \begin{enumerate*}[label=(\alph*)] \item AlexNet, \item VGG-16, and \item ResNet-18 \end{enumerate*} architectures. For each architecture, the \textit{top} figure represents the activations robustness to quantization, while the \textit{bottom} figure is designated for weights robustness. The rows of each figure correspond to the quantization bit-precision level, while the columns designate quantization robustness at different network layers. The white column means that the data-structure in that layer is not quantized because it is not needed. Additionally, we show the amount of degradation for configurations that causes the accuracy to drop by 10\% or more.}
\label{fig:sensitivity}
\end{figure*}

\subsection{Data-Structures Sensitivity to Quantization} \label{sec:sensitivity_results}

In this experiment, we investigate the role each layer plays in forming a good predictor. To this end, we designed a test that shows the robustness of each layer to quantization. The test differentiates between activations and weights robustness. For each layer, we quantize its weights and output activations separately, while keeping the data-structures for other layers intact. After optimizing the $\mathrm{QPs}$, the performance of the network is evaluated by measuring the top-1 error-rate. To measure the robustness of activations/weights in the $\ell$-th layer, we compare the measured top-1 error-rate to that of the original full-precision network. When a network exhibits a significant increase in top-1 error-rate after quantizing $\ell$-th layer activations/weights, we say that the $\ell$-th layer activation/weight data-structure is sensitive to quantization, and otherwise, it is called robust to quantization. 

Each data-structure is quantized into five levels; $16\textnormal{-,}$ $8\textnormal{-,}$ $4\textnormal{-,}$ $2\textnormal{-,}$ and $1$-bit fixed-point format. A dataset of size $5,000$ images is used for $\mathrm{QPs}$ calibration. Thereafter, top-1 error-rate is measured on a test datset containing another $5,000$ images. Observing the robustness results, as shown in Figure~\ref{fig:sensitivity}, several conclusions can be drawn; \begin{enumerate*}[label=(\roman*)] \item overall, quantizing different DNN data-structures causes various levels of accuracy drop, \item data-structures in deeper layers are more robust to quantization than the ones in shallower layers, and \item the bit-precision level can be reduced up to $8$ bits without causing noticeable performance degradation\end{enumerate*}.

For AlexNet, we can see that almost all layers are sensitive to rigorous quantization with activations being slightly less sensitive than weights. A plausible explanation for this could be attributed to the size of AlexNet architecture. When the network is small, all layers are vigilant participants in forming a good predictor. As the network size increases, which is the case for VGG-16, deeper layers become more robust to quantization. The reason for this is due to more abstraction occurring in deeper layers. More interestingly, the results show that VGG-16 ${\textnormal{\textit{FC}}}$ layers, which contributes to $\sim \!\!\! 89$\% of VGG-16 model size, can use aggressively quantized weights. 

With regard to ResNet-18 architecture, we found that the sensitivity patterns are not as easily pronounced as those of AlexNet and VGG-16. However, focusing on the weights of ResNet-18, we can interpret the ${\textnormal{\textit{CONV}}}$ layers in the residual connections of each stage as a sub-network with sensitivity characteristics roughly similar to AlexNet and VGG-16. Furthermore, by comparing the results of the ${\textnormal{\textit{CONV}}}$ layer in the downsample skip connection, i.e., weights in \textit{R$\star$a\_B1}, at the different residual stages, we can see that the robustness increases for the deeper layers in a similar fashion to what was observed in the weights of VGG-16. Concerning activations sensitivity, unlike AlexNet and VGG-16 architectures that place sensitive layers at the bottom of the network, ResNet-18 distributes them across the network. More precisely, we can notice that the activation data-structure of the last ${\textnormal{\textit{CONV}}}$ layer in each residual connection is more sensitive to quantization than that of the other layers. This could be potentially attributed to the higher dynamic range of activations in these layers.

\begin{figure*}[t!]
\vspace{-0.9mm}
\captionsetup[subfloat]{captionskip=0.5pt}
\centering
\subfloat[AlexNet architecture.]{\includegraphics[width=0.315\textwidth, trim=0.35cm -0.225cm 0.35cm 0.35cm, clip]{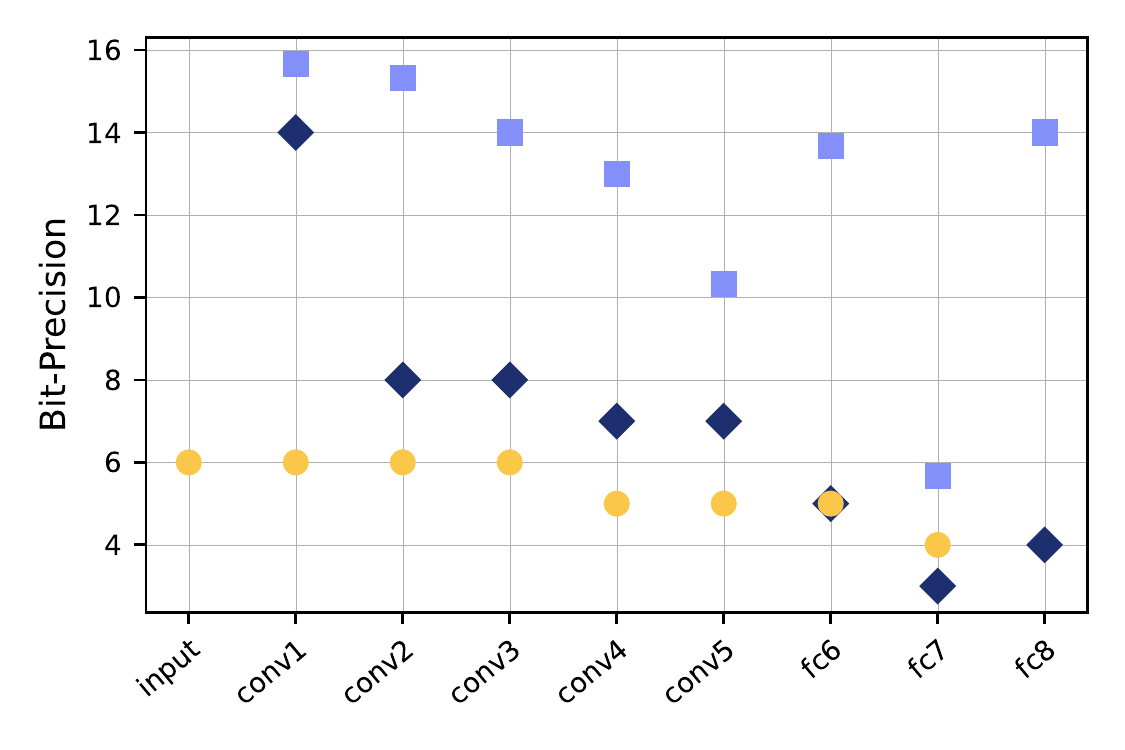}}
\hspace{2mm}
\subfloat[VGG-16 architecture.]{\includegraphics[width=0.628\textwidth, trim=0.35cm 0.0cm 0.35cm 0.35cm, clip]{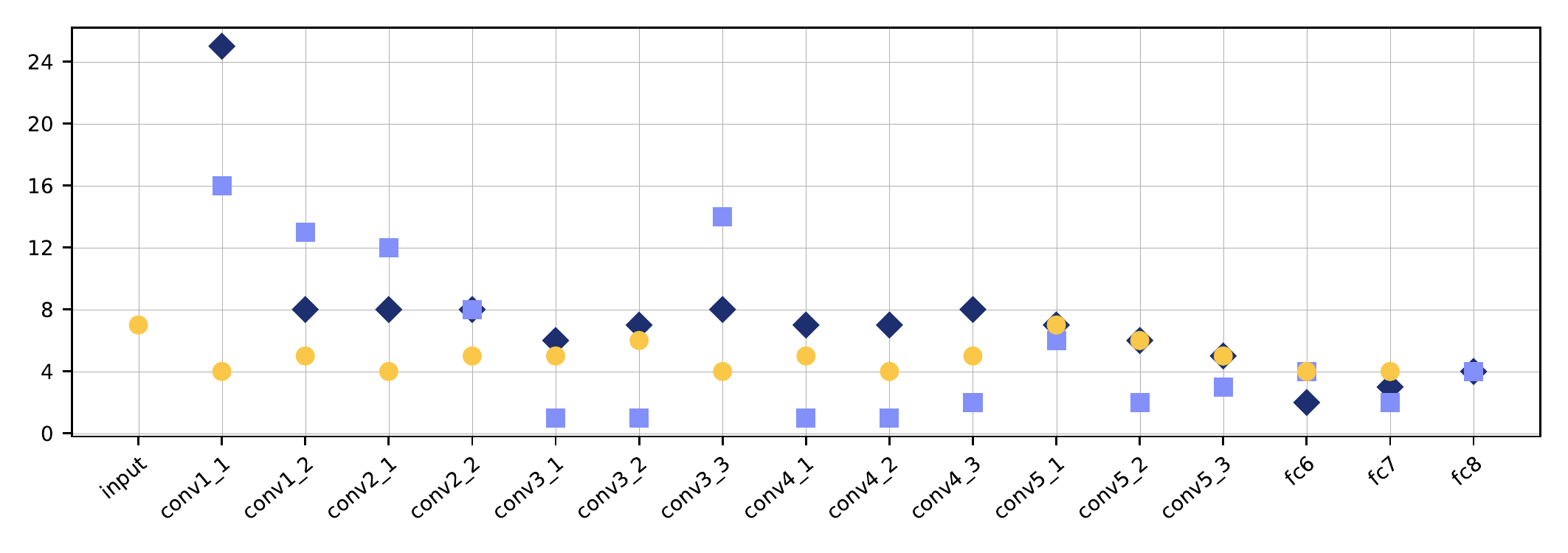}}
\hfil\\[-2.5ex]
\subfloat[ResNet-18 architecture.]{\includegraphics[width=0.81\linewidth, trim=2.55cm 0.4cm 2.4cm 1.0cm, clip]{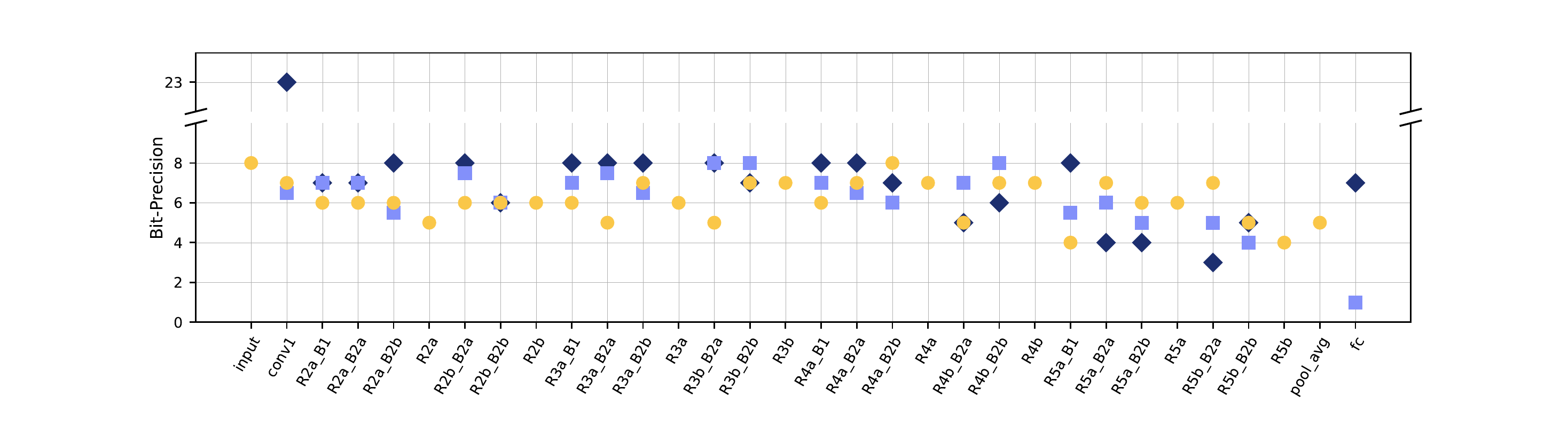}}
\subfloat{\includegraphics[width=0.115\linewidth, trim=16.95cm 0.55cm 0cm 0cm,clip]{imgs/data_structure_size_legend}}
\caption[]{Quantization wordlengths allocated to the data-structures of each layer for \begin{enumerate*}[label=(\alph*)] \item AlexNet, \item VGG-16, and \item ResNet-18 \end{enumerate*} architectures. For Biases/Others data-structures, we show the average allocated wordlength in each layer in case of more than one parameter exist in that layer.}
\label{fig:bit_precision}
\end{figure*}

According to the results and findings that we observed in this section, the layers in a DNN should not be treated in a monolithic way. Furthermore, we argue that better preservation of sensitive data-structures information, through higher wordlength allocation, boosts the performance of a quantized DNN evidently. At the same time, it allows for further compression in deeper layers which are likely to constitute the highest ratio of model size as seen with VGG-16. Therefore, in the proposed framework, we check the sensitivity of the data-structure to quantization, line $24$ in Algorithm~\ref{Algorithm:QFO}, and accordingly either accept or reject the reduction, lines $19-24$ in Algorithm~\ref{Algorithm:BPR_2} - Part $2$. Also, we can clearly see that the last ${\textnormal{\textit{FC}}}$ layer in each architecture is less robust to quantization than the ${\textnormal{\textit{FC}}}$ layers preceding it, if any. Thus, the \textit{preprocessor} generates an initial solution where the weights and input activations of the last ${\textnormal{\textit{FC}}}$ layer are at $32$-bit fixed-point bit-precision level.

\subsection{Mixed Low-Precision Quantization} 

Unlike previous sections where only \textit{preprocessor} or \textit{preprocessor} and \textit{forward optimizer} units were used for evaluation, this section employs all the components of ${\textnormal{\textbf{FxP-QNet}}}$ to design and evaluate a mixed low-precision DNN. We should emphasize here that our work focuses on a hardware-friendly design of quantized DNNs that maps well on generic fixed-point hardware. We use the full ILSVRC-2012 ImageNet validation dataset to evaluate the performance of quantized DNNs, while a sample set of $1,000$ images is used to calibrate the quantization parameters. The top-1/5 accuracy degradation threshold is set to $1$\%, $1$\%, and $2$\% for AlexNet, VGG-16, and ResNet-18, respectively. We assign a higher accuracy degradation threshold to ResNet-18 due to its complex structure and lighter model with significantly less parameters compared to AlexNet and VGG-16.

\subsubsection{Wordlength Allocation and Network Performance} \label{sec:fxp_qnet_wordlength_allocation}

In Figure~\ref{fig:bit_precision}, we visualize the allocated wordlength for the data-structures of each layer in AlexNet, VGG-16, and ResNet-18 architectures. As can be seen, ${\textnormal{\textbf{FxP-QNet}}}$ shows a spectrum of varying wordlength allocations to the layers. In all cases, there is significant heterogeneity in the wordlength assignment. From the results, we can also notice that the number of bits allocated to each data-structure is consistent with our findings in the sensitivity analysis. Thus, we conclude that ${\textnormal{\textbf{FxP-QNet}}}$ succeeds in distinguishing different layers and their varying robustness to quantization while choosing their respective wordlength. 

Furthermore, we can see that all MAC operations are performed on a maximum of $8$-bit numbers, except for the first ${\textnormal{\textit{CONV}}}$ layer. We believe that assigning higher bits to the weights of the first layer allows for more compression and higher performance for the quantized DNNs. Given that the ${\textnormal{\textit{FC}}}$-6 layer performs a $102.76$ Mega floating-point multiply operation, with $2$-bit integer weights, it is now possible to replace these expensive operations with simple add-subtract operations resulting in a significant improvement in computation and energy efficiency.

\begin{table*}[t!]
 \centering
 \captionsetup{justification=centering}
 \caption{Quantization results of AlexNet, VGG-16, and ResNet-18 on ImageNet validation dataset. The baseline network performs operations with 32-bit floating-point type. In comparison, the quantized DNNs perform operations with integer arithmetic. Furthermore, we compare the performance of mixed-precision networks designed by ${\textnormal{\textbf{FxP-QNet}}}$ with the networks that assign the same number of bits to all layers (uniform quantizer).}\vspace{-0.5mm}
 \setlength\tabcolsep{1.5pt}
 \renewcommand{\arraystretch}{1.4}
 \begin{tabular}{| l || l || >{\centering}p{0.72in} | >{\centering}p{0.72in} | >{\centering}p{0.72in} || c | c |} 
 \hline
 \multicolumn{1}{| c||}{\multirow{2}{*}{\textbf{Architecture}}} & \multicolumn{1}{ c||}{\multirow{2}{*}{\textbf{Benchmark}}} & \multicolumn{3}{c||}{\textbf{Bit-Precision}} & \multicolumn{2}{c|}{\textbf{Accuracy}}\\ 
 \cline{3-7}
 &  &   \textbf{Activations} & \textbf{Weights} & \textbf{Biases/Others} & \textbf{\;\;\;\;Top-1\;\;\;\;} & \textbf{\;\;\;\;Top-5\;\;\;\;}\\ 
 \hline\hline
 \multirow{4}{*}{\:AlexNet} & Baseline (Floating-Point) & 32 & 32 & 32 & 57.20\% & 79.68\% \\ 
 \cdashline{2-7}
  & Uniform (Preprocessor) & \multicolumn{1}{ c |}{\multirow{2}{*}{6}} & \multicolumn{1}{ c |}{\multirow{2}{*}{5}} & \multicolumn{1}{ c ||}{\multirow{2}{*}{11}} & 28.95\% & 50.99\% \\ 
 \cdashline{2-2} \cdashline{6-7}
  & Uniform (Forward Optimizer) &  &  &  & 50.64\% & 74.13\% \\ 
 \cdashline{2-7}
  & ${\textnormal{\textbf{FxP-QNet}}}$ (Mixed-Precision) & 5.85 & 4.48 & 10.33 & 56.25\% & 78.80\% \\ 
 \hline
 \multirow{4}{*}{\:VGG-16} & Baseline (Floating-Point) & 32 & 32 & 32 & 68.35\% & 88.44\% \\ 
 \cdashline{2-7}
  & Uniform (Preprocessor) & \multicolumn{1}{ c |}{\multirow{2}{*}{5}} & \multicolumn{1}{ c |}{\multirow{2}{*}{3}} & \multicolumn{1}{ c ||}{\multirow{2}{*}{4}} & 0.11\% & 0.54\% \\ 
 \cdashline{2-2} \cdashline{6-7}
  & Uniform (Forward Optimizer) &  &  &  & 18.41\% & 37.68\% \\ 
 \cdashline{2-7}
  & ${\textnormal{\textbf{FxP-QNet}}}$ (Mixed-Precision) & 4.66 & 2.68 & 3.34 & 67.40\% & 87.50\% \\ 
 \hline
 \multirow{4}{*}{\:ResNet-18} & Baseline (Floating-Point) & 32 & 32 & 32 & 66.62\% & 87.49\% \\ 
 \cdashline{2-7}
  & Uniform (Preprocessor) & \multicolumn{1}{ c |}{\multirow{2}{*}{7}} & \multicolumn{1}{ c |}{\multirow{2}{*}{5}} & \multicolumn{1}{ c ||}{\multirow{2}{*}{6}} & 49.93\% & 74.85\% \\ 
 \cdashline{2-2} \cdashline{6-7}
  & Uniform (Forward Optimizer) & & & & 59.76\% & 83.00\% \\ 
 \cdashline{2-7}
  & ${\textnormal{\textbf{FxP-QNet}}}$ (Mixed-Precision) & 6.32 & 4.81 & 5.52 & 64.63\% & 86.16\% \\ 
 \hline
 \end{tabular}
 \label{tab:mixed_v_uniform}
\end{table*}

We can also observe that ${\textnormal{\textbf{FxP-QNet}}}$ assigns less bits to the weights of ${\textnormal{\textit{FC}}}$ layers than that of ${\textnormal{\textit{CONV}}}$ layers in AlexNet and VGG-16. Intuitively, this is because the size as well as the robustness of the former is much more than the latter. Remember that big data-structures, in terms of memory, are visited early for quantization, and at that time, the \textit{network designer} has a relatively large space to work on quantization before the accuracy degradation threshold is violated. Comparatively, the weights of the ${\textnormal{\textit{FC}}}$ layer in ResNet-18 are allocated with $3$ more bits, on average, than the weights of ${\textnormal{\textit{CONV}}}$ layers in the residual connections of the last stage due to the same reasons.

The results also show that the first and last layers can be quantized to lower bit-precision levels without having a drastic effect on the accuracy. For instance, the proposed framework was able to reduce the bit-precision level of the first layer weights and activations of AlexNet by $18$ and $26$ bits from their initial $32$ bits level while degrading the top-1 accuracy only by $0.06$\% and $0.08$\%, respectively. With regards to the last layer of AlexNet, the proposed framework successfully reduced the wordlength to $4$ bits for both the activations and the weights. This reduction was accompanied by a $0.12$\% and $0.02$\% increase in top-1 and top-5 error-rates, respectively.

A summary of the average bit-precision level assigned to each data-structure as well as the classification accuracy achieved for the evaluated architectures is presented in Table~\ref{tab:mixed_v_uniform}. Because the post-training quantization works available in the literature for integer-only deployment do not exceed the $8$-bit limit, we make a comparison with the accuracy of the full-precision baselines. Accuracy of $32$-bit floating-point baselines is reported as validated on our end. Additionally, we compare the performance of the mixed-precision networks obtained by ${\textnormal{\textbf{FxP-QNet}}}$ with the uniform-precision networks designed by the \textit{preprocessor} and \textit{forward optimizer}.

More precisely, the \textit{preprocessor} is used to design a uniform network, in which the bit-precision level of each data-structure is set to the average bit-precision level obtained by ${\textnormal{\textbf{FxP-QNet}}}$. Thereafter, we employ the \textit{forward optimizer} to further improve the uniform networks designed by the \textit{preprocessor}. Note that we set the search space limit in Algorithm~\ref{Algorithm:QFO} to $1$ in order to speed up the optimization process. This is because we found that searching only adjacent $\mathrm{QPs}$ with expanding the search space in the direction of minimum \textit{cost} insures global optimum $\mathrm{QPs}$ when $\mathcal{F}$ is used as an optimization function.

First, for the uniform quantization, the networks designed by the \textit{preprocessor} recorded relatively very low accuracy compared to the baseline, especially for VGG-16. It can be seen that the classification in VGG-16 is rarely performed correctly. On the other hand, we can observe the effectiveness of the \textit{forward optimizer} in improving the performance of the homogeneous networks designed by the \textit{preprocessor}. The \textit{forward optimizer} increased the top-1 accuracy for AlexNet, VGG-16, and ResNet-18 by $21.69$\%, $18.30$\%, and $9.83$\%, respectively. However, there is still a large gap between the performance of the full-precision networks and uniformly quantized networks even after optimizing the quantization parameters. Thus, a uniform assignment of the bits is not the desired choice for maintaining accuracy.

In contrast, in the case of the proposed ${\textnormal{\textbf{FxP-QNet}}}$, it significantly reduced the wordlength of model parameters as well as intermediate activations while achieving comparable accuracy with full-precision baselines. Furthermore, the results show that the proposed ${\textnormal{\textbf{FxP-QNet}}}$ dramatically outperforms uniform quantization methods. For the top-1 accuracy on VGG-16, ${\textnormal{\textbf{FxP-QNet}}}$ improves the \textit{forward optimizer} result by $48.99$\%. The gap to the full-precision model is only $0.95$\%. The results on AlexNet and ResNet-18 are similar, only $0.95$\% and $1.99$\% loss in top-1 classification accuracy is shown compared with the baseline models.

Thus, it is easy to conclude that mixed-precision quantization has advantages over its homogeneous counterparts in terms of performance and model size. It is noteworthy that despite our method not needing any retraining and fine-tuning, it still poses a very competitive quantization scheme. It can be directly deployed without any retraining or fine-tuning, making our quantization scheme highly appealing for hardware implementation.

\subsubsection{Memory Requirement and Footprint}

\begin{figure*}[t!]
\vspace{-1mm}
\captionsetup[subfloat]{captionskip=0.01pt}
\centering
\subfloat{\includegraphics[width=0.0125\textwidth, trim=0.05cm -1.05cm 24.00cm 0.35cm, clip]{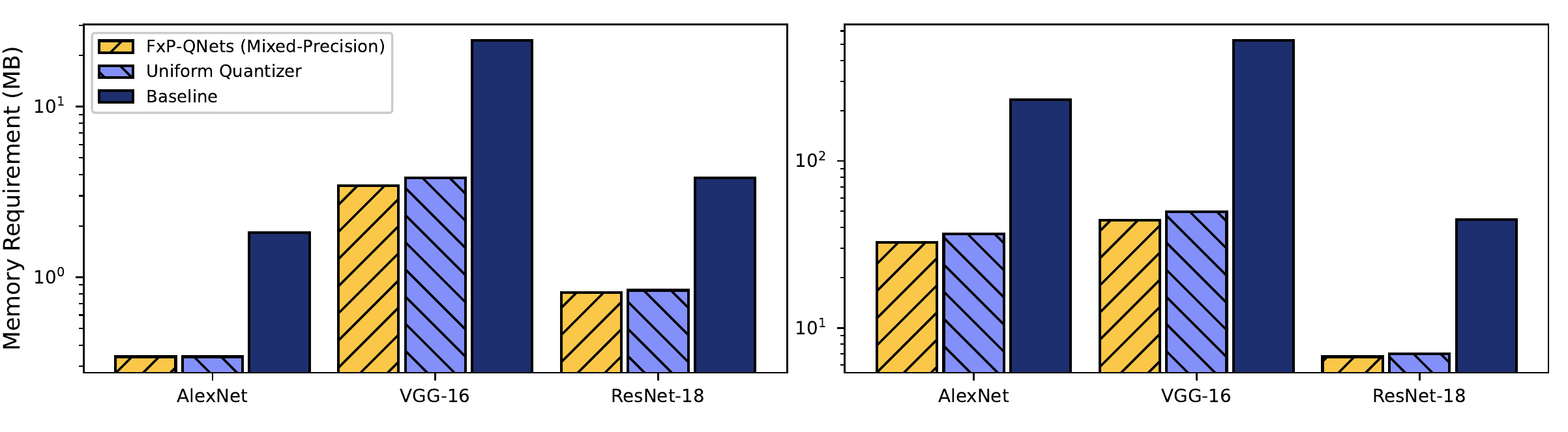}}
\hspace{0.2mm}\setcounter{subfigure}{0}
\subfloat[Read-write memory.]{\includegraphics[width=0.46\textwidth, trim=0.55cm 0.0cm 12.1cm 0.35cm, clip]{imgs/memory_requirement}}
\hspace{2.5mm}
\subfloat[Read-only memory.]{\includegraphics[width=0.46\textwidth, trim=12.4cm 0.0cm 0.25cm 0.35cm, clip]{imgs/memory_requirement}}
\caption[]{The requirement for (a) read-write memory, and (b) read-only memory, in Megabyte (MB), for AlexNet, VGG-16, and ResNet-18 architectures.}
\label{fig:memory_requirement}
\end{figure*}

At any step of the inference, a pair of temporary activation vectors and the whole set of model parameters must be present in the memory. As the quantized input and output activations of a layer are temporarily stored, a read-write (RW) memory is usually used for this purpose. On the other hand, the quantized learnable parameters are not changed during the lifetime of DNN inference, so a read-only (RO) memory is usually adopted to store the frozen parameters. It is worth noting that the last layer of each residual block in ResNet-18 performs an element-wise sum on the input activations from the skip connection and the residual connection. Thus, if considering a DNN consisting of $L$ layers where $M$ of them with learnable parameters, $M \leq L$, the above requirement is translated as\vspace{-2mm}

\begin{equation} \label{eq:RW_R}
\begin{aligned}
 \!\!\!\!\!\!\mathrm{RW}_R\left(\mathcal{Q}_a\right) = & \; \max\!\Bigg(\!\!\!\!\!\!\!\!\!\!\!\!\!\!\!\!\!\sum_{\;\;\;\;\;\;\;\;\;\;\;\;\;\ell_\text{inp} \, \in \, \mathrm{INP}_\ell^{(a)}}\!\!\!\!\!\!\!\!\!\!\!\!\!\!\!\!\mathrm{MFP}(a, \ell_\text{inp}, \mathcal{Q}_a) \; + \\
 & \;\;\;\;\;\;\;\;\;\;\;\;\;\mathrm{MFP}(a, \ell, \mathcal{Q}_a) \;, \;\;\;\; \forall \, \ell \in [L]\Bigg)
\end{aligned}
\end{equation}

\begin{equation} \label{eq:RO_R}
\begin{aligned}
\!\!\!\!\!\!\!\!\!\!\!\!\!\!\!\!\!\!\!\!\!\!\!\!\!\!\!\!\!\!\!\!\!\!\!\!\!\!\mathrm{RO}_R\left(\mathcal{Q}_w, \mathcal{Q}_b\right) = & 
\sum_{\ell \, \in \, [M]}\!\Big(\mathrm{MFP}(w, \ell, \mathcal{Q}_w) \; + \\
& \;\;\;\;\;\;\;\;\;\;\;\,\mathrm{MFP}(b, \ell, \mathcal{Q}_b)\Big)
\end{aligned}
\end{equation}

where $\mathcal{Q}_d$ indicates whether the data-structure type $d$, ${d \in \left\{a, w, b\right\}}$ is in full-precision ($\mathrm{FP32}$) or quantized ($\mathrm{FxP}$), $\mathrm{INP}_\ell^{(a)}$ is a set of all layers providing input activations to layer $\ell$, and the function $\mathrm{MFP}(d, \ell, \mathcal{Q}_d)$ returns the memory footprint in MB for the data-structure type $d$ in layer $\ell$ based on its the size, $\nu$, as well as its quantization status, $\mathcal{Q}_d$, as

\begin{equation} \label{eq:mfp}
 \mathrm{MFP}(d, \ell, \mathcal{Q}_d) = \frac{\nu_\ell^{(d)}}{2^{23}} \times
 \begin{cases}
 32 \;\:\,, & \mathcal{Q}_d = \mathrm{FP32}\\
 k_\ell^{\left(d\right)}, & \mathcal{Q}_d = \mathrm{FxP}
 \end{cases}
\end{equation}
\\
where ${k_\ell^{\left(d\right)}}$ is the bit-precision level of the data-structure $d$ in layer $\ell$. Note that we neglected the memory footprint for the shift and ${\textnormal{\textit{FxP-QLayer}}}$ parameters in the aforementioned computations because their size, in terms of bytes, is too small as we are applying a layer-wise quantization. Thus, we calculate the compression rate of activations and model parameters as follows

\begin{equation} \label{eq:compression_a}
 \mathrm{COMP}_{\text{Act}} = \mathrm{RW}_R\left(\scalebox{.85}{$\mathrm{FP32}$}\right) 
/ \; \mathrm{RW}_R\left(\scalebox{.85}{$\mathrm{FxP}$}\right)
\end{equation}

\begin{equation} \label{eq:compression_model}
 \mathrm{COMP}_{\text{Model}} = \mathrm{RO}_R\left(\scalebox{.85}{$\mathrm{FP32}$}, \scalebox{.85}{$\mathrm{FP32}$}\right)
/ \; \mathrm{RO}_R\left(\scalebox{.85}{$\mathrm{FxP}$}, \scalebox{.85}{$\mathrm{FxP}$}\right)
\end{equation}

Figure~\ref{fig:memory_requirement} reports the memory requirements of the proposed ${\textnormal{\textbf{FxP-QNet}}}$. For the read-write memory requirement, the results show that ${\textnormal{\textbf{FxP-QNet}}}$ is $11$\% and $3$\% less memory-demanding than using the uniform quantization methodology for VGG-16 and ResNet-18, respectively. Furthermore, the read-only memory demand for ${\textnormal{\textbf{FxP-QNet}}}$ is $12$\%, $12$\%, and $4$\% less than uniform quantizer for AlexNet, VGG-16, and ResNet-18, respectively. Thus, it is easy to conclude that mixed-precision quantization has advantages over its uniform counterpart in terms of memory requirements. Nevertheless, the achievements of the uniform quantizer come at the cost of being less accurate than the mixed-precision quantizer.

Compared to baseline requirements, the quantized DNNs reduce the read-write and read-only memory requirements by about $1.5$MB and $200$MB for AlexNet, $21$MB and $483.5$MB for VGG-16, $3$MB and $38$MB for ResNet-18, respectively. To view from the off-chip bandwidth side, a $128$-bit off-chip bandwidth can transfer $4$ filter weight data per cycle when using full-precision weights. But after ${\textnormal{\textbf{FxP-QNet}}}$ quantization, it can transfer $28$, $47$, and $26$ filter weight data per cycle, on average, for AlexNet, VGG-16, and ResNet-18 architectures, respectively, without extra power consumption for data-movement.

\begin{figure*}[t!]
\vspace{-1mm}
\captionsetup[subfloat]{captionskip=0.01pt}
\centering
\subfloat[High-level architecture of the proposed hardware accelerator.\label{fig:hardware_architecture_1}]{\includegraphics[width=0.60\textwidth, trim=0.0cm -1.0cm 0.0cm 0.0cm, clip]{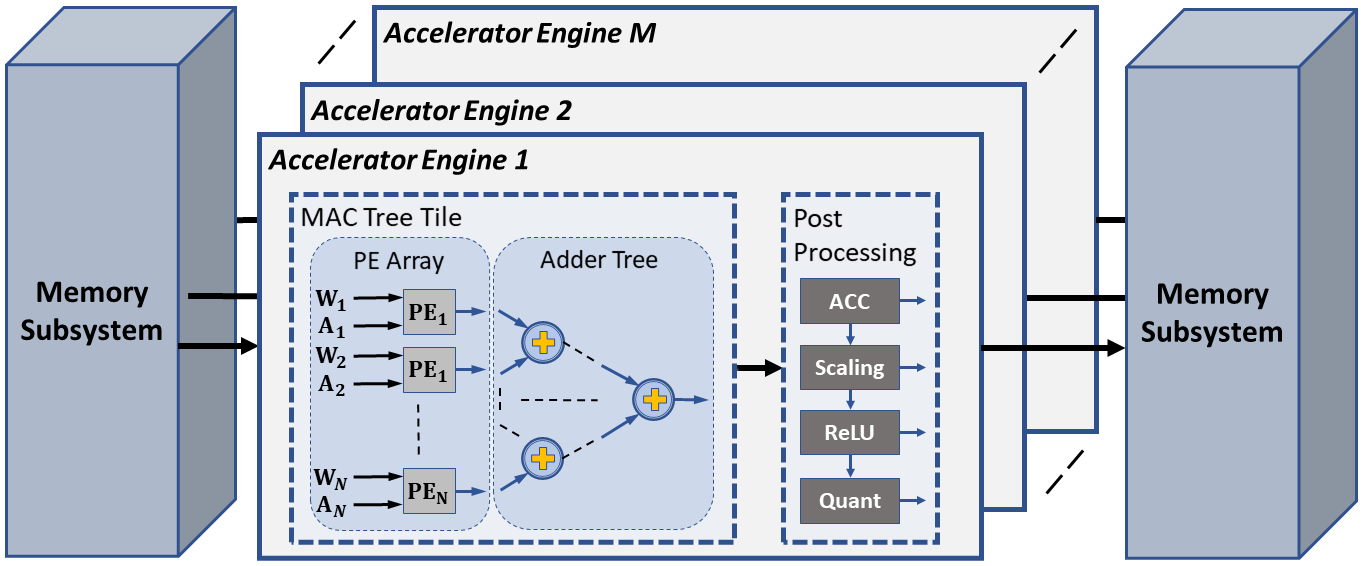}}
\hspace{5.5mm}
\subfloat[FxP-QLayer hardware design.\label{fig:hardware_architecture_2}]{\includegraphics[width=0.34\textwidth, trim=0.0cm -0.8cm 0.0cm 0.0cm, clip]{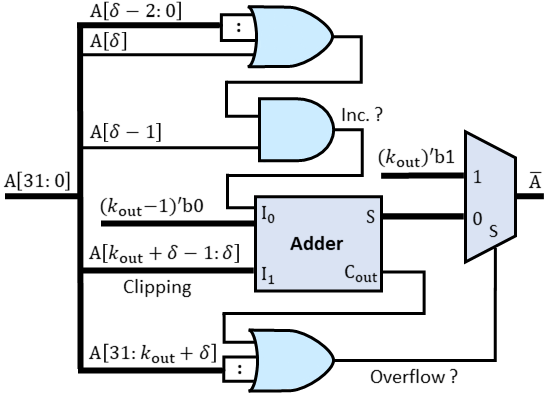}}
\caption[]{The architecture of (a) the proposed hardware accelerator, and (b) ${\textnormal{\textit{FxP-QLayer}}}$ based on the right-shift ($\delta$) and output bit-width ($k_\textnormal{out}$) parameters obtained during the design phase by ${\textnormal{\textbf{FxP-QNet}}}$ for ResNet-18 model.}
\end{figure*}

\subsubsection{Hardware Efficiency}

The proposed hardware architecture consists of $M$ accelerator engines that work in parallel to increase the computational throughput. Each accelerator engine entails a MAC tree tile and a post-processing unit as illustrated in Figure~\ref{fig:hardware_architecture_1}. The MAC tree tile contains $N$ processing elements (PEs) and an adder tree. Each PE performs multiplication on a $k_w$-bit input weight filter and a $k_a$-bit input activation feature. Then, the resultant products are accumulated using the binary adder tree. On the other hand, the post-processing unit includes the hardware circuitry required for partial result and bias accumulation, ${\textnormal{\textit{BN}}}$ and $\textnormal{\textit{AVG-POOL}}$ scaling, ${\textnormal{\textit{ReLU}}}$ activation, and quantization. Additionally, two memory subsystems are employed to derive the accelerator engines and hold output data.

We modeled the behavior of the hardware accelerator in Verilog and synthesized it using Xilinx Vivado Design Suite targeting the Xilinx Artix-7 FPGA (XC7A100TCSG324-1). 
We kept all synthesis properties to their default. 
In this experiment, we evaluate the hardware efficiency when implementing ResNet-18 model. Precisely, we consider the test-case of implementing the hardware accelerator for ${\textnormal{\textit{CONV}}}$ layers that require a large number of $3 \times 3$ MAC operations to produce final output activations. It is noteworthy that $3 \times 3$ is the most widely used kernel size in DNNs. The depth of the PE array is adjusted to achieve the highest possible utilization and data reuse, whereas the value of $M$ is constrained by the hardware resources.

We compare the performance of the proposed ${\textnormal{\textbf{FxP-QNet}}}$ mixed-precision quantization with the conventional $8$-bit linear quantization (\textit{Quant8}) scheme. The \textit{Quant8} scheme implements each PE using an $8$-bit fixed-point multiplier. After computing MAC tree output, \textit{Quant8} converts the result to a floating-point number and performs the computations of the post-processing unit using IEEE-754 single format floating-point arithmetic. 
The arithmetic operations on the floating-point numbers are implemented using Vivado IP integrator.
Note that \textit{Quant8} is a popular quantization technique in Pytorch and TensorRT~\cite{migacz20178}. Conversely, ${\textnormal{\textbf{FxP-QNet}}}$ scheme performs all the operations with fixed-point arithmetic, including the quantization process which is performed through ${\textnormal{\textit{FxP-QLayer}}}$. In doing so, PEs and ${\textnormal{\textit{FxP-QLayer}}}$ are configured based on the wordlength allocation and the quantization parameters obtained from the bit-precision level reduction framework.

For ${\textnormal{\textbf{FxP-QNet}}}$, we consider two scenarios as it allocates a different bit-precision level to different layers. These scenarios evaluate the hardware accelerator when it is used to perform the computations of (i) the least compact layer (LCL), i.e., the ${\textnormal{\textit{CONV}}}$ layer with the largest weights and activations wordlengths allocated, and (ii) the most compact layer (MCL), i.e., the ${\textnormal{\textit{CONV}}}$ layer with the smallest weights and activations wordlengths allocated. We utilize the dynamic partial reconfiguration feature offered by modern FPGAs to configure the hardware accelerator at run-time for each ${\textnormal{\textit{CONV}}}$ layer. The overhead introduced by the reconfiguration can be compensated by using the reconfigured hardware accelerator many times, which is the case for the accelerator engines. Thus, the reconfiguration overhead becomes negligible with respect to the duration of the computation interval.

\begin{table}[tb!]
\begin{center}
\captionsetup{justification=centering}
\caption{Implementation results for ResNet-18 on Artix-7.}
\label{tab:hardware_anaysis_results}
\renewcommand{\arraystretch}{1.6}
\setlength\tabcolsep{2.9pt}
\begin{tabular}{| l | >{\centering}p{0.3in} | >{\centering}p{0.307in} | >{\centering}p{0.307in} | >{\centering}p{0.307in} | >{\centering}p{0.307in} | >{\centering}p{0.307in} | c |}
\hline
\multicolumn{2}{| c |}{\multirow{2}{*}{\textbf{Scheme}}} & \multirow{2}{*}{\makecell{\textbf{Num.} \\ \textbf{PEs}}} & \multirow{2}{*}{\textbf{LUTs}} & \multirow{2}{*}{\textbf{FFs}} & \multirow{2}{*}{\textbf{DSPs}} & \multirow{2}{*}{\makecell{\textbf{Power} \\ \textbf{(mW)}}} & \multirow{2}{*}{\makecell{\textbf{F$_{\bm{max}}$} \\ \textbf{(MHz)}}} \\
\multicolumn{2}{| c |}{} & & & & & & \\
\hline\hline 
\multicolumn{2}{| l |}{\textbf{Quant8}} & 512 & 51,408 & 31,176 & 48 & 665 & 100 \\
\hline
\multirow{2}{*}{\textbf{${\textnormal{\textbf{FxP-QNet}}}$\,}} & \textbf{LCL} & 384 & 25,448 & 11,772 & 6 & 363 & 135 \\
\cdashline{2-8}
 & \textbf{MCL} & 512 & 12,239 & 10,284 & 1 & 210 & 135 \\
\hline
\end{tabular}
\end{center}
\end{table}

Table~\ref{tab:hardware_anaysis_results} shows the FPGA hardware resource usage in terms of the look-up tables (LUTs), flip-flops (FFs), and digital signal processors (DSPs). It also provides the power consumption, the maximum operating frequency (F$_{max}$), and the number of PEs implemented for each scheme. The number of PEs in \textit{Quant8} is limited by the available resources in the device. Consequently, the hardware accelerator for \textit{Quant8} scheme is implemented with $512$ PEs. On the other hand, ${\textnormal{\textbf{FxP-QNet}}}$ PEs is constrained by the allocated resources to the reconfigurable region. Therefore, the ${\textnormal{\textbf{FxP-QNet}}}$ LCL uses $25$\% less PEs while the ${\textnormal{\textbf{FxP-QNet}}}$ MCL employs the same number of PEs as in \textit{Quant8}. Note that the size of the reconfigurable region, in terms of hardware resources, involves a trade-off between high acceleration and low reconfiguration overhead.

The effectiveness of ${\textnormal{\textbf{FxP-QNet}}}$ can be seen from the utilized resources. In the worst-case scenario, ${\textnormal{\textbf{FxP-QNet}}}$ uses $50.5$\%, $62.2$\%, and $87.5$\% less LUTs, FFs, and DSPs, respectively, than \textit{Quant8} to design a hardware accelerator integrating $384$ PEs. This is despite the fact that the operating frequency of ${\textnormal{\textbf{FxP-QNet}}}$ LCL is $1.35\times$ that of \textit{Quant8}, and the former is $1.83\times$ better in saving power than the latter on average. Overall, it is evident that ${\textnormal{\textbf{FxP-QNet}}}$ has a comparable hardware efficiency for the worst-case allocated wordlength, but a significant improvement in hardware cost, power efficiency, and operating frequency for best-case wordlength allocated.

Concerning the overhead of quantizing activations, \textit{Quant8} de-quantizes resultant activations from the MAC tile and then re-quantizes them back after ${\textnormal{\textit{BN}}}$ computation due to the use of FP32 quantization scalars. That is why linear quantization techniques adopt floating-point arithmetic for the computations of post-processing unit, which consumes much energy and hardware resources. Hence, \textit{Quant8} has a non-negligible quantization overhead. In contrast, ${\textnormal{\textbf{FxP-QNet}}}$ is tailored for integer-only deployment. Both ${\textnormal{\textit{BN}}}$ and quantization are performed with integer arithmetic. Specifically, ${\textnormal{\textit{FxP-QLayer}}}$ reduces the bit-precision level of a $32$-bit integer input activation (the output of ${\textnormal{\textit{BN}}}$ and ${\textnormal{\textit{ReLU}}}$ layers) on-the-fly using an adder, a multiplexer, and logic gates, as shown in Figure~\ref{fig:hardware_architecture_2}, and its implementation uses LUTs only. The multiplexer selects outputs between the clipped-then-rounded value and the maximum representable value in case of an overflow. Consequently, the overhead of ${\textnormal{\textit{FxP-QLayer}}}$ is negligible.

\subsubsection{Comparison with Prior Work}

\begin{table*}[t!]
 \centering
 \captionsetup{justification=centering}
 \caption{Quantization results of AlexNet, VGG-16, and ResNet-18 on ImageNet validation dataset. We calculate the compression rate of activations and model parameters using Equation~(\ref{eq:compression_a}) and Equation~(\ref{eq:compression_model}), respectively.}
 \setlength\tabcolsep{1.5pt}
 \renewcommand{\arraystretch}{1.6}
 \centerline{
 \begin{tabular}{| l || l || c || c || c || c ||  >{\centering}p{0.56in} | >{\centering}p{0.56in} ||>{\centering}p{0.5in} | >{\centering}p{0.5in} || c | c |} 
 \hline
 \multicolumn{1}{| c||}{\multirow{2}{*}{\textbf{Architecture}}} & \multicolumn{1}{ c||}{\multirow{2}{*}{\textbf{Framework}}} & \multirow{2}{*}{\makecell{\textbf{\textbf{Integer-Only}}\\\textbf{Inference}}} & \multirow{2}{*}{\makecell{\textbf{\textbf{Training/}}\\\textbf{Fine-Tuning}}} & \multirow{2}{*}{\makecell{\textbf{\textbf{Quantization}}\\\textbf{Granularity}}} & \multirow{2}{*}{\makecell{\textbf{Mixed-}\\\textbf{Precision}}} & \multicolumn{2}{c||}{\textbf{CONV/FC Bit-Precision}} & \multicolumn{2}{c||}{\textbf{Compression}} & \multicolumn{2}{c|}{\textbf{Degradation}}\\ 
 \cline{7-12}
 & & & & &  &   \textbf{Activations} & \textbf{Weights}& \textbf{Model} & \textbf{Activation} & \textbf{Top-1} & \textbf{Top-5}\\ 
 \hline\hline
 \multirow{13}{*}{\:AlexNet} & DoReFa-Net~\cite{zhou2016dorefa} & No & Yes & Per-Layer & No & $\text{13.47}^{(\ref{itm:tab_com_first_layer}, \ref{itm:tab_com_last_layer})}$
 & $\text{4.03}^{(\ref{itm:tab_com_first_layer}, \ref{itm:tab_com_last_layer})}$ & 7.89$\times$ & 2.58$\times$ & 1.50\% & 0.20\% \\
 \cdashline{2-12}
 & WRPN~\cite{mishra2017wrpn} & No & Yes & Per-Layer & No & $\text{13.47}^{(\ref{itm:tab_com_first_layer}, \ref{itm:tab_com_last_layer})}$ & $\text{4.03}^{(\ref{itm:tab_com_first_layer}, \ref{itm:tab_com_last_layer})}$ & 7.89$\times$ & 2.58$\times$ & 1.40\% & N/A \\
 \cdashline{2-12}
 & PACT~\cite{choi2018pact} & No & Yes & Per-Layer & No & $\text{13.47}^{(\ref{itm:tab_com_first_layer}, \ref{itm:tab_com_last_layer})}$
 & $\text{4.03}^{(\ref{itm:tab_com_first_layer}, \ref{itm:tab_com_last_layer})}$ & 7.89$\times$ & 2.58$\times$ & 1.00\% & -0.70\% \\
 \cdashline{2-12}
 & KDE-KM~\cite{seo2019efficient} & No & No & Per-Group & No & 32
 & 4 & 7.99$\times$ & 1.00$\times$ & 9.09\% & 6.97\% \\ 
 \cdashline{2-12}
 & LQ-Nets~\cite{zhang2018lq} & No & Yes & Per-Channel & No & $\text{13.47}^{(\ref{itm:tab_com_first_layer}, \ref{itm:tab_com_last_layer})}$
 & $\text{4.03}^{(\ref{itm:tab_com_first_layer}, \ref{itm:tab_com_last_layer})}$ & 7.89$\times$ & 2.58$\times$ & 4.40\% & 3.40\% \\
 \cdashline{2-12}
 & TSQ~\cite{wang2018two} & No & Yes & Per-Channel & No & $\text{13.47}^{(\ref{itm:tab_com_first_layer}, \ref{itm:tab_com_last_layer})}$
 & $\text{4.03}^{(\ref{itm:tab_com_first_layer}, \ref{itm:tab_com_last_layer})}$ & 7.89$\times$ & 2.58$\times$ & 0.50\% & 1.00\% \\
 \cdashline{2-12}
 & MMSE~\cite{kravchik2019low} & No & No  & Per-Group & No & 4 & 4 & 8.00$\times$ & 5.13$\times$ & 2.15\% & 1.21\% \\
 \cdashline{2-12}
 & ACIQ~\cite{banner2019post} & No & No & Per-Channel & Yes & 5.53 & 8 & 4.00$\times$ & 4.00$\times$ & 4.32\% & N/A \\
 \cdashline{2-12}
 & QIL~\cite{jung2019learning} & No & Yes  & Per-Layer & No & $\text{13.47}^{(\ref{itm:tab_com_first_layer}, \ref{itm:tab_com_last_layer})}$
 & $\text{4.03}^{(\ref{itm:tab_com_first_layer}, \ref{itm:tab_com_last_layer})}$ & 7.89$\times$ & 2.58$\times$ & 3.70\% & N/A \\
 \cdashline{2-12}
 & SegLog~\cite{xu2020memory} & No & No  & Per-Layer & No & 32 & 5 & 6.40$\times$ & 1.00$\times$ & N/A & 1.73\% \\
 \cdashline{2-12}
 & MP-QNN~\cite{chu2021mixed} & No & Yes & Per-Layer & Yes & $\text{13.47}^{(\ref{itm:tab_com_first_layer}, \ref{itm:tab_com_last_layer})}$
 & $\text{3.19}^{(\ref{itm:tab_com_first_layer}, \ref{itm:tab_com_last_layer})}$ & 9.95$\times$ & 2.57$\times$ & 3.42\% & 4.30\% \\
 \cdashline{2-12}
 & \multicolumn{1}{ l||}{\multirow{2}{*}{${\textnormal{\textbf{FxP-QNet}}}$ (ours)}} & \multicolumn{1}{ c||}{\multirow{2}{*}{Yes}} & \multicolumn{1}{ c||}{\multirow{2}{*}{No}} & \multicolumn{1}{ c||}{\multirow{2}{*}{Per-Layer}} & \multicolumn{1}{ c||}{\multirow{2}{*}{Yes}} & 5.79 & 4.48 & 7.14$\times$ & 5.33$\times$ & 0.95\% & 0.88\% \\ 
 \cdashline{7-12}
 &  &  &  &  &  & 5.44 & 3.85 & 8.30$\times$ & 5.62$\times$ & 1.95\% & 1.61\% \\ 
 \hline
 \multirow{10}{*}{\:VGG-16} & LogQuant~\cite{miyashita2016convolutional} & Yes & Yes & Per-Layer & No & $\text{4.46}^{(\ref{itm:tab_com_first_layer})}$ & 4.11 & 7.79$\times$ & 8.00$\times$ & N/A & 6.40\% \\
 \cdashline{2-12}
  & Going Deeper~\cite{qiu2016going} & Yes & No & Per-Layer & No & 8.00 & 4.43 & 7.23$\times$ & 4.00$\times$ & 1.14\% & 0.40\% \\
 \cdashline{2-12}
 & KDE-KM~\cite{seo2019efficient} & No & No & Per-Group & No & 32
 & 4 & 7.99$\times$ & 1.00$\times$ & 3.79\% & 2.19\% \\ 
 \cdashline{2-12}
 & TSQ~\cite{wang2018two} & No & Yes & Per-Channel & No & $\text{2.51}^{(\ref{itm:tab_com_first_layer}, \ref{itm:tab_com_last_layer})}$
 & $\text{2.89}^{(\ref{itm:tab_com_first_layer}, \ref{itm:tab_com_last_layer})}$ & 11.07$\times$ & 16.0$\times$ & 2.00\% & 0.70\% \\
 \cdashline{2-12}
 & MMSE~\cite{kravchik2019low} & No & No  & Per-Group & No & 4 & 4 & 8.00$\times$ & 7.87$\times$ & 2.82\% & 1.40\% \\
 \cdashline{2-12}
 & ACIQ~\cite{banner2019post} & No & No  & Per-Channel & Yes &  4.07 & 4.12 & 7.76$\times$ & 5.33$\times$ & 1.10\% & N/A\\ 
 \cdashline{2-12}
 & ALT~\cite{jain2019trained} & Yes & Yes  & Per-Layer & No & 8 & 4 & 8.00$\times$ & 4.00$\times$ & 0.40\% & 0.20\%\\
 \cdashline{2-12}
 & SegLog~\cite{xu2020memory} & No & No  & Per-Layer & No & 32 & 5 & 6.40$\times$ & 1.00$\times$ & N/A & 0.74\% \\
 \cdashline{2-12}
 & MXQN~\cite{huang2021mxqn} & No & No & Per-Layer & No & $\text{6.43}^{(\ref{itm:tab_com_first_layer})}$ & $\text{9.68}^{(\ref{itm:tab_com_last_layer})}$ & 3.30$\times$ & 5.33$\times$ & 1.05\% & 0.50\% \\ 
 \cdashline{2-12}
 & ${\textnormal{\textbf{FxP-QNet}}}$ (ours) & Yes & No & Per-Layer & Yes & 4.55 & 2.68 & 11.93$\times$ &  7.11$\times$ & 0.95\% & 0.94\%\\ 
 \hline
 \multirow{12}{*}{\:ResNet-18} & DoReFa-Net~\cite{zhou2016dorefa} & No & Yes & Per-Layer & No & $\text{9.67}^{(\ref{itm:tab_com_first_layer} - \ref{itm:tab_com_downsample_layer})}$ & $\text{4.72}^{(\ref{itm:tab_com_first_layer} - \ref{itm:tab_com_downsample_layer})}$ & 6.71$\times$ & 1.00$\times$ & 2.90\% & 2.00\% \\
 \cdashline{2-12}
 & PACT~\cite{choi2018pact}  & No & Yes & Per-Layer & No & $\text{9.67}^{(\ref{itm:tab_com_first_layer} - \ref{itm:tab_com_downsample_layer})}$ & $\text{4.72}^{(\ref{itm:tab_com_first_layer} - \ref{itm:tab_com_downsample_layer})}$ & 6.71$\times$ & 1.00$\times$ & 2.30\% & 1.40\% \\
 \cdashline{2-12}
 & KDE-KM~\cite{seo2019efficient} & No & No & Per-Group & No & 32
 & 4 & 7.90$\times$ & 1.00$\times$ & 7.91\% & 5.15\%\\ 
 \cdashline{2-12}
 & LQ-Nets~\cite{zhang2018lq} & No & Yes & Per-Channel & No & $\text{5.01}^{(\ref{itm:tab_com_first_layer}, \ref{itm:tab_com_last_layer})}$ & $\text{4.29}^{(\ref{itm:tab_com_first_layer}, \ref{itm:tab_com_last_layer})}$ & 7.37$\times$ & 1.00$\times$ & 2.10\% & 1.60\%\\
 \cdashline{2-12}
 & MMSE~\cite{kravchik2019low} & No & No  & Per-Group & No & 4 & 4 & 7.93$\times$ & 6.75$\times$ & 2.23\% & 1.21\% \\
 \cdashline{2-12}
 & ACIQ~\cite{banner2019post} & No & No  & Per-Channel & Yes & 4.28 & 4.18 & 7.57$\times$ & 4.00$\times$ & 2.70\% & N/A\\ 
 \cdashline{2-12}
 & QIL~\cite{jung2019learning} & No & Yes & Per-Layer & No & $\text{8.91}^{(\ref{itm:tab_com_first_layer} - \ref{itm:tab_com_downsample_layer})}$
 & $\text{3.78}^{(\ref{itm:tab_com_first_layer} - \ref{itm:tab_com_downsample_layer})}$ & 8.36$\times$ & 1.00$\times$ & 4.50\% & N/A \\
 \cdashline{2-12}
 & CCQ~\cite{khan2020learning}  & No & Yes & Per-Layer & Yes & N/A & $\text{3.71}^{(\ref{itm:tab_com_downsample_layer})}$ & 8.52$\times$ & N/A & 2.60\% & N/A \\
 \cdashline{2-12}
 & MP-QNN~\cite{chu2021mixed} & No & Yes & Per-Layer & Yes & $\text{11.66}^{(\ref{itm:tab_com_first_layer} - \ref{itm:tab_com_first_residual_stage})}$
 & $\text{3.24}^{(\ref{itm:tab_com_first_layer} - \ref{itm:tab_com_downsample_layer})}$ & 9.73$\times$ & 1.00$\times$ & 4.27\% & 3.20\% \\
 \cdashline{2-12}
 & MXQN~\cite{huang2021mxqn} & No & No & Per-Layer & No & $\text{9.65}^{(\ref{itm:tab_com_first_layer})}$ & 8 & 3.98$\times$ & 2.71$\times$ & 0.63\% & 0.40\% \\ 
 \cdashline{2-12}
 & DoubleQExt~\cite{see2021doubleqext} & No & No  & Per-Layer & Yes & 8 & $\text{5.77}^{(\ref{itm:tab_com_doubleqext})}$ & 5.55$\times$ & 4.00$\times$ & 1.79\% & 1.04\% \\
 \cdashline{2-12}
 & ${\textnormal{\textbf{FxP-QNet}}}$ (ours) & Yes & No & Per-Layer & Yes & 6.29 & 4.81 & 6.65$\times$ & 4.57$\times$ & 1.99\% & 1.33\%\\ 
 \hline
 \multicolumn{12}{l}{\rule{0pt}{4ex} 
 \text{\footnotesize \raisebox{1.3ex}[0ex][0ex]{\textbf{Note:}}
 \parbox{.9\textwidth}{
\renewcommand\labelenumi{(\theenumi)}
 \begin{enumerate*}
 \item Do not quantize first layer. \label{itm:tab_com_first_layer} 
 \item Do not quantize last layer. \label{itm:tab_com_last_layer}
 \item Do not quantize downsample skip connection layer. \label{itm:tab_com_downsample_layer}
 \item Do not quantize the input activations to first residual stage. \label{itm:tab_com_first_residual_stage}
 \item Quantizes the weights to either 5 or 8 bits. \label{itm:tab_com_doubleqext}
\end{enumerate*}
\noindent
}
}}
 \end{tabular}
 }
 \label{tab:comparison}
\end{table*}

In this section, we present the results of ${\textnormal{\textbf{FxP-QNet}}}$ and other quantization frameworks considering seven criteria as summarized in Table~\ref{tab:comparison}. For each framework, we indicate whether the inference computations are hardware-friendly because of using integer-only arithmetic or they include floating-point operations. We also mention whether the framework facilitates rapid deployment of quantized models or requires a training phase. Furthermore, we show the quantization granularity for each framework as it reflects the additional computational overhead during quantization. Moreover, we highlight whether the framework designs a homogeneous or a mixed-precision quantized DNN.

For the computation cost comparison, we refer to the average wordlength of input activations and weights used to perform the calculation of convolution and fully-connected layers. On the other hand, the compression denotes the amount of reduction in memory requirements for activations and model parameters. Last but not the least, since all the other reported works have different baselines, we use the degradation in top-1 and top-5 accuracy for comparison rather than using absolute accuracy numbers. Note that some of these frameworks report several results such as when quantizing DNN models to different bit-precision levels. In this case, we present the best results that meet our compression criteria for fair comparison.

Here, we should emphasize that our work focuses on a hardware-friendly design of quantized DNNs that accelerates the inference, reduces the memory requirements, and at the same time is tailored for integer-only deployment. Thus, it is completely unfair to compare our framework with those interested mainly in speeding up the MAC operation by quantizing activations and weights to a low wordlength before performing MAC operation, and then multiplying each resulting activation by a $32$-bit floating-point scalar to recover the dynamic range. Nevertheless, we make such a comparison to show the effectiveness of ${\textnormal{\textbf{FxP-QNet}}}$. 

Overall, from Tables~\ref{tab:comparison}, it is evident that despite the ${\textnormal{\textbf{FxP-QNet}}}$-quantized architectures perform integer-only inference and are not retrained, they achieve the best trade-off between the computational efficiency, memory requirement, and classification accuracy. For AlexNet architecture, we achieve the highest I/O activation compression ratio ($5.62\times$) and the second-highest compression ratio for model parameters after MP-QNN~\cite{chu2021mixed}. This is despite the fact that the latter framework has a top-1 accuracy degradation of $3.42$\% even though it employs training and floating-point multiplication to maintain network accuracy, whereas ${\textnormal{\textbf{FxP-QNet}}}$ has degradation of $1.95$\% only.

For VGG-16 architecture, ${\textnormal{\textbf{FxP-QNet}}}$ achieves the highest model compression ratio ($11.93\times$). Furthermore, we can see that ${\textnormal{\textbf{FxP-QNet}}}$ performs better than those frameworks that support integer-only deployment; it attains higher accuracy, in general, with the smallest model size. Note that the improvement in ALT~\cite{jain2019trained} classification accuracy comes at the cost of computational efficiency as it uses $3.45$ and $1.32$ more bits for activations and weights, respectively, during the calculation of ${\textnormal{\textit{CONV}}}$ and ${\textnormal{\textit{FC}}}$ layers. With regards to frameworks including floating-point arithmetic, and focusing on TSQ~\cite{wang2018two}, the framework with the lowest number of bits involved in the computation of ${\textnormal{\textit{CONV}}}$ and ${\textnormal{\textit{FC}}}$ layers, this wordlength reduction comes at the cost of being less accurate, even though TSQ is a quantization-aware training framework, in addition to having a large memory overhead since it requires a $32$-bit floating-point scalar for each output channel in each layer.

Additionally, we also compare ${\textnormal{\textbf{FxP-QNet}}}$ with other frameworks when quantizing a deeper and a more complex network structure such as ResNet-18 and show that the results attained by ${\textnormal{\textbf{FxP-QNet}}}$ are competitive even though ${\textnormal{\textbf{FxP-QNet}}}$ requires no retraining or fine-tuning and does not require floating-point scaling, as well. For instance, ${\textnormal{\textbf{FxP-QNet}}}$ is $1.68\times$ better in increasing the compression rate as compared to MXQN~\cite{huang2021mxqn} with only a $0.93$\% higher top-5 accuracy degradation. Further, while the latter framework performs the MAC operation 
%of ${\textnormal{\textit{CONV}}}$ and ${\textnormal{\textit{FC}}}$ layers 
using $9.65$-bit activations and $8$-bit weights, on average, the former is more computationally efficient because it uses $6.29$-bit activations and $4.81$-bit weights. Compared to DoubleQExt~\cite{see2021doubleqext}, ${\textnormal{\textbf{FxP-QNet}}}$ reduces memory requirements for model parameters by about $1.35$MB with only $0.20$\% and $0.29$\% extra drop in top-1 and top-5 accuracies, respectively. This is despite the fact that the former framework requires floating-point arithmetic to transform the output of MAC operations, whereas ${\textnormal{\textbf{FxP-QNet}}}$-quantized ResNet-18 is designed for integer-only deployment.

\section{Conclusion} \label{sec:conclusion}

In this paper, we presented ${\textnormal{\textbf{FxP-QNet}}}$, a framework to efficiently quantize DNN data-structures to low-precision numbers in dynamic fixed-point representation. Specifically, the ${\textnormal{\textbf{FxP-QNet}}}$ computes the low-precision representation of DNN data-structures as mentioned in Equations~\eqref{eq:fp_k_bit_quantization} and \eqref{eq:fp_1_bit_quantization}. Furthermore, we experimentally demonstrated that optimizing the quantization parameters for each layer and data-structure independently produces a local optimum solution. Thus, ${\textnormal{\textbf{FxP-QNet}}}$ selects the quantization parameters that minimize the post-training self-distillation error as well as network prediction errors as summarized in Algorithm~\ref{Algorithm:QFO}. Thereafter, the paper presented an iterative accuracy-driven bit-precision level reduction framework. ${\textnormal{\textbf{FxP-QNet}}}$ employs this framework to flexibly design a quantized DNN with a mixed-precision taking into account the robustness of data-structures to quantization as illustrated in Algorithm~\ref{Algorithm:BPR_1}.

With all the techniques aforementioned, ${\textnormal{\textbf{FxP-QNet}}}$ transforms current DNNs to hardware-friendly networks while still preserving accuracy at a high rate. The appealing aspects in our framework are as follows; \begin{enumerate*}[label=(\roman*)]
\item designing a pure fixed-point DNNs without expensive floating-point operations and thus providing unique computational and hardware efficiencies, \item adopting layer-wise quantization and therefore there is almost no additional hardware cost overhead for low-precision scaling as bit-shift operation is used for this, \item reducing quantization-induced degradation enough so as to make quantization-aware training unnecessary, \item employing a simple and very fast-to-implement quantization process, and \item facilitating rapid deployment of pre-trained off-the-shelf models.\end{enumerate*} 

Finally, the paper demonstrated the effectiveness of ${\textnormal{\textbf{FxP-QNet}}}$ through extensive experiments on several CNN-based architectures on the benchmark ImageNet dataset. The results showed that ${\textnormal{\textbf{FxP-QNet}}}$-quantized AlexNet, VGG-16, and ResNet-18 can reduce the overall memory requirements by $7.16\times$, $10.36\times$, and $6.44\times$ with only $0.95$\%, $0.95$\%, and $1.99$\% accuracy drop on ImageNet validation dataset, respectively. 

Evaluating the presented technique on other deep learning architectures such as recurrent neural networks (RNNs) and Transformer models is an interesting future work. Moreover, the current work was evaluated on image classification task. It would be interesting to see how ${\textnormal{\textbf{FxP-QNet}}}$ would perform on other related tasks like object detection, and image segmentation or even in other domains such as natural language processing. Future work shall focus on validating ${\textnormal{\textbf{FxP-QNet}}}$-quantized networks on hardware accelerating platforms such as FPGA and ASIC. Ongoing work shall also explore channel-wise quantization, which could be potentially useful for rigorous quantization of advanced network structures like MobileNets.

%note: uncomment to return references back

%\bibliography{main.bib}
%\bibliographystyle{ieeetr}

\phantomsection

\begin{IEEEbiography}[{\includegraphics[width=1in,height=1.25in,clip,keepaspectratio]{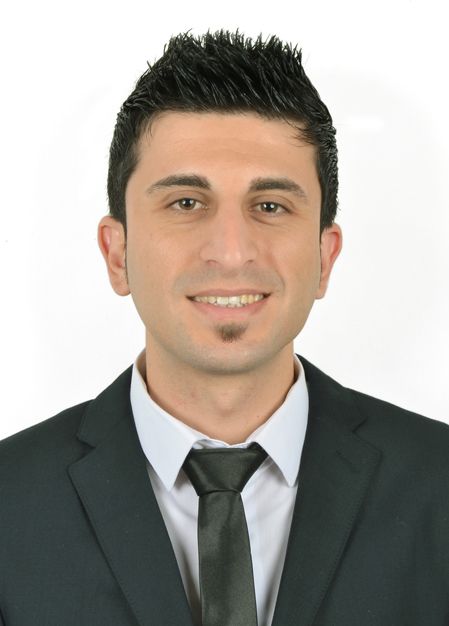}}]{Ahmad Shawahna} obtained M.S. in computer engineering from King Fahd University of Petroleum and Minerals (KFUPM), Saudi Arabia, in 2016. He also received the B.Sc degree in computer engineering from An-Najah National University, Palestine, in 2012. Ahmad Shawahna is a Ph.D. student in the department of computer engineering of KFUPM. In addition, he is currently working at the Center for Communications and IT Research (CCITR), KFUPM. His research interests include hardware accelerator, deep learning, CNNs, FPGA, wireless security, network security, Internet of Things (IoT), and cloud computing.
\end{IEEEbiography}

\begin{IEEEbiography}[{\includegraphics[width=1in,height=1.75in,clip,keepaspectratio]{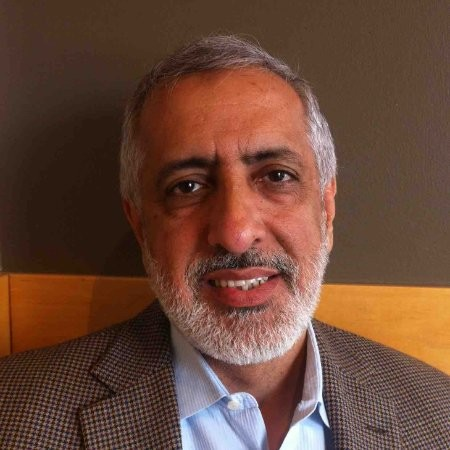}}]{Sadiq M. Sait} (Senior Member, IEEE) obtained his Bachelor's degree in Electronics Engineering from Bangalore University in 1981, and Master's and Ph.D. degrees in Electrical Engineering from KFUPM in 1983 and 1987, respectively. In 1981, Sait received the best Electronic Engineer award from the Indian Institute of Electrical Engineers, Bangalore (where he was born). Sait has authored over 250 research papers, contributed chapters to technical books, and lectured in over 25 countries. Sadiq M. Sait is also the principal author of two books. He is currently Professor of Computer Engineering and the Director of Industry Collaboration at KFUPM.
\end{IEEEbiography}

\begin{IEEEbiography}[{\includegraphics[width=1in,height=1.25in,clip,keepaspectratio]{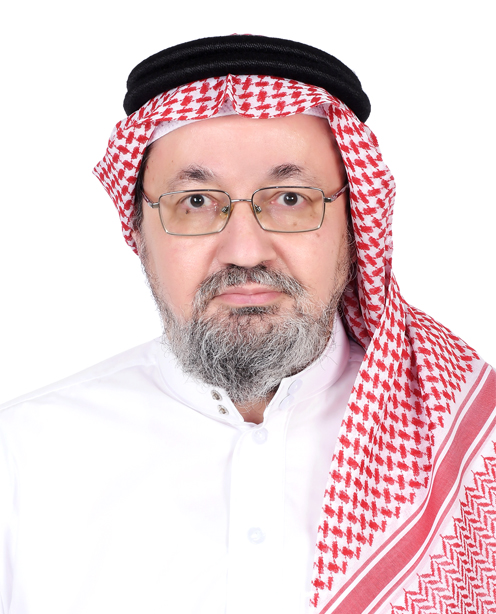}}]{Aiman El-Maleh}
is a Professor in the Computer Engineering Department at King Fahd University of Petroleum \& Minerals. He holds a B.Sc. in Computer Engineering, with first honors, from King Fahd University of Petroleum \& Minerals in 1989, a M.A.SC. in Electrical Engineering from University of Victoria, Canada, in 1991, and a Ph.D in Electrical Engineering, with dean’s honor list, from McGill University, Canada, in 1995. He was a member of scientific staff with Mentor Graphics Corp., a leader in design automation, from 1995-1998. Dr. El-Maleh received the Excellence in Teaching award from KFUPM in 2001/2002, 2006/2007 and 2011/2012, the Excellence in Advising award from KFUPM in 2013/2014 and 2017/2018, the Excellence in Research award from KFUPM in 2010/2011 and 2015/2016, and the First Instructional Technology award from KFUPM in 2009/2010. Dr. El-Maleh's research interests are in the areas of synthesis, testing, and verification of digital systems. In addition, he has research interests in defect and soft-error tolerance design, VLSI design, design automation and efficient FPGA implementations of deep learning algorithms and data compression techniques. Dr. El-Maleh is the winner of the best paper award for the most outstanding contribution in the field of test at the 1995 European Design \& Test Conference. His paper presented at the 1995 Design Automation Conference was also nominated for best paper award. He holds five US patents.
\end{IEEEbiography}

\begin{IEEEbiography}[{\includegraphics[width=1in,height=1.65in,clip,keepaspectratio]{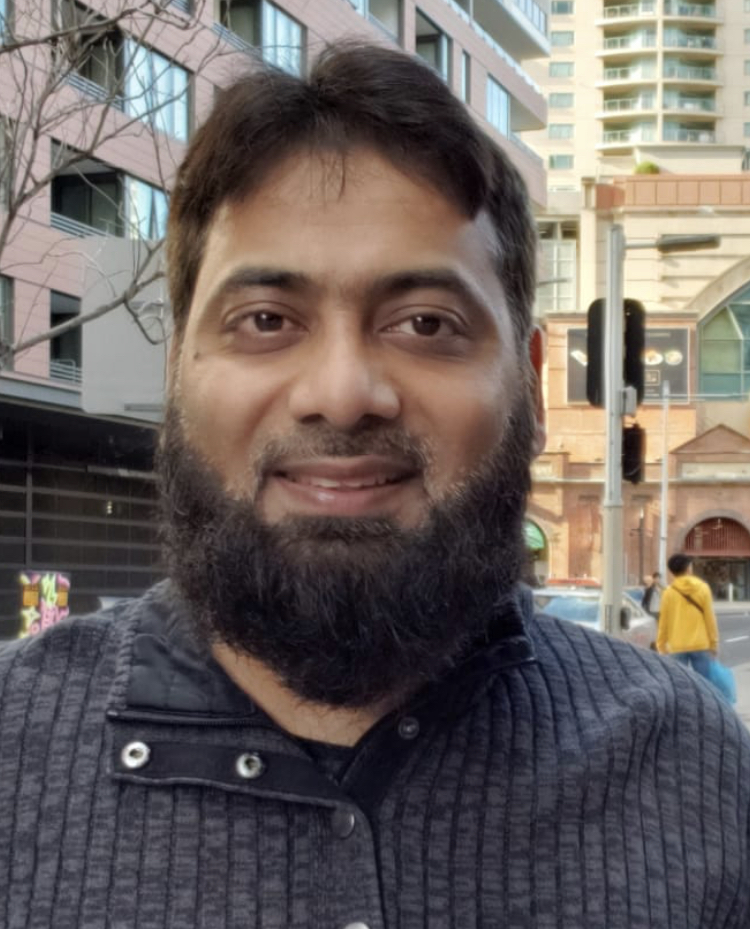}}]{Irfan Ahmad} received the Ph.D. degree in computer science from TU Dortmund, Germany, in 2017. He is currently an Assistant Professor with the Department of Information and Computer Science, KFUPM, Saudi Arabia. He has published several articles in high-quality journals and international conferences. He has also coauthored two book chapters and holds three U.S. patents. His research interests include pattern recognition and machine learning. He regularly reviews articles for well-known journals and conferences in his area of research in addition to being a program committee member for some of the reputed international conferences.
\end{IEEEbiography}
\EOD
\end{document}